%% file: main.tex
\documentclass{article}

\usepackage{iclr2024_conference,times}
\iclrfinalcopy

\usepackage[utf8]{inputenc} 
\usepackage[T1]{fontenc}    
\usepackage{hyperref}       
\usepackage{url}            
\usepackage{booktabs}       
\usepackage{amsfonts}       
\usepackage{nicefrac}       
\usepackage{microtype}      
\usepackage{xcolor}         
\usepackage{multirow}
\usepackage[export]{adjustbox}
\usepackage{graphicx}
\usepackage{subfigure}
\usepackage{comment}
\usepackage[normalem]{ulem}
\usepackage{enumitem}
\usepackage{url}            
\usepackage{floatrow}       
\newfloatcommand{capbtabbox}{table}[][\FBwidth]

\def\yabin{\textcolor{black}}

\def\yabin{\textcolor{black}}

\title{Linguistic Profiling of Deepfakes: An Open Database for Next-Generation Deepfake Detection}

\author{
  Yabin Wang\textsuperscript{\rm 1}, Zhiwu Huang\textsuperscript{\rm 2}, Xiaopeng Hong\textsuperscript{\rm 3, 4}, Zhiheng Ma\textsuperscript{\rm 5}\\
  \textsuperscript{\rm 1}Xi'an Jiaotong University, P. R. China, \textsuperscript{\rm 2}University of Southampton, United Kingdom, \\
  \textsuperscript{\rm 3}Harbin Institute of Technology, P. R. China, \textsuperscript{\rm 4}Pengcheng Laboratory, P. R. China, \\
  \textsuperscript{\rm 5}Shenzhen Institute of Advanced Technology, Chinese Academy of Science, P. R. China,\\
  \texttt{iamwangyabin@stu.xjtu.edu.cn, zhiwu.huang@soton.ac.uk} \\ \texttt{hongxiaopeng@ieee.org},  \texttt{zh.ma@siat.ac.cn}\\
}

\begin{document}

\maketitle

\input{00_abstract}
\input{01_introduction}
\input{02_relatedwork}

\input{03_dataset}

\input{04_benchmark}

\input{05_evaluation}

\input{06_conclusion}
\input{supp}

\bibliographystyle{iclr2024_conference}
\bibliography{references}

\end{document}

%% file: 00_abstract.tex
\begin{abstract}
The emergence of text-to-image generative models has revolutionized the field of deepfakes, enabling the creation of realistic and convincing visual content directly from textual descriptions. However, this advancement presents considerably greater challenges in detecting the authenticity of such content.
Existing deepfake detection datasets and methods often fall short in effectively capturing the extensive range of emerging deepfakes and offering satisfactory explanatory information for detection.
To address the significant issue, this paper introduces a deepfake database (DFLIP-3K) for the development of convincing and explainable deepfake detection.
It encompasses about 300K diverse deepfake samples from approximately 3K generative models, which boasts the largest number of deepfake models in the literature. Moreover, it collects around 190K linguistic footprints of these deepfakes. The two distinguished features enable DFLIP-3K to develop a benchmark that promotes progress in linguistic profiling of deepfakes, which includes three sub-tasks namely deepfake detection, model identification, and prompt prediction. 
The deepfake model and prompt are two essential components of each deepfake, and thus dissecting them linguistically allows for an invaluable exploration of trustworthy and interpretable evidence in deepfake detection, which we believe is the key for the next-generation deepfake detection.
Furthermore, DFLIP-3K is envisioned as an open database that fosters transparency and encourages collaborative efforts to further enhance its growth. Our extensive experiments on the developed benchmark verify that our DFLIP-3K database is capable of serving as a standardized resource for evaluating and comparing linguistic-based deepfake detection, identification, and prompt prediction techniques. 

\end{abstract}

%% file: 01_introduction.tex
\section{Introduction}

The field of deepfakes has experienced a significant shift in recent years, moving beyond traditional generative models focused on realistic images and videos to the emergence of text-to-image (T2I) generative models. These new models have unlocked new possibilities for creating highly realistic and convincing deepfakes. 
In particular, recent T2I generative models like DALL·E \cite{ramesh2022hierarchical}, Imagen \cite{saharia2022photorealistic} and Stable Diffusion \cite{rombach2021highresolution} have revolutionized deepfakes by enabling the direct synthesis of images from textual descriptions or prompts. However, the proliferation of T2I generative models raises serious ethical concerns. The potential for misuse and the rapid dissemination of misleading or false information further intensify concerns about the authenticity and trustworthiness of visual content.

\begin{figure}[t]
\begin{floatrow}
\ffigbox{%
  \centering
  \includegraphics[width=0.42\textwidth]{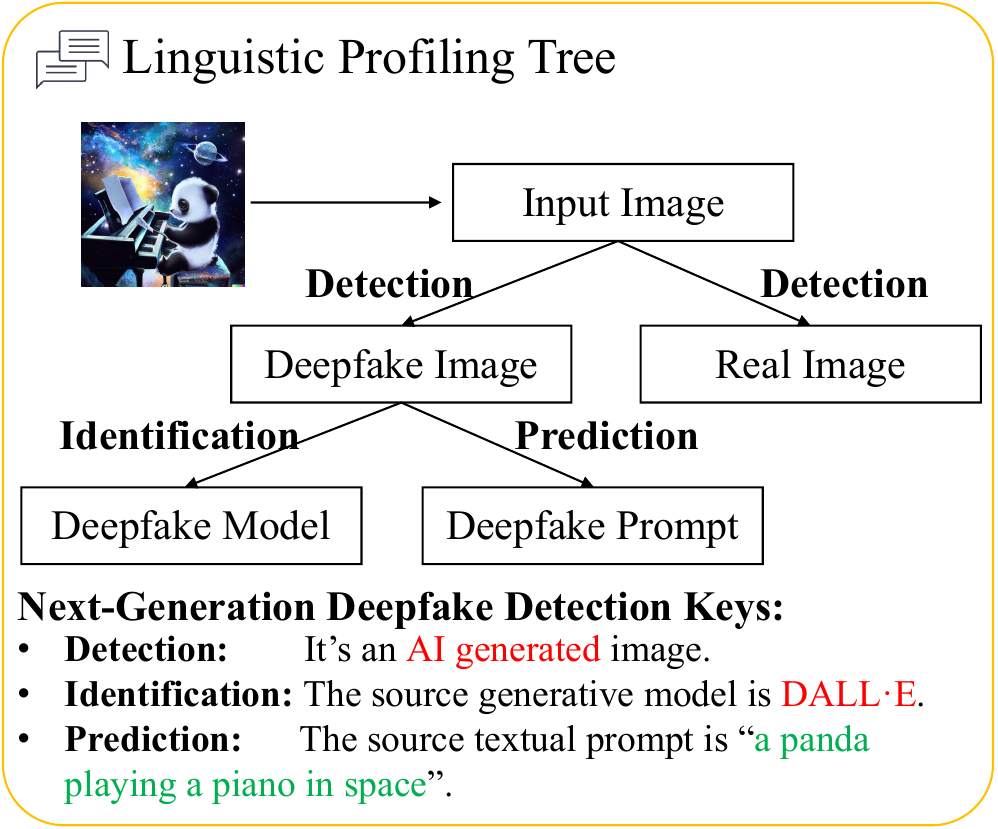}%
}{%
  \caption{DFLIP-3K enables linguistic profiling of deepfakes, including the assessment of authenticity (deepfake detection), the identification of source model (deepfake identification), and the prediction of source prompt (prompt prediction).}%
  \label{fig:dflip_overview}
}
\capbtabbox{%
\centering
        \begin{footnotesize}
        \resizebox{0.96\linewidth}{!}{
        \begin{tabular}{ccc}\hline
        \textbf{Dataset}  & \textbf{\# Generative Models} \\
        \hline
        \multicolumn{2}{c}{\textbf{Early Datasets}}         \\ \hline
        FaceForensic++ \cite{rossler2019faceforensics++}  &  4 \\
        DFDC \cite{dolhansky2020deepfake}      & 8 \\
        WildDeepfake \cite{zi2020wilddeepfake}  & Unknown \\
        CDDB \cite{li2022continual}  &  11 \\
       \hline
        \multicolumn{2}{c}{\textbf{Recent Datasets}}         \\ \hline
        DiffusionDB \cite{wang2022diffusiondb} &    1 \\
        SAC \cite{pressmancrowson2022}    &    4   \\
        Pick-a-Pic \cite{kirstain2023pick}    &   2 \\ 
        HP \cite{wu2023better}    &    1 \\
        MidJourney(250K) \cite{turc2023midjourney}     &   Unknown \\ 
        \textbf{DFLIP-3K (Ours)} & \textbf{$\approx$ 3K }                    \\
          \hline
        \end{tabular}
        }
        \end{footnotesize}
}{%
  \caption{A comparison of DFLIP-3K against the early and recent deepfake datasets. It collects deepfakes from about 3K generative models, representing the largest scale in terms of the number of deepfake models.}%
  \label{tab:deepfake_databases}
}
\end{floatrow}
\vspace{-0.5cm}
\end{figure}

Although many datasets and methods (e.g., \cite{rossler2019faceforensics++,dolhansky2020deepfake,zi2020wilddeepfake,li2022continual,wang2022diffusiondb,pressmancrowson2022,kirstain2023pick,wu2023better,turc2023midjourney}) exist for deepfake detection, they suffer from significant limitations. Firstly, these methods typically focus solely on binary classification, i.e., distinguishing between real and deepfake images, without providing any explanatory information. However, in practice, countering forgery requires comprehensive evidence and explanations to effectively challenge fake content, going beyond a simple binary judgment. Secondly, as in Table~\ref{tab:deepfake_databases}, individual datasets are often generated by a limited number of generative models, resulting in a lack of representation for the extensive diversity of existing deepfakes.

In response to this pressing issue, this paper focuses on the development of \emph{convincing and explainable deepfake detection}, which serves as a foundational step towards next-generation deepfake detection. We posit that the key lies not only in binary deepfake detection but, more importantly, in dissecting the results and presenting them in a manner understandable to humans. Hence, we place significant emphasis on the importance of this new and challenging task, which we term \emph{linguistic profiling of deepfake detection}. 
\yabin{While "deepfakes" is often associated with FaceSwap videos, we broaden the term to include all synthetically generated images, an important area to investigate for combating digital disinformation.}
As shown in Fig.\ref{fig:dflip_overview}, the task of \emph{linguistic profiling of deepfake detection} can be further decomposed into three sub-tasks, namely \emph{the detection of deepfakes, the identification of deepfake source models}, and \emph{the prediction of textual prompts} that are used for text-to-image generation. 
To facilitate this emerging study, we curate an open dataset (DFLIP-3K), which standardizes resources for analyzing the linguistic characteristics of deepfake contents, and provides a benchmark for evaluating and comparing new approaches for linguistic profiling of deepfake detection.

In particular, the established DFLIP-3K database encompasses approximately 300K deepfake samples produced from about 3K generative models, which is the largest scale in the literature of deepfake detection. In addition, we gather about 190K textual prompts that are used to create images. The collected prompts allow for the exploitation of linguistic profiling in simultaneous deepfake detection, identification, and prompt prediction. 
Apart from curating this database, we thoroughly examine the ethical considerations and potential flaws associated with large-scale data collection. By making DFILP-3K publicly available, we provide the community with the initial opportunity to further enhance a database of such application and magnitude.

To assess the potential value of DFLIP-3K, we establish a benchmark for linguistic profiling in simultaneous deepfake detection, identification, and prompt prediction. Based on this benchmark, we conduct several experiments. Our evaluation focuses on state-of-the-art vision-based and vision-language models adapted for deepfake profiling. The results reveal that these vision-language models outperformed traditional vision models in detecting and identifying deepfakes. Unlike vision-based models, our suggested vision-language models have the ability to generate either regular image captions or textual prompts in specific formats for image generation. The evaluations demonstrate that training the vision-language models with our collected prompts leads to more sensible prompt prediction, resulting in reconstructed images that closely resemble the input deepfake images in terms of perceptual, semantic, and aesthetic similarities. Furthermore, we also show that the suggested deepfake profiling facilitates the reliable interpretability of the generated content.

Despite the validation results, DFLIP-3K is not considered a finalized data product at this stage. Given the continuous emergence of generative models, the comprehensive curation of DFLIP-3K for widespread usage extends beyond the scope of a single research paper. Therefore, in addition to releasing the dataset, we are also sharing the software stack that we developed for assembling DFLIP-3K. We consider this initial data release and accompanying paper as an initial step towards creating a widely applicable deepfake dataset suitable for a broader range of linguistic profiling tasks. Consequently, we are making all collection methods, metadata, and deepfake data within our database openly accessible to fully unleash its potential for next-generation deepfake detection.

%% file: 02_relatedwork.tex
\section{Related Works}

\textbf{Deepfake Datasets with Generative Models.} 
Deepfake datasets are vital for advancing development in the field of deepfake detection, manipulation, and understanding. 
The progress in generating deepfake datasets is closely tied to the advancements in generative models. As generative models have evolved significantly, they have enabled the creation of more realistic and diverse deepfakes.

\yabin{Traditional generative models, such as generative adversarial nets (GANs) \cite{goodfellow2020generative, arjovsky2017wasserstein, karras2019style} and variational autoencoders (VAEs) \cite{tolstikhin2017wasserstein, hinton2011transforming}, have played a significant role in creating early deepfake datasets like FaceForensic++ \cite{rossler2019faceforensics++}, DFDC \cite{dolhansky2020deepfake}, WildDeepfake \cite{zi2020wilddeepfake}, Deephy~\cite{narayan2022deephy} and CDDB~\cite{li2022continual}, to name a few.}
However, as the field evolved, the limitations of traditional generative models became apparent. While they were kind of success in generating realistic deepfakes, they lacked fine-grained control and interpretability. This led researchers to explore new approaches for deepfake dataset creation. Text-to-image generative (T2I) models have been emerging as a promising solution, allowing for the generation of deepfake datasets based on textual descriptions. Current state-of-the-art generative models for text-guided image synthesis include the scaled-up GANs like GigaGAN \cite{kang2023gigagan}, autoregressive models such as  DALL·E \cite{ramesh2021zero} and Parti \cite{yu2022scaling}, and diffusion models like DALL·E 2 \cite{ramesh2022hierarchical} and Stable Diffusion \cite{rombach2021highresolution}. Their integration of textual information enables researchers to generate new deepfake datasets (e.g., DiffusionDB \cite{wang2022diffusiondb}, SAC \cite{pressmancrowson2022}, Pick-a-Pic \cite{kirstain2023pick}, HP \cite{wu2023better}, MidJourney(250K) \cite{turc2023midjourney})  that align with specific prompts, facilitating high-degree control and excellent interpretability. 
These datasets are enumerated in Table \ref{tab:deepfake_databases}, and their additional details are provided in the supplementary material.

It is noteworthy that these datasets mainly focus on a limited number of specific models such as Stable Diffusion and MidJourney. Nonetheless,
the field is experiencing rapid emergence of numerous other generative models. To fill this gap, our DFLIP-3K database encompasses deepfake images generated by a significantly larger volume (about 3K) of T2I generative models. This provides a comprehensive resource that best reflects the current landscape of T2I deepfake technologies.

\textbf{Deepfake Detection and Identification.}
Previous studies on deepfake detection have predominantly utilized deep neural networks, such as Residual Network (ResNet) \cite{wang2020cnn}, Vision Transformer (ViT) \cite{wang2021learning}, and Contrastive Language-Image Pretraining (CLIP)~\cite{wang2022sprompt}, as the backbone for training binary classifiers. 
Two recent works \cite{guarnera2023level, ricker2022towards} propose to detect and recognize both GAN-based and recently appeared diffusion-based deepfakes. Their experimental results demonstrated that detecting deepfakes generated by T2I models is more challenging compared to traditional GAN-based deepfakes. However, their solutions mainly focused on Stable Diffusion and LDM \cite{rombach2021highresolution}.

These deepfake detection methods typically concentrate solely on binary classification, while lacking the exploration of evidential and explanatory information. However, combating forgery necessitates comprehensive evidence and explanations to effectively challenge fraudulent content, surpassing a mere binary judgment. Our DFLIP-3K database is proposed to bridge this gap, aiming for the development of convincing and explainable deepfake detection. Particularly, this database allows for leveling up the traditional deepfake detection task to linguistic profiling of deepfakes, which is composed of three sub-tasks, i.e., deepfake detection, deepfake model identification, and prompt prediction. These three sub-tasks serve as crucial elements in dissecting and understanding deepfakes, providing valuable insights for effective detection and analysis.

\textbf{Deepfake Prompt Prediction.}
Prompt prediction (a.k.a., prompt engineering) involves analyzing the text prompts that were used to generate deepfakes. 
Emerging prompt engineering tools like Lexica~\cite{shameemLexicaBuildingCreative2022} allow users to explore textual prompts with various variations and find visually similar images from a database that match the input deepfake. However, the primary drawback of such tools is the requirement for users to be engaged in costly iteration. 
Prompt auto-completers  \cite{oppenlaender2022prompt} complements user-provided prompts using additional keywords provided by statistical distribution predictions to generate higher-quality deepfakes with desired styles.
However, refining these keywords often requires expensive human intervention or assistance.
Others employ image caption models such as CLIP \cite{radford2021learning} and BLIP \cite{li2022blip} to generate prompts. However, merely describing image elements using these models does not guarantee visually appealing results. 

The development of our DFLIP-3K database, which collects textual prompts used for image generation, opens up new avenues for researchers to explore the realm of automatic prompt prediction systems. By profiling the prompts, we can better understand how to describe the style and improve image quality, and gain insights into the intentions, biases, or specific characteristics of the individuals or groups behind the creation of deepfakes.

%% file: 03_dataset.tex
\begin{figure}[t]
\begin{floatrow}
\capbtabbox{%
\centering
        \begin{footnotesize}
        \resizebox{1\linewidth}{!}{
        \begin{tabular}{lrrr} 
        \hline
        Models & \# Models & \# Prompts & \# Images  \\ 
        \hline
        Stable Diffusion \cite{rombach2021highresolution} & 1 & 15,000 & 15,000 \\
        Personalized Diffusion & 3,434(at least) &  120,978 & 280,753 \\
        DALL·E 2 \cite{ramesh2022hierarchical} & 1 & 31,315 & 33,705 \\
        MidJourney \cite{david2022mj} & Unkown & 21,858 & 27,241 \\
        Parti \cite{yu2022scaling} & 1 & 0 & 195 \\
        Imagen \cite{saharia2022photorealistic} & 1 & 0 & 224 \\
        \hline
        \end{tabular}
        }
        \end{footnotesize}
}{
  \caption{Statistics of DFLIP-3K that consists of six big families of generative models. Personalized Diffusions include deepfakes from 3,434 known generative models and a set of unknown models.}%
  \label{tab:dflip3k_statistics}
}
\capbtabbox{%
\centering
        \begin{footnotesize}
        \resizebox{0.95\linewidth}{!}{
        \begin{tabular}{l|ll|ll}
        \hline
        \multirow{2}{*}{Methods} & \multicolumn{2}{c|}{Train} & \multicolumn{2}{c}{Test} \\ \cline{2-5} 
         & Real & Fake & Real & Fake \\ \hline
        Stable Diffusion \cite{rombach2021highresolution} & 5,000 & 5,000 (15,000) & 1,000 & 1,000 \\
        PD 1 & 10,000 & 10,000 (40,000) & 4,000 & 4,000 \\
        DALL·E 2 \cite{ramesh2022hierarchical} & 5,000 & 5,000 (30,961) & 1,000 & 1,000 \\
        MidJourney \cite{david2022mj} & 5,000 & 5,000 (20,867) & 1,000 & 1,000 \\
        Imagen \cite{saharia2022photorealistic} & - & - & 223 & 223 \\
        Parti \cite{yu2022scaling} & - & - & 193 & 193 \\ 
        PD 2 & - & - & - & 72,721 \\ 
        \hline
        \end{tabular}
        }
        \end{footnotesize}
}{%
  \caption{The proposed benchmark setting on DFLIP-3K. Digits outside parentheses are for deepfake detection, and the ones in parentheses are for the other two tasks. Personalized Diffusion (PD) is divided into PD1 (51 Models) for in-distribution training/test,
  and PD2 ($\approx$ 3K models) for out-of-distribution test.}
  \label{tab:traindata}
}

\end{floatrow}
\vspace{-0.5cm}
\end{figure}

\section{Database Collection}
DFLIP-3K is constructed by scraping publicly available high-quality images. It encompasses deepfakes generated by prominent T2I models, namely Stable Diffusion \cite{rombach2021highresolution}, DALL·E 2 \cite{ramesh2022hierarchical}, MidJourney \cite{david2022mj}, Imagen \cite{saharia2022photorealistic}, Parti \cite{yu2022scaling}, and a substantial number of personalized models based on Stable Diffusion. 
Table~\ref{tab:dflip3k_statistics} presents an overview of the DFLIP-3K database.

\subsection{Database Collection Procedure}
\textbf{Overall Dataset Generation Pipeline.}
The fundamental principles guiding our data collection efforts are three-fold: public availability, high-quality images, and reliable sources. To this end, our collection comprises four distinct parts: 1) website selection, 2) web scraping, 3) parsing and filtering, and 4) storing data.
The code used for the whole collection pipeline will be openly available.

\textbf{Website Selection.}
Our objective is to collect a diverse and high-quality database of synthetic data that accurately reflects the latest advancements in T2I deepfakes. To achieve this goal, we first select social media platforms such as Instagram and Twitter, as well as art-sharing websites and imageboards.
\yabin{We meticulously select source websites to ensure that all deepfakes included in our database are created by using recent T2I tools and exhibit high aesthetic quality. }
For DALL·E 2, we retrieve its public database~\cite{dalle2gallery}. It is a website that enables users to search for images by texts. It boasts over 30,000 images, although their quality is fair given the immense quantity. In addition, we also scrape DALL·E generated images that were posted by OpenAI on their social media channels like Instagram and Twitter. These images are carefully selected and are of superior quality.
For Imagen and Parti, their models are not publicly available, and thus we use their officially released data from their respective project pages~\cite{yu2022scaling, saharia2022photorealistic}
as well as their social media channels.
For MidJourney, we use the MidJourney User Prompts \& Generated Images (250k) dataset~\cite{turc2023midjourney}, which crawls messages from the public MidJourney Discord servers. We sample a subset of images that are upscaled by users from this dataset.
Furthermore, we scrape the official showcase gallery~\cite{midjourney} to obtain more aesthetically pleasing images.

To obtain deepfakes generated by Stable Diffusion, we utilize a subset of the data released by DiffusionDB, which is a large-scale T2I prompt dataset. This dataset contains 14 million images with prompts and hyperparameters generated by Stable Diffusion (SD) from official Discord channels. 
However, the aesthetic quality of the images in this dataset are fair. Hence, we additionally select Lexica~\cite{shameemLexicaBuildingCreative2022}, which is an art gallery for artwork created with Stable Diffusion. This platform provides more aesthetically pleasing images.

To avoid any confusion, we use the term 'Stable Diffusion' to refer exclusively to the official base models developed by StabilityAI. Custom models built on top of Stable Diffusion through personal data finetuning are referred to as 'Personalized Diffusions' throughout this paper. In order to enhance the quality of the collected synthetic images, we select several popular imageboards or galleries that specialize in AI-generated illustrations, such as Civitai~\cite{civitai2022}, Aigodlike~\cite{aigodlike}, AIBooru~\cite{aibooruonline}, ArtHub.ai~\cite{arthubai}, FindingArt ~\cite{findingart}, and MajinAI~\cite{majinaiart}. 
These platforms are dedicated to AI art enthusiasts and provide a means for sharing generative models and deepfake images. The images available on these platforms are sourced from both direct user uploads and web crawlers. 
Additionally, some images come with generating parameters, such as prompts, samplers, and other hyper-parameters.

\textbf{Web Scraping.}
We exclusively select publicly available image-sharing platforms that do not require any login credentials. Accordingly, we develop web-scraping tools to extract a comprehensive collection of images. Our web-scraping tool primarily utilizes the BeautifulSoup Python libraries to parse the image download URLs and accessible parameters, such as prompts and models, from the returned HTML messages. To ensure compliance with SFW (safe for work) regulations and avoid downloading age-restricted content, we configure the web-scraping tool to crawl only safe content (if possible). We download all raw images from the parsed URLs and maintain their original formats, such as WEBP, PNG, JPG, etc. We publicly disclose all image URLs and sanitized metadata as we do not claim ownership of the data.

\textbf{Parsing \& Filtering.}
To aggregate images from diverse sources, we archive them based on their generative models. Following the processing and filtering of Common Crawl, we have obtained over 300K samples.
This section provides a guide on processing data into a clean data format.

Parsing data for the official releases of DALL·E, Parti, Imagen, and MidJourney is simple and straightforward, as these sources already identify their generative methods and adhere to strict regulations for publishing SFW content in standardized formats.

Our primary focus is on processing data from other imageboards and websites. 
\yabin{After downloading all the collected deepfake images from various platforms, we apply publicly available NSFW detectors, LAION-SAFETY~\cite{laionsafety}, GantMAN~\cite{man}, CLIP-based-NSFW-Detector~\cite{clipnsfw} and SD-SAFETY~\cite{SDSAFETY}, to filter out potentially inappropriate images. For more implementation details and information, please refer to the supplementary material.}
To address the issue of duplicate images generated with the same prompt and hyper-parameters but using different random seeds, we use fastdup~\cite{visuallayer2022fastdup} to remove duplicates and similar images with the same prompts. Besides, we filter out any collapsed images that failed to download.
To obtain the parameters for generated images, we employ two methods: scraping metadata from websites and parsing original raw images. Standard format of the scraped data suffices for the former method. For the latter, we try to extract raw information used to generate the images if possible. 
Currently, widely used open-source AI-generated tools such as Automatic1111-Stable-Diffusion-WebUI~\cite{Automatic11112022} automatically embed information about the hyperparameters utilized to generate images into PNG files. Therefore, it is feasible to directly obtain uploaded images in PNG format and extract the crucial parameters, including prompts, the name of the model used for generation, and the model hash.

We utilize search engines, such as Google and CivitAI, to search for model information and download URLs. However, there exist a significant number of images with unknown model hashes, which are likely generated by personalized or merged models. Additionally, enormous images are generated using additional networks, such as embeddings (via textural inversion~\cite{gal2022image}), LoRA~\cite{hu2021lora}, and ControlNet~\cite{zhang2023adding}, among others. It is challenging, and sometimes impossible, to extract the exact additional networks utilized, as individuals tend to withhold this information. However, these networks are primarily used to encode personalized characters or poses and have a relatively minor impact on image quality and style. Thus, we rely on base models as the primary source for generating images.

Since some deepfakes contain watermarks or logos (e.g., DALL·E, Imagen, and Parti), we address this issue by cropping these images to eliminate potentially trivial methods for detection. 

\subsection{Real Data Collection and Deepfake Database Preparation}
To construct a deepfake detection database, it is crucial to include high-quality real data as a counterpart to the deepfake samples. All real images utilized in DFLIP-3K are gathered from LAION-5B~\cite{schuhmann2022laion}, which is a database containing 5 billion image-text-pairs crawled from web pages between 2014 and 2021.
We choose LAION-5B for two primary reasons. Firstly, given the recent emergence of deepfake technologies, it is very unlikely that high-quality deepfakes were easily accessible on the internet before 2021. 
Secondly, this dataset has been used to train the Stable Diffusion model, and as such, we can leverage it as negative samples to construct a training set for deepfake detection.

We select a subset of deepfakes from our database and employ an image retrieval tool~\cite{clipretrieval} provided by LAION-5B to search for high-quality and visually similar real images. We search for images that have an aesthetic score of over $8$, and select the most similar image from LAION-5B.

\subsection{Statistics}
Table~\ref{tab:dflip3k_statistics} presents the statistics of our DFLIP-3K database. We collect over 300K images, and after cleaning and filtering, we obtain a dataset of 189,151 images with prompts, paired with over 3K models. For Personalized Diffusions, we gather 280,753 images from either 3,434 parsed models or a set of unknown models.
For Stable Diffusion, we consider different versions as one model, such as v1.4 and v1.5, as they only differ in the number of training iterations, while using the same training dataset. Similarly, in Personalized Diffusions, we treat different versions of models with the same name as the same model, as their versions only have slight differences in style.
It is worth noting that half of the models in Personalized Diffusions have less than 10 images. 
Regarding the number of models in MidJourney subset, it is unknown to us since we are unsure of their model structures and whether they use the same model for each operation.

\textbf{Prompt Analysis.} 
After collecting data, we analyze the prompts provided by users who utilize T2I models.
We observe that these prompts differ significantly from our usual natural language. 
Many users submit prompts containing comma-delimited phrases that impose desirable constraints. These instructions often include the elements that should be present in the generated images, and begin with words that describe the quality of the image, such as "Masterpiece" or "8K", followed by specific elements such as "cat", "dog", or "person".
Furthermore, we find that users of DALL·E, MidJourney, and Stable Diffusion tend to use longer sentences that are closer to natural language, whereas many personalized diffusion models use pure words or tags to describe images. 
This phenomenon may be due to the fact that many fine-tuned models use image tag estimation tools, such as DeepDanbooru~\cite{deepdanbooru}, to label the training images, while other models' captions are often more closely related to natural language annotations.

For personalized diffusion models, their textual prompts follow specific grammatical structures, i.e., the Stable-Diffusion-Webui~\cite{Automatic11112022} grammar.
This grammar utilizes various symbols to manipulate the model's attention towards specific words or implement certain control. For instance, the use of parentheses "()" increases the model's focus on the enclosed words, while the use of brackets "[]" decreases it. In addition, there are negative prompts that instruct the model to avoid generating particular objects in the deepfake image. This is achieved by using the negative prompt for unconditional conditioning in the sampling process, instead of an empty string. 
Furthermore, additional networks, such as LoRA~\cite{hu2021lora}, can be incorporated into the base model by using the format "<lora:LoRA\_Name>". 
\yabin{To facilitate easy training purposes, we also offer a clean prompt version alone with the raw prompts that excludes symbols and special characters. }

\begin{figure}[t]
\begin{floatrow}

\ffigbox{%
  \centering
    {\scriptsize
    \def\teaserwid{0.15\linewidth}
    \begin{tabular}{c@{\hspace{.5mm}}*{11}{c@{\hspace{.5mm}}}}
    \raisebox{-0.45\height}{\includegraphics[width=\teaserwid]{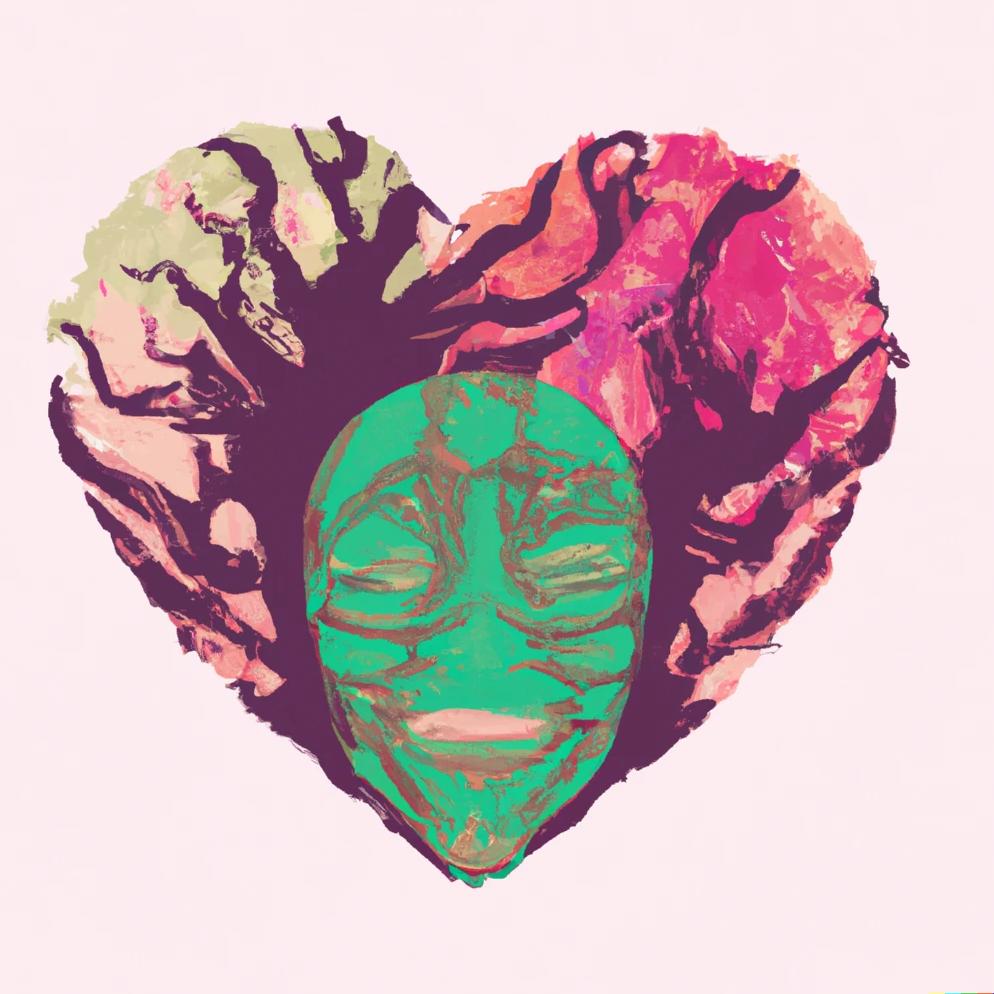}}&  
    \raisebox{-0.45\height}{\includegraphics[width=\teaserwid]{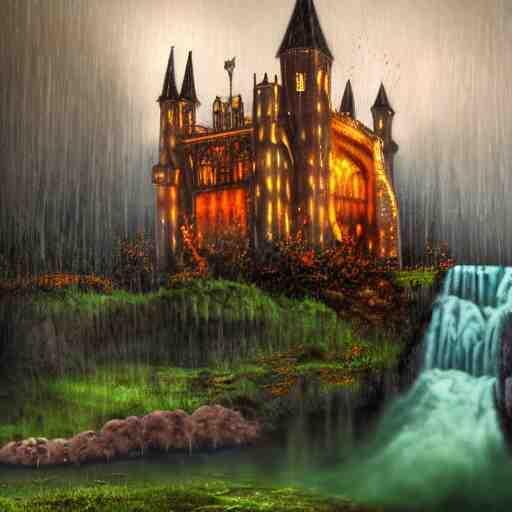}}&
    \raisebox{-0.45\height}{\includegraphics[width=\teaserwid]{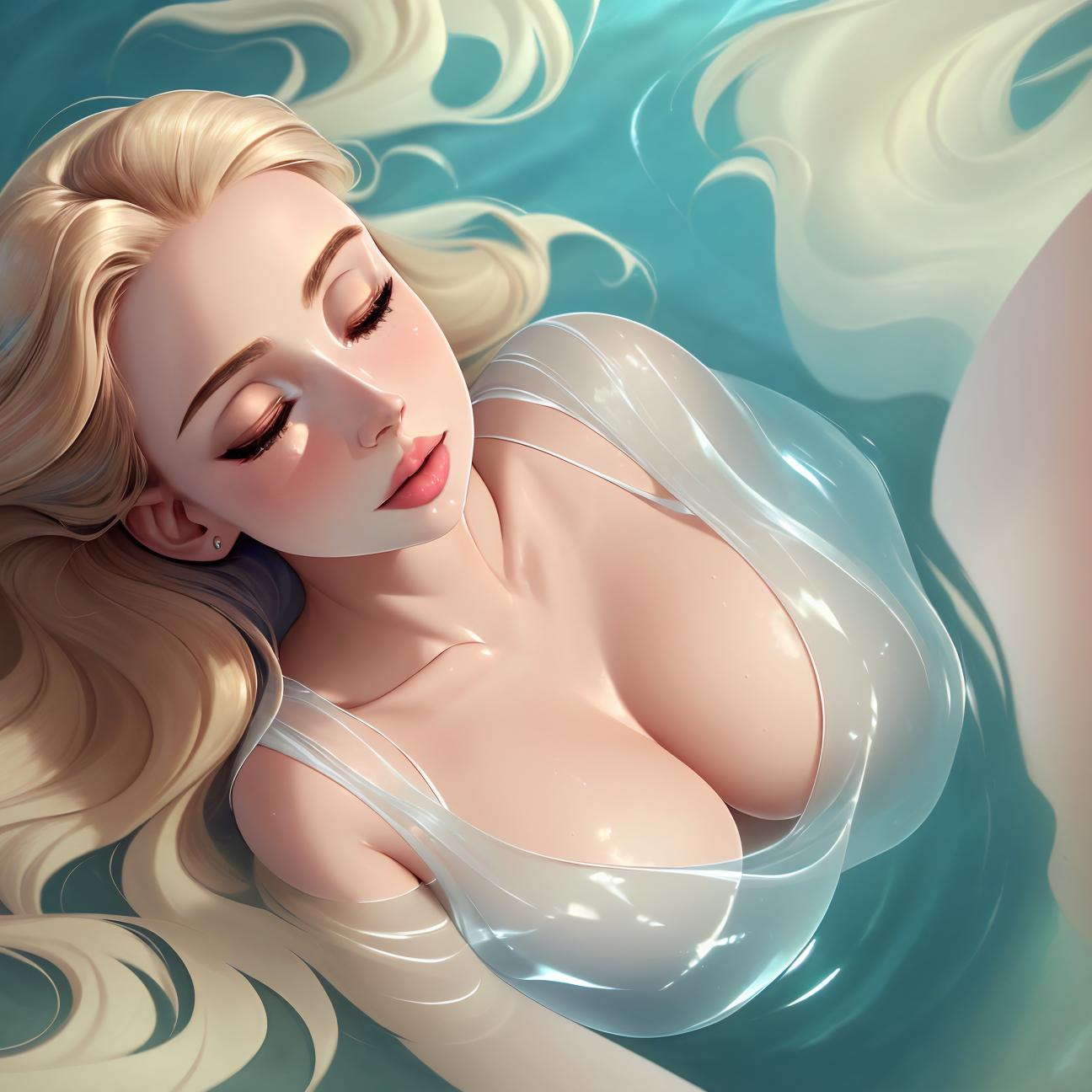}}&
    \raisebox{-0.45\height}{\includegraphics[width=\teaserwid]{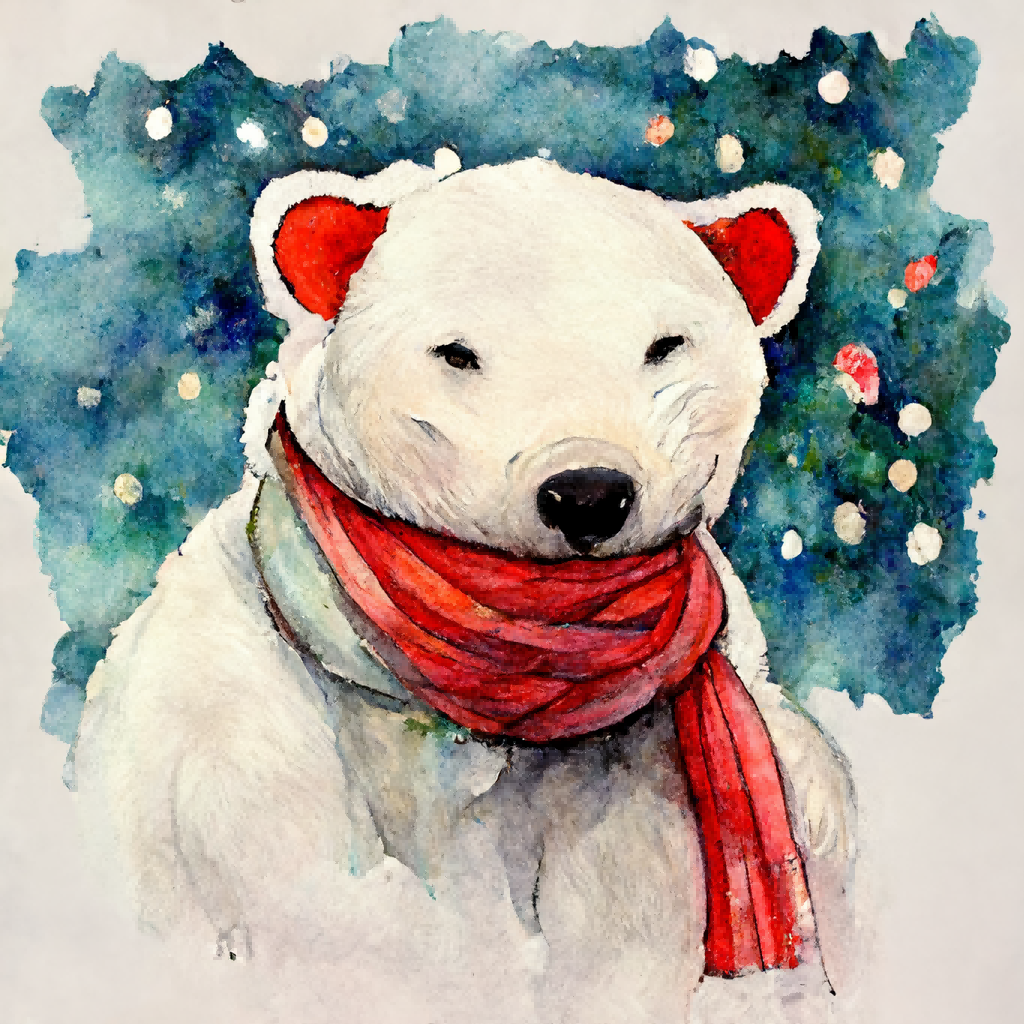}}&
    \raisebox{-0.45\height}{\includegraphics[width=\teaserwid]{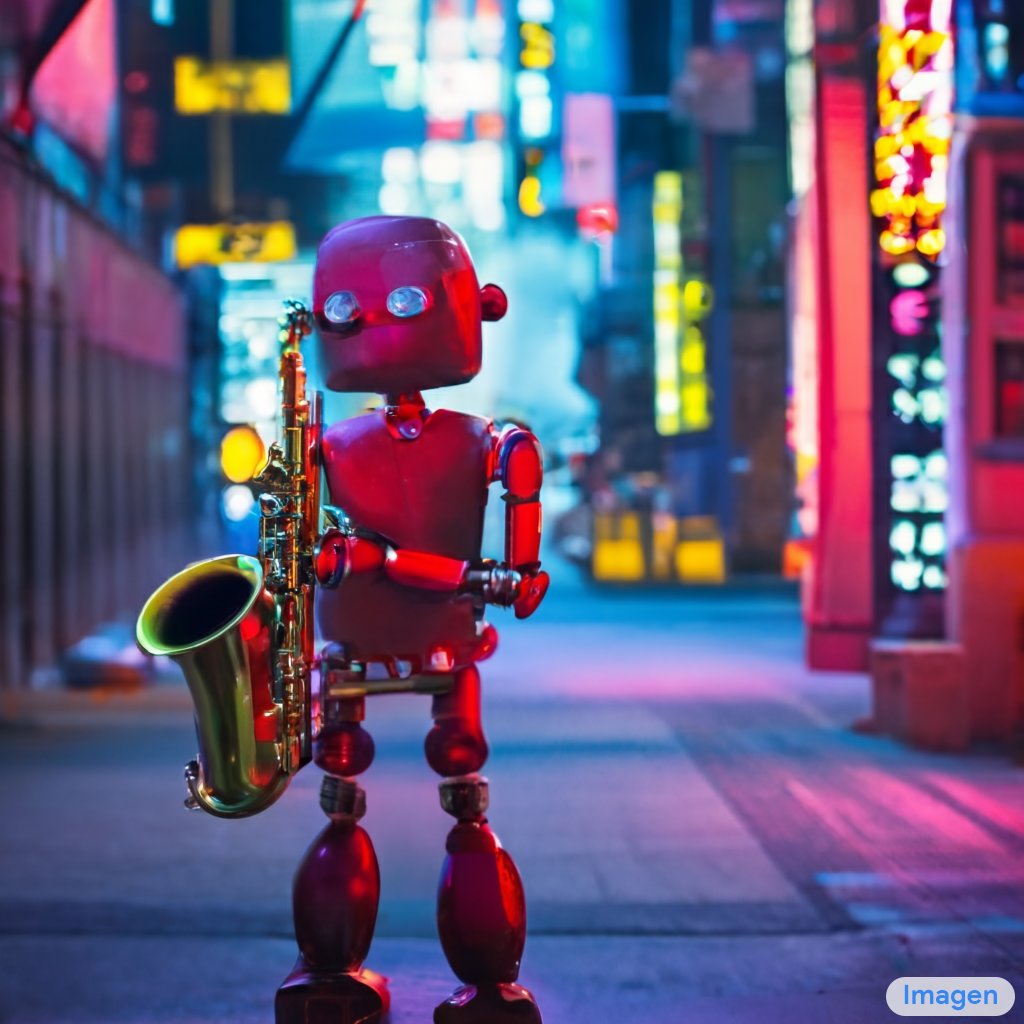}}&
    \raisebox{-0.45\height}{\includegraphics[width=\teaserwid]{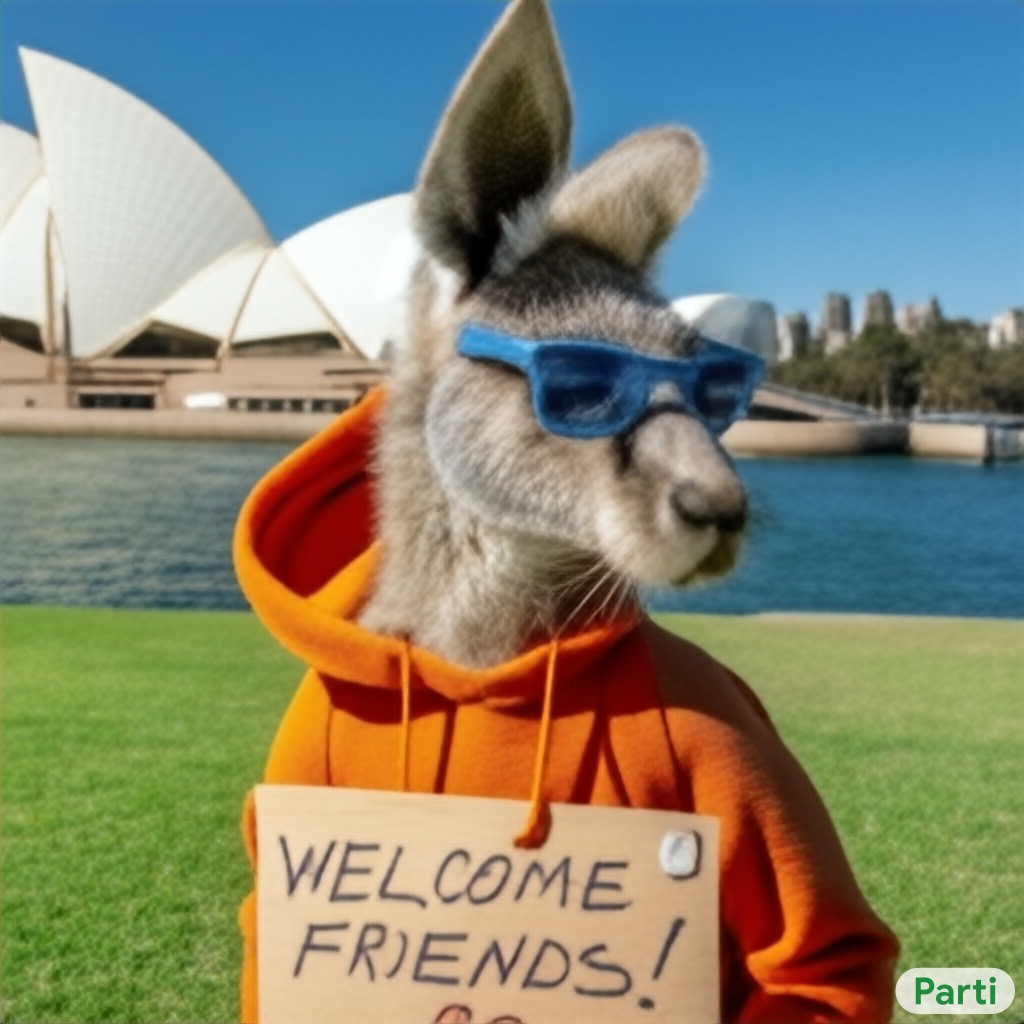}}&
    \vspace{.05in}
    \\
    \raisebox{-0.45\height}{\includegraphics[width=\teaserwid]{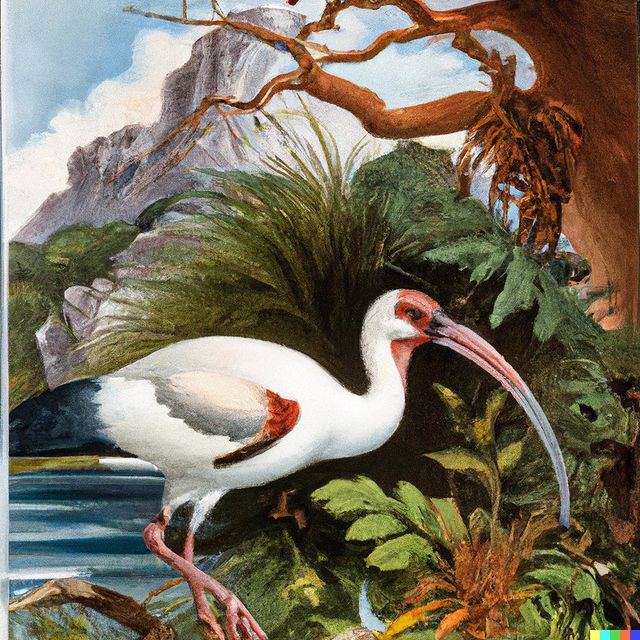}}&  
    \raisebox{-0.45\height}{\includegraphics[width=\teaserwid]{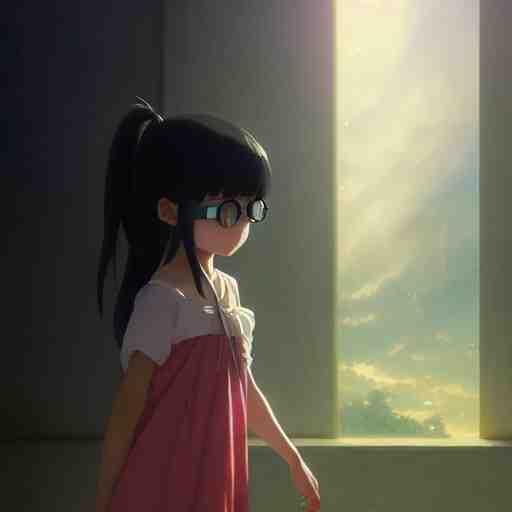}}&
    \raisebox{-0.45\height}{\includegraphics[width=\teaserwid]{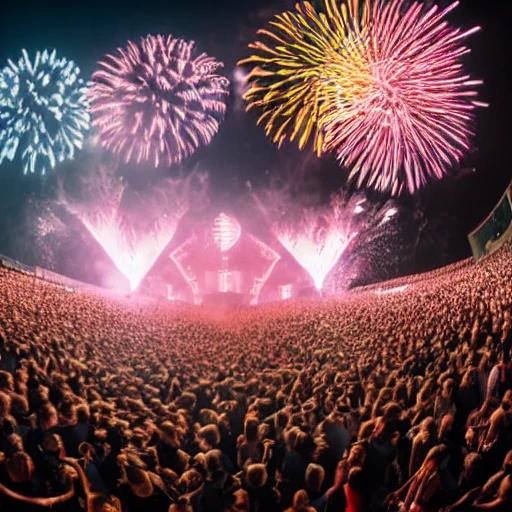}}&
    \raisebox{-0.45\height}{\includegraphics[width=\teaserwid]{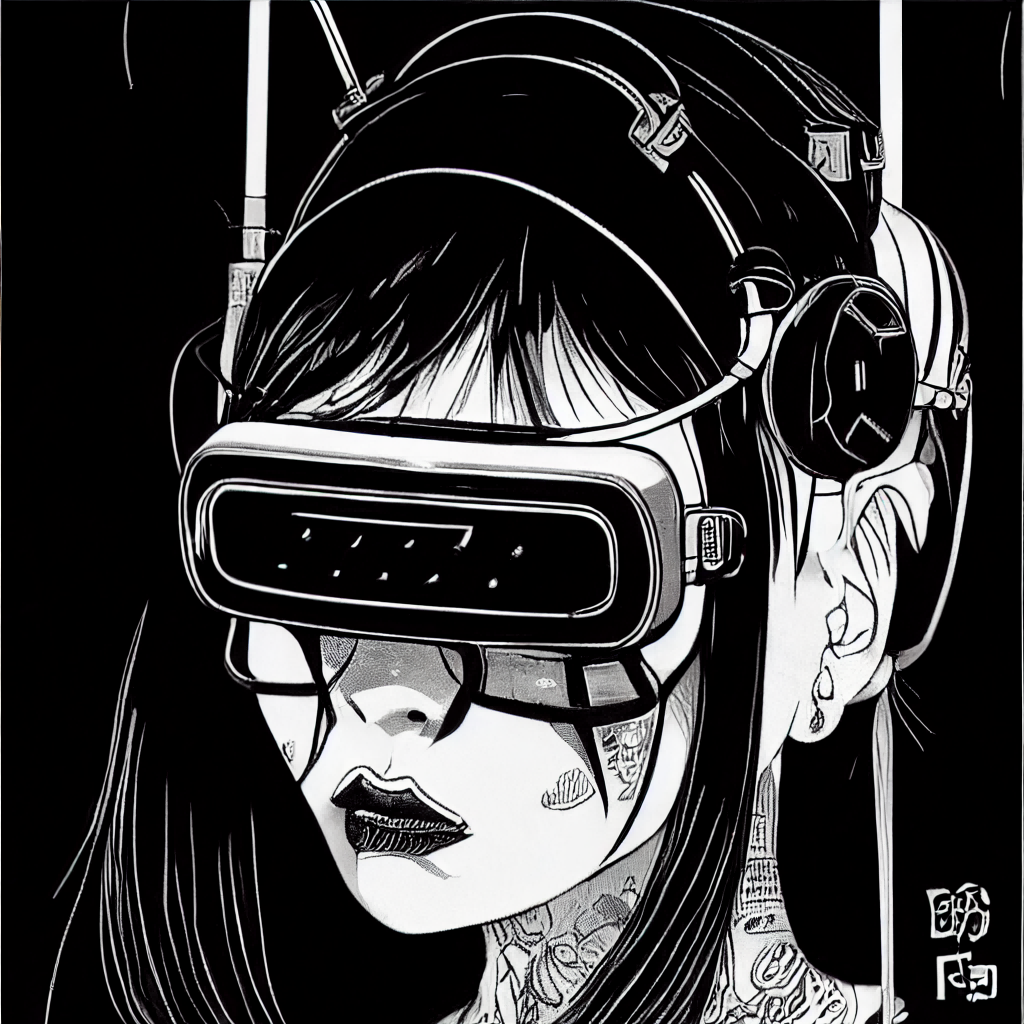}}&
    \raisebox{-0.45\height}{\includegraphics[width=\teaserwid]{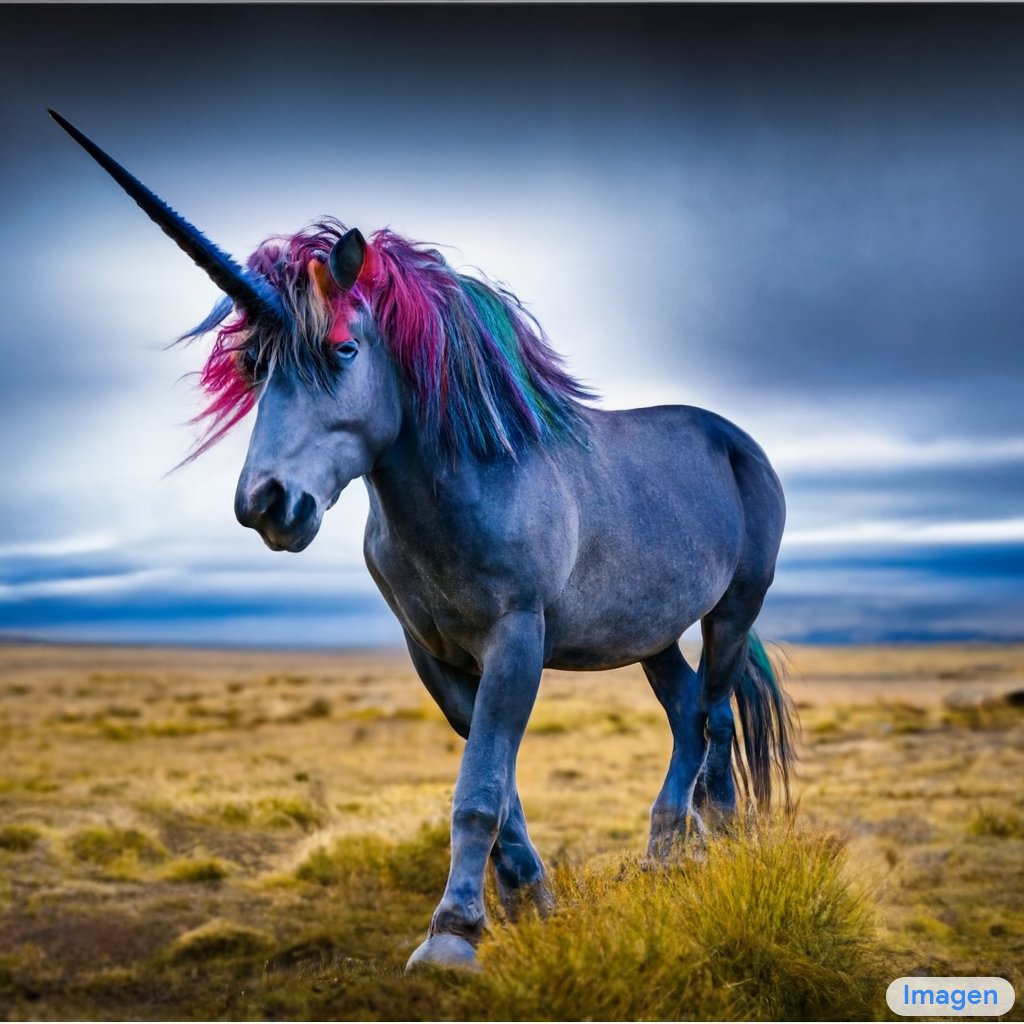}}&
    \raisebox{-0.45\height}{\includegraphics[width=\teaserwid]{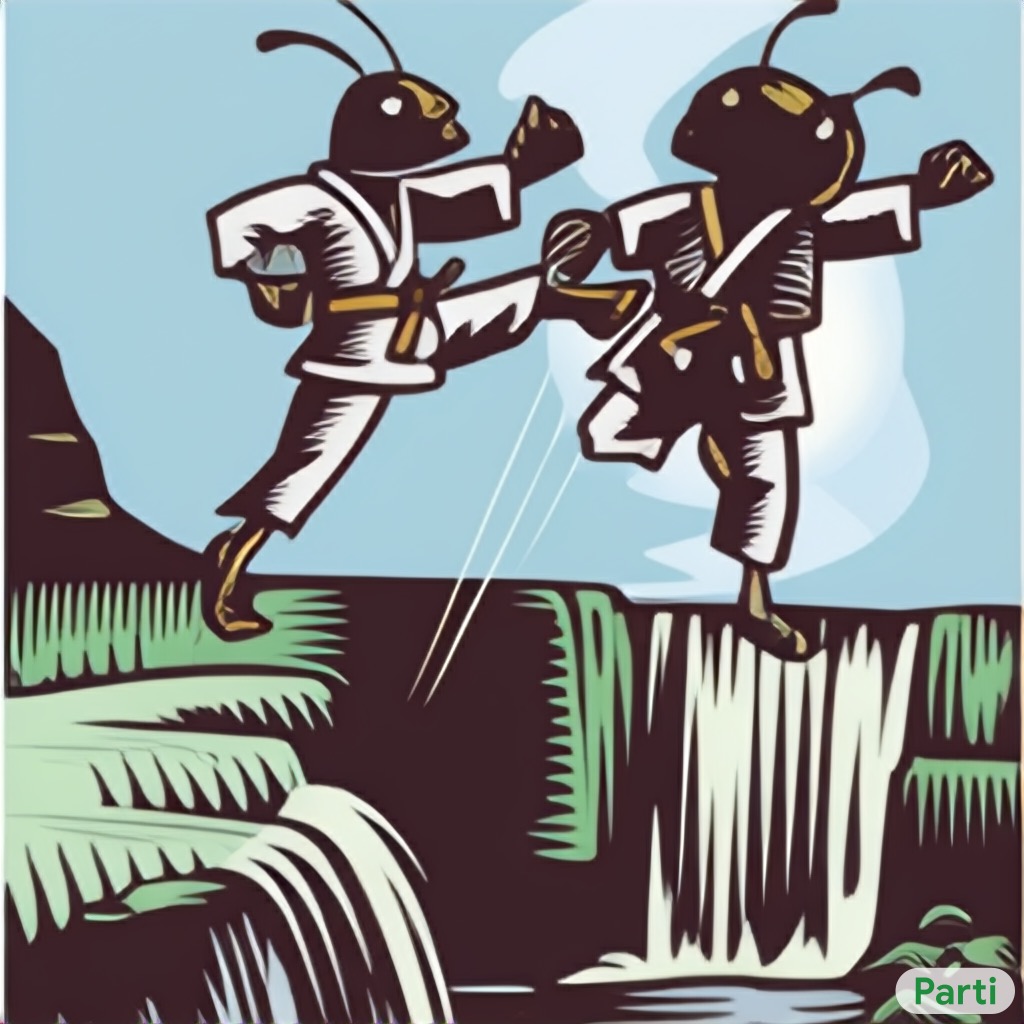}}
    \vspace{.05in}
    \\
    DALL·E\cite{ramesh2022hierarchical} &  
    SD \cite{rombach2021highresolution} &
    PD &
    MJ \cite{david2022mj}&
    Imagen \cite{saharia2022photorealistic} &
    Parti \cite{yu2022scaling}
    \vspace{-.1in}
    \\
    \end{tabular}
    }
}{
  \caption{DFLIP-3K examples generated by the six groups of generative models. More examples are presented in the Supplementary Material.}%
  \label{fig:dflip3k_example}
}
\ffigbox{
  \centering
  \includegraphics[width=0.4\textwidth]{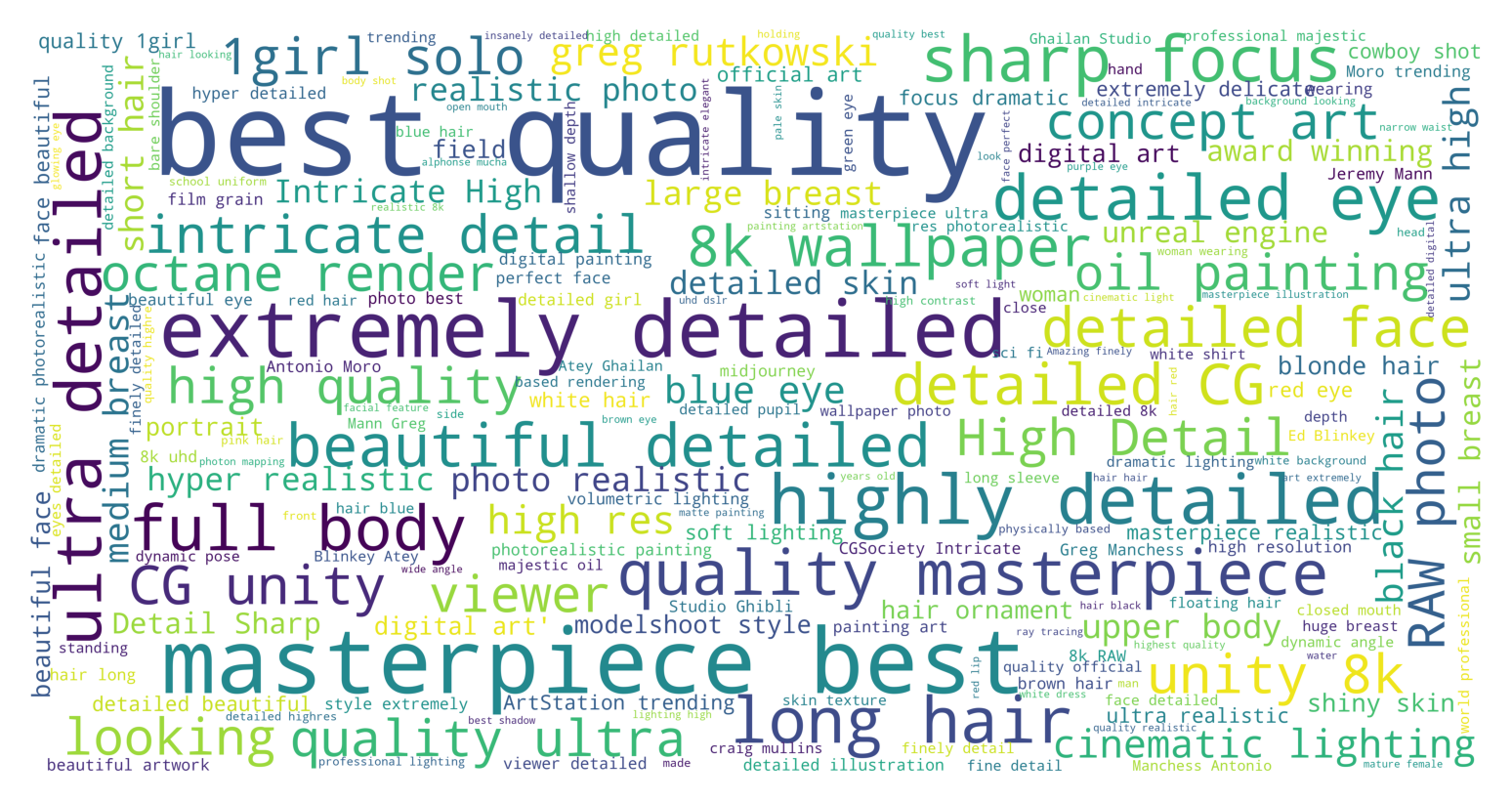}%
}{
  \caption{Word clouds of our collected textual prompts for all associated generative models. More word clouds are in Supplementary Material.}
  \label{fig:wordcloud}
}

\end{floatrow}
\vspace{-0.5cm}
\end{figure}

To illustrate the commonly used words in our collected textual prompts, 
we conduct frequency analysis, excluding special words and punctuation. The resulting word cloud (Figure~\ref{fig:wordcloud}) shows the top 200 most frequent words.

%% file: 04_benchmark.tex
\section{Benchmark Setup}
Based on DFLIP-3K, we develop a benchmark to validate its efficacy as a standardized resource for evaluating methods for linguistic profiling of deepfake detection,
which include the three sub-tasks: 1) Deepfake Detection, 2) Deepfake Model Identification, and 3) Prompt Prediction (Fig.\ref{fig:dflip_overview}).

{Deepfake detection} is one of the fundamental sub-tasks, particularly in light of recent advances in large-scale T2I generative models that enable individuals to create high-quality deepfakes. As these deepfakes blur the distinction between reality and fantasy, differentiating them from non-AI generated images becomes increasingly challenging.
Moreover, it is crucial to identify which deepfake model is used to generate an image, as this can either serve as an evidence for deepfake detection or aid model publishers in safeguarding their models. Nonetheless, identifying the origins of deepfakes remains a non-trivial task, given their high diversity and the prevalence of personalized models trained on private data.
Furthermore, recent advancements in T2I models have led to the emergence of the prompt prediction problem. The predicted prompts forms the other evidence for deepfake detection. Based on the resulting prompts, we can better understand and interpret the created content.

To effectively train deep learning models that can accomplish the three tasks at hand, it is imperative to perform additional data processing.
This is primarily due to some models having few images, rendering them inadequate for training purposes.
Following our initial data collection, we have conducted a thorough pre-screening process and present the resulting processed dataset in Table~\ref{tab:traindata}. Our aim is to ensure that all generative models included in the training set have a minimum of 100 images, with 50 images for testing. As a result, we now have a total of 54 models for training, comprising 51 personalized diffusion models (PD1 in Table~\ref{tab:traindata}), as well as DALL·E, MidJourney, and Stable Diffusion. Further details on the selected models can be found in the supplementary material.
In addition to the generative models, we also selected 25,000 real images to serve as a separate category in the deepfake identification task. Moving forward, we have categorized the remaining models of Personalized Diffusion (PD2 in Table~\ref{tab:traindata}), Parti, and Imagen for out-of-distribution test.

%% file: 05_evaluation.tex
\section{Experiments and Results}
In this section, we present our experimental setting along with the detailed training procedures. 
We report the performance of baseline methods and discuss limitations.

\subsection{Evaluated Methods}
For deepfake detection and deepfake model identification, we select two state-of-the-art methods, CNNDet~\cite{wang2020cnn} and S-Prompts~\cite{wang2022sprompt} as our baseline approaches.
In particular, we follow them to use vision-based network ResNet-50~\cite{he2016deep}, ViT-base-16~\cite{dosovitskiy2020image}, as well as vision-language based network CLIP~\cite{radford2021learning} as our baselines for these two tasks.
Prompt prediction is a recently emerging task, for which there are no dedicated methods available yet. Hence, we choose the state-of-the-art captioning method, BLIP~\cite{li2022blip}, as our baseline, which merely performs prompt prediction.

To accomplish all the three sub-tasks of linguistic deepfake profiling, we suggest exploiting Flamingo~\cite{Alayrac2022FlamingoAV} that leverages pre-trained vision-language models to accept images and texts as input and generate free-form texts as output. We discover that language models can be used to unify the three different sub-tasks in a simple and efficient manner. For instance, in the case of deepfake detection, we convert the real and fake labels in the dataset to a dialogue format, such as `Question: Is this image generated by AI? Answer: This is an AI generated image by Stable Diffusion' or `Answer: This is a real image.' Similarly, for deepfake identification, the answer to the above question can determine the deepfake source model. 
For prompt prediction, we design the image prompts as a question-answer format, such as `Question: Give me prompts using Stable Diffusion to generate this image. Answer: I suggest using ChilloutMix and this prompt: 8k, RAW photo, best quality, masterpiece, realistic, photo-realistic, ultra-detailed, 1 girl, solo, beautiful detailed sky street, standing, nose blush, closed mouth, beautiful detailed eyes, short hair, white shirt, belly button, torn jeans.'

We use the OpenFlamingo-9B~\cite{anas_awadalla_2023_7733589} implementation, which uses a CLIP ViT-Large vision encoder and a LLaMA-7B~\cite{touvron2023llama} language model that is trained on large multimodal datasets like Multimodal C4~\cite{zhu2023multimodal} and LAION-2B~\cite{schuhmann2022laion}. 
We further fine-tune OpenFlamingo-9B on our DFLIP-3K database using the aforementioned question-answering data format.


\begin{figure*}[t] 
\centering 
\includegraphics[width=0.9\textwidth]{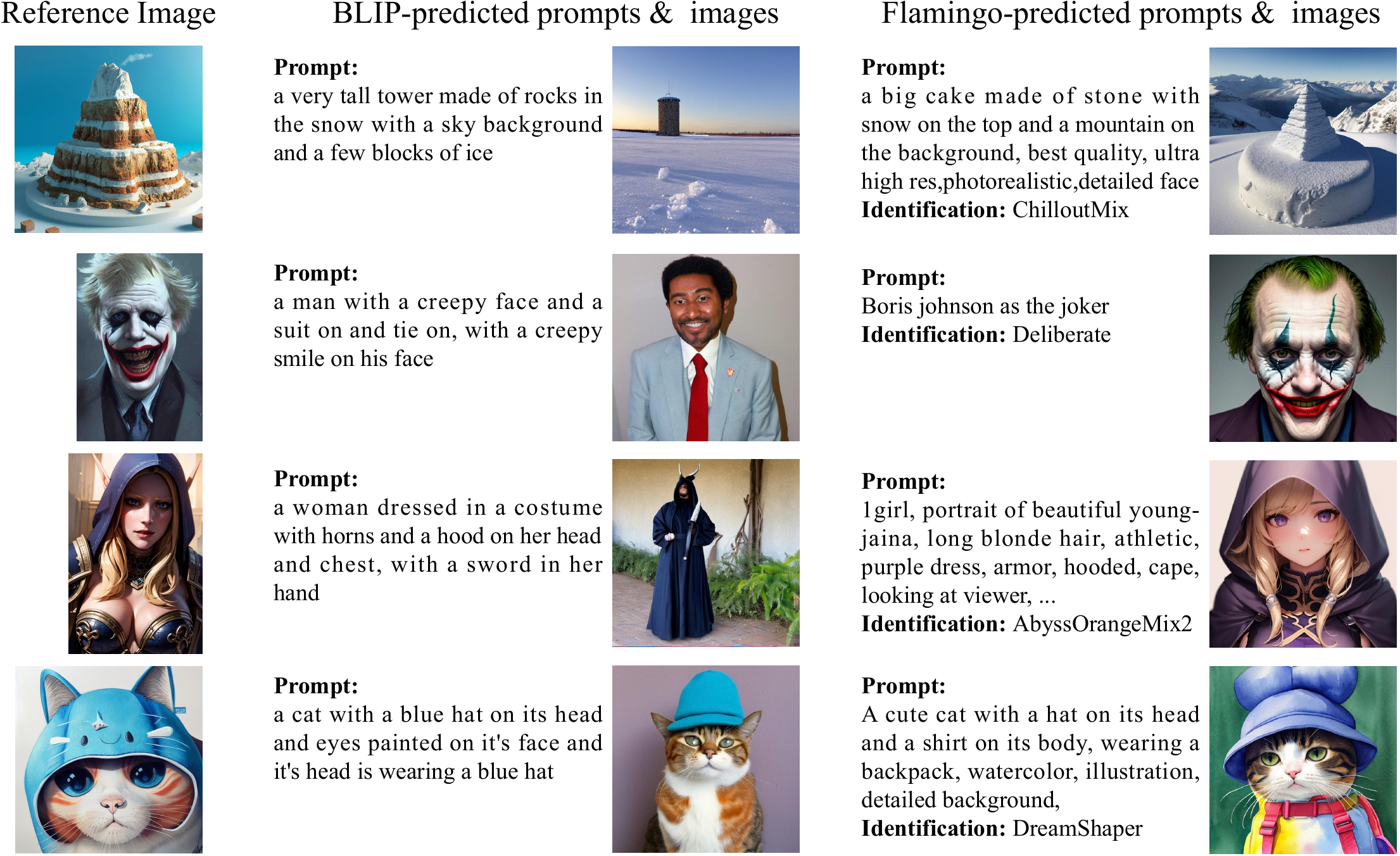} 
\caption{Visualization of Prompt Prediction Results. The first column displays the input reference images used for prompt prediction, while the second and third columns showcase the prompts predicted by BLIP and the corresponding images generated by Stable Diffusion v1.5, respectively. Similarly, the fourth and fifth columns exhibit the fine-tuned Flamingo-predicted prompts and the images generated by models predicted by Flamingo, respectively. \emph{Compared with BLIP, Flamingo can jointly perform model identification and prompt prediction, and its predicted prompts interpret the image contents more faithfully (like `cake' vs. 'tower'), providing more global styles (like `high res') and  more local attributes (like `a cute cat'). Thanks to the better dissection on the two essentials (source model and prompt), Flamigo's resulting images look visually closer to the reference images.} }
\label{fig:vis1}
\vspace{-0.2cm}
\end{figure*}

\subsection{Evaluation Metrics}
Following~\cite{wang2020cnn, li2022continual}, we use average detection accuracy to measure the deepfake detection results.
For deepfake model identification, we calculate multiclass accuracy, which calculates the accuracy that the picture is classified into the correct deepfake model or real image.
To evaluate the quality of predicted prompts, we select $1,000$ images from test set as reference images to do prompt prediction. To assess the quality of predicted prompts and models serving as two essential evident for deepfakes, we suggest measuring the similarity between each input deepfake and the reconstructed deepfakes, which is generated by feeding the predicted prompt into the identified model.
For BLIP, as it only outputs image captions, we use Stable Diffusion v1.5 to produce the reconstructed deepfakes.
In contrast, since Flamingo can identify deepfake model jointly, we use its predicted models and prompts to generate the reconstructed deepfakes.
We adopt CLIP-Score~\cite{gal2022image}, Learned Perceptual Image Patch Similarity (LPIPS)~\cite{zhang2018perceptual} and LAION-Aesthetic Score~\cite{laionaesthetics} to measure the semantic, perceptual, and aesthetic similarities respectively.
Please find training and testing details in Supplementary Material.

\begin{table}[]
\resizebox{1\linewidth}{!}{
\begin{tabular}{lrrrrrrrrrr}
\hline
\multicolumn{1}{c}{} & \multicolumn{4}{c}{In-Distribution Eval.} & \multicolumn{1}{c}{} & \multicolumn{3}{c}{Out-of-Distribution Eval.}  \\ \cline{2-5} \cline{7-9}
\multicolumn{1}{c}{\multirow{-2}{*}{Methods}} & \multicolumn{1}{c}{SD} & \multicolumn{1}{c}{PD1} & \multicolumn{1}{c}{DALL·E} & \multicolumn{1}{c}{MJ} & \multicolumn{1}{c}{\multirow{-2}{*}{AC $(\uparrow)$}} & \multicolumn{1}{c}{Imagen} & \multicolumn{1}{c}{Parti} & \multicolumn{1}{c}{PD2} & \multicolumn{1}{c}{\multirow{-2}{*}{AC $(\uparrow)$}} \\ 
\hline
CNNDet-ResNet \cite{wang2020cnn} & 66.70 & 85.65 & 80.95 & 77.50 & 77.70 & 76.45 & 66.32 & 88.94 & 77.24  \\
CNNDet-ResNet(0.1) \cite{wang2020cnn} & 69.00 & 81.70 & 75.55 & 74.15 & 75.10 & 77.13 & 61.92 & 91.09 & 76.71\\
CNNDet-ResNet(0.5) \cite{wang2020cnn} & 66.40 & 80.67 & 72.35 & 70.80 & 72.56 & 70.85 & 62.18 & \textbf{93.30} & 75.44\\
SPrompts-ViT \cite{wang2022sprompt}   & 65.25 & \textbf{88.72} & \underline{83.35} & \textbf{82.80} & \underline{80.03} & 75.56 & 74.35 & 88.90 & 79.60\\
SPrompts-ViT(0.1) \cite{wang2022sprompt} & \textbf{74.00} & 84.68 & 79.85 & 77.45 & 79.00 & 75.78 & 68.13 & 91.80 & 78.57 \\
SPrompts-ViT(0.5) \cite{wang2022sprompt} & 70.75 & 82.76 & 75.30 & 73.45 & 75.57 & 74.89 & 63.47 & 91.12 & 76.49 \\
SPrompts-CLIP \cite{wang2022sprompt}  & 65.95 & 85.56 & 80.50 & 77.30 & 77.30 & \textbf{82.51} & \underline{86.27} & 88.07 & \underline{85.62}\\
Flamingo \cite{Alayrac2022FlamingoAV}  & \underline{72.00} & \underline{88.32} & \textbf{86.90} & \underline{82.10} & \textbf{82.33} & \underline{78.25} & \textbf{89.90} & \underline{92.14} & \textbf{86.76} \\ 
\hline
\end{tabular}
}
\caption{Results of deepfake detection. (0.1) and (0.5) mean the augmentation with 10\% or 50\% probability~\cite{wang2020cnn}, respectively. SD: Stable Diffusion, MJ: MidJourney, PD1/PD2: two subsets of Personalized Diffusions, AC: average detection accuracy, \textbf{Bold}: \textbf{best}, \underline{Underline}: \underline{second best}}
\label{tab:deepfake}
\vspace{-0.3cm}
\end{table}

\begin{figure}[t]
\begin{floatrow}
\capbtabbox{%
  \centering
        \resizebox{0.8\linewidth}{!}{
        \begin{tabular}{lll} 
        \hline
        Methods    & Accuracy $(\uparrow)$  \\ 
        \hline
        CNNDet-ResNet \cite{wang2020cnn} & 46.43    \\
        SPrompts-ViT  \cite{wang2022sprompt} & 51.74    \\
        SPrompts-CLIP \cite{wang2022sprompt} &  53.10 \\
        Flamingo \cite{Alayrac2022FlamingoAV} & 63.39    \\
        \hline
        \end{tabular}
        }        
}{%
  \caption{Results of deepfake model identification.}
  \label{tab:identification}
}
\capbtabbox{%
\centering
        \resizebox{1\linewidth}{!}{
        \begin{tabular}{lllll} 
        \hline
        Methods & CLIP $(\uparrow)$ & LPIPS $(\uparrow)$ & Aesthetic $(\uparrow)$  \\ 
        \hline
        BLIP~\cite{li2022blip} & 0.65 & 0.67 & 6.30  \\
        Flamingo~\cite{Alayrac2022FlamingoAV} & 0.66   & 0.72   & 6.97  \\
        \hline
        \end{tabular}
        }
}{%
  \caption{Similarities between reference and reconstructed deepfakes by either identified (Flamingo) or default models (BLIP) over predicted prompts.}
  \label{tab:prompt}
}
\end{floatrow}
\vspace{-0.6cm}
\end{figure}



\subsection{Evaluation Results}

\textbf{Deepfake detection.} 
Table ~\ref{tab:deepfake} reports the results of deepfake detection.
Some conclusions are listed as follows:
(1) 
Different from the observation from~\cite{wang2020cnn} over GAN-based deepfakes, the Gaussian blur and JPEG augmentation brings clear performance degradation over the non-augmentation case in terms of detection accuracy. 
(2) Pre-trained vision-language models, such as CLIP and Flamingo, show superior performance in detecting out-of-distribution deepfakes compared to traditional vision models, such as ResNet and ViT. These models leverage both visual and textual information, enabling a more comprehensive analysis of the input data. The incorporation of multimodal information lead to clear advancements in deepfake detection.

\textbf{Deepfake Identification.}
Table~\ref{tab:identification} reports the results of deepfake model identification.
Our results demonstrate that the pre-trained vision-language models, such as CLIP and Flamingo, outperform traditional pre-trained vision models in accurately identifying the deepfake model used to generate deepfakes.
In particular, Flamingo exhibits superior performance in deepfake identification, possibly due to its massive size, pretraining data, huge amount of network parameters as well as its excellent vision-language learning mechanism. 
Furthermore, the correctly identified deepfake models provide a substantial amount of evidence for convincing deepfake detection.

\textbf{Prompt Prediction.}
Figure~\ref{fig:vis1} visualizes the predicted prompts from BLIP and Flamingo. From the results, we can find that Flamingo's predicted prompts describe the given reference images more faithfully. For example, Flagmio understands the main content of the first reference image correctly as `a big cake', while BLIP misunderstands it as `a very tall tower'. Moreover, the predicted prompts of Flamingo is capable of providing more global styles like `ultra high res' and `detailed background', as well as more local attributes like `a cute cat' and `beautiful young jaina'. The more faithful textual prompts contribute to a more reliable interpretation of deepfakes. In addition, we also show their reconstructed deepfakes produced by feeding the predicted prompts to either the predicted models (Flamingo) or the default model (BLIP). The results reflect that Flamingo's reconstructed deepfakes look clearly closer to their reference deepfakes. This shows that our suggested linguistic profiling mechanism over DFLIP-3K is able to provide the two types of trustworthy evidences for deepfake detection. In addition, table~\ref{tab:prompt} presents the corresponding quantitative evaluation. The results demonstrate that Flamingo's reconstructed deepfakes are more similar to the reference deepfakes in terms of  semantic, perceptual and aesthetic scores, aligning well with the visual comparison case.



%% file: 06_conclusion.tex
\section{Conclusion and Discussion}
DFLIP-3K exhibits a significant potential to advance next-generation deepfake detection. Its vast size and quality make it an invaluable resource for furthering research in linguistic profiling of deepfakes.

\textbf{Intended Use:}
During the collection and processing of the DFLIP-3K database, we dedicated great effort to ensuring its high quality. All data were obtained from AI-generated
images and underwent a meticulous selection process to ensure their high aesthetic value. Furthermore, we processed the data to enable clear inference for users, including getting prompts and models.
To ensure the safety of our data, we applied safe-for-work filters to our images and will continue to enhance our filter tools. We encourage researchers to develop more effective and general models on the DFLIP-3K database.

\textbf{Limitations:}
It is crucial to acknowledge that DFLIP-3K, like many image generation datasets, is subject to biases during data collection, such as those related to gender and race. Researchers should take care to consider these biases when using this database and use it judiciously.
Additionally, DFLIP-3K is a highly imbalanced dataset, with varying sizes of data generated by different sources. However, we caution against assuming that models with fewer pictures are less important. On the contrary, small sets of deepfakes that are difficult to detect may have significant impacts. As with any other datasets, there is a risk that individuals may misuse the information contained therein. 

%% file: supp.tex









\newpage
\vspace*{\fill}
{
\centering
\LARGE Linguistic Profiling of Deepfakes: An Open Database for Next-Generation Deepfake Detection\\
\centering
Supplementary Material \\ [1em]
}

\vspace*{\fill}



The supplementary material of this paper provides a comprehensive datasheet of the DFLIP-3K database. Additionally, it presents in-depth information about the database, the implementation of benchmarking methods, as well as the training and testing procedures employed in the study. The material also includes supplementary experiments and visualizations that further support the findings of the main paper. Furthermore, it offers expanded descriptions into related datasets and features a checklist that complements the main paper. The DFLIP-3K database and the associated code can be accessed and downloaded from the GitHub repository available at \url{https://github.com/dflip3k/DFLIP-3K}.

\section{Database Details}
We conduct further analysis on several important factors related to our constructed DFILP-3K database, including the collected textual prompts and the used hyper-parameters. Our investigations shed light on key aspects that contribute to the overall performance of our database, and provide valuable insights for the future research.

\subsection{Prompt Lengths} 
\begin{figure}[http]
  \centering
    \includegraphics[width=0.8\linewidth]{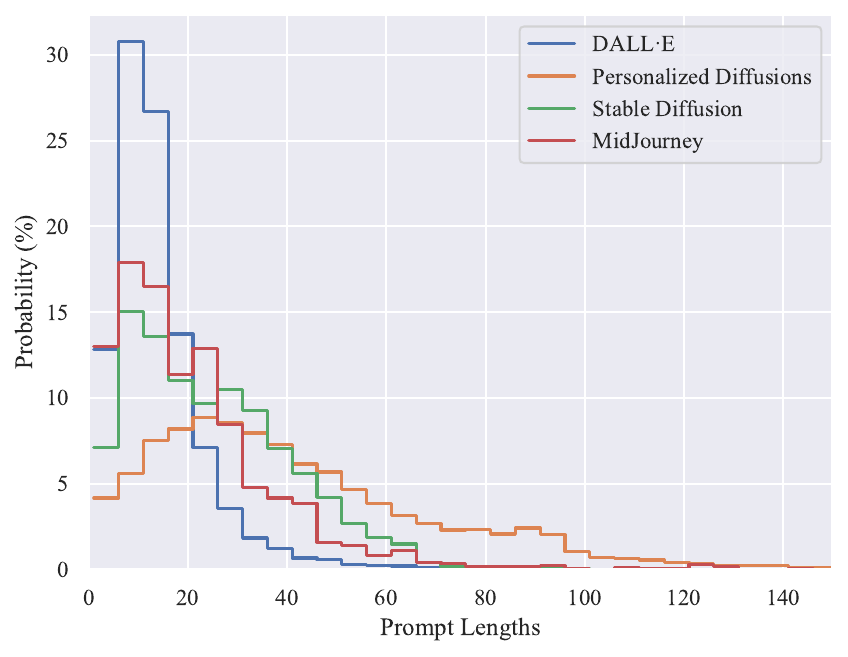}
  \caption{Prompt Lengths of different deepfake methods.}
  \label{Fig.promptlength}
\end{figure}

We segment each textual prompt into a list of words by delimiters such as commas, and periods, and then calculate the number of words in each prompt as its prompt length. 
We calculate the proportion of prompts with varying prompt lengths applied on different generative models and illustrate them in Figure~\ref{Fig.promptlength}.
As per our observations, the majority of the textual prompts have a length of 20 words. DALL·E, Stable Diffusion, and MidJourney exhibit similar prompt length distributions, whereas Personalized Diffusions have a significantly higher proportion of longer prompts.

\subsection{Prompt Frequency Analysis}

It is also crucial to provide a comprehensive analysis of the most commonly employed words in prompts across various generative models. To this end, we collect the frequency of each word, and filter out unique tokens, punctuation, stop words, and subtokens that are difficult to assign to a single concept.

\begin{figure}
  \centering
  \subfigure[Wordcloud of DALL·E]{
    \includegraphics[width=0.45\linewidth]{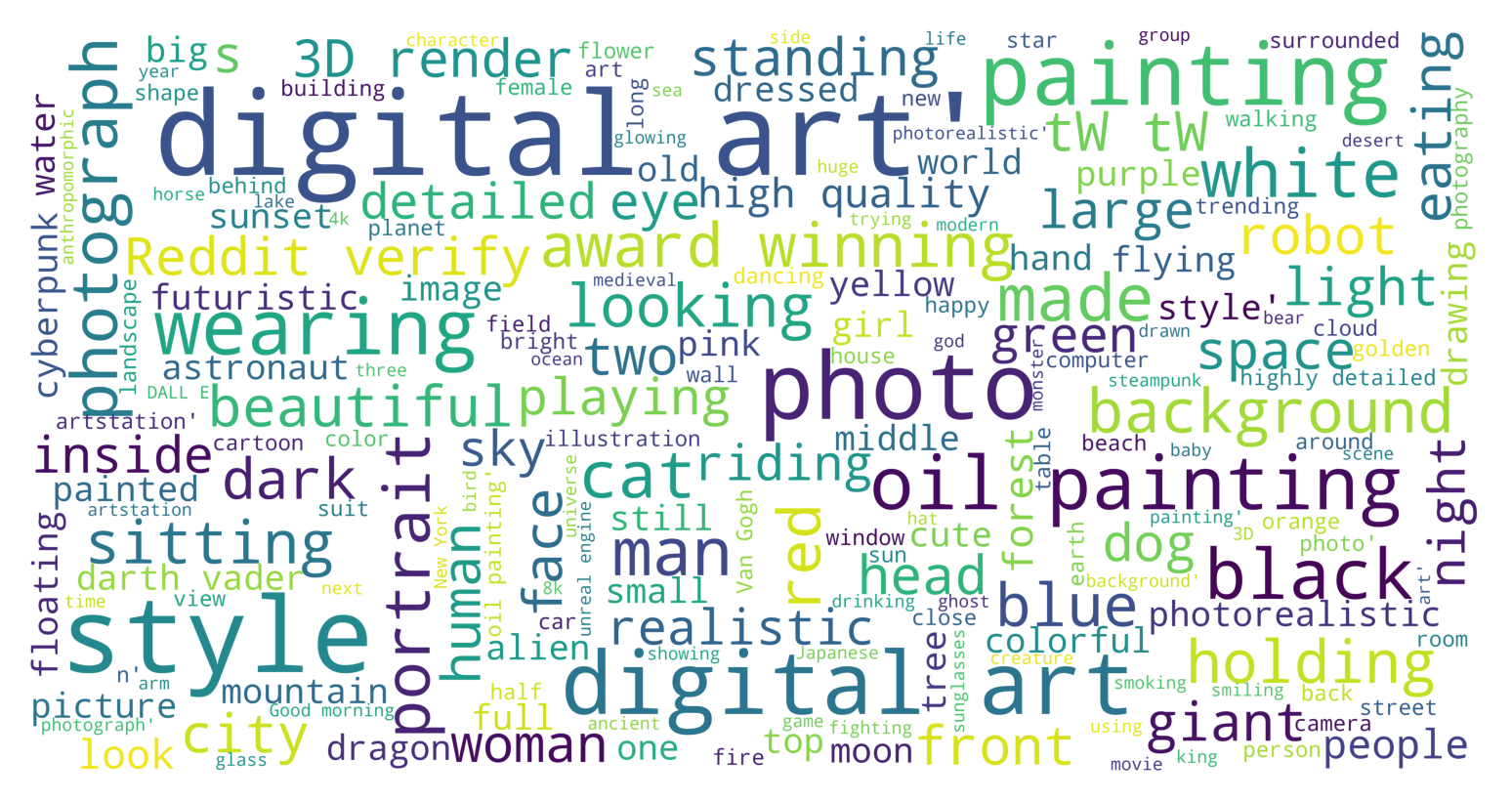}
  }
  \subfigure[Top 20 words in DALL·E's prompts]{
    \includegraphics[width=0.45\linewidth]{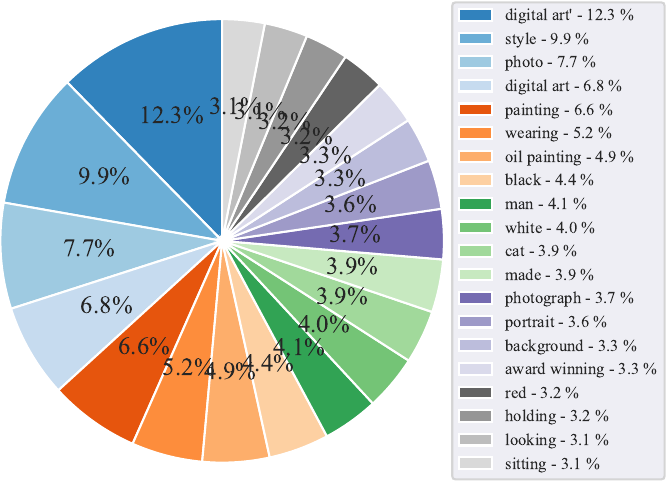}
  }
  \\
  \subfigure[Wordcloud of Stable Diffusion]{
    \includegraphics[width=0.45\linewidth]{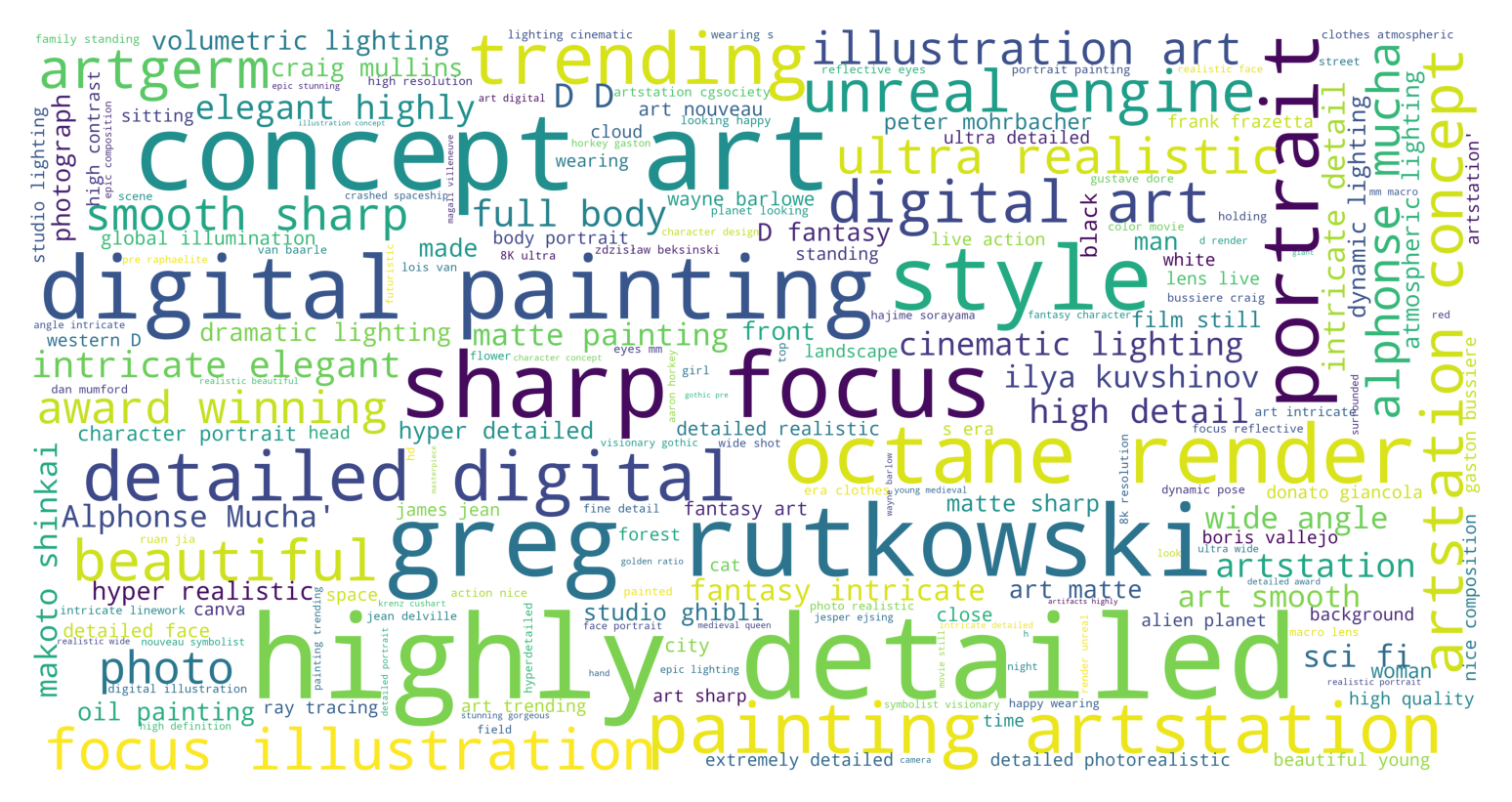}
  }
  \subfigure[Top 20 words in Stable Diffusion's prompts]{
    \includegraphics[width=0.45\linewidth]{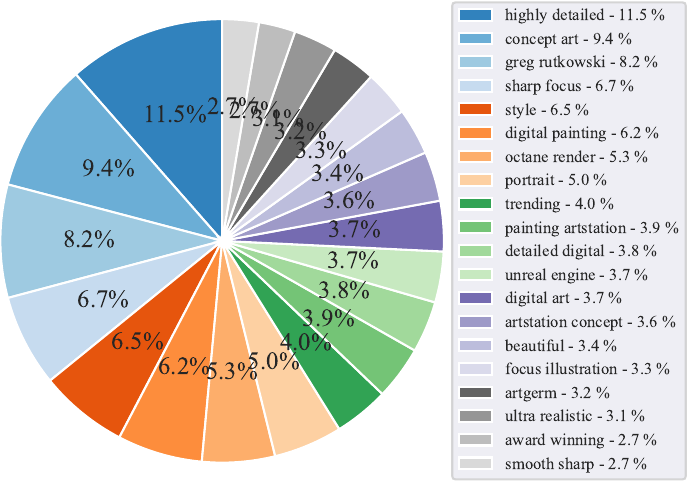}
  }
  \\
  \subfigure[Wordcloud of MidJourney]{
    \includegraphics[width=0.45\linewidth]{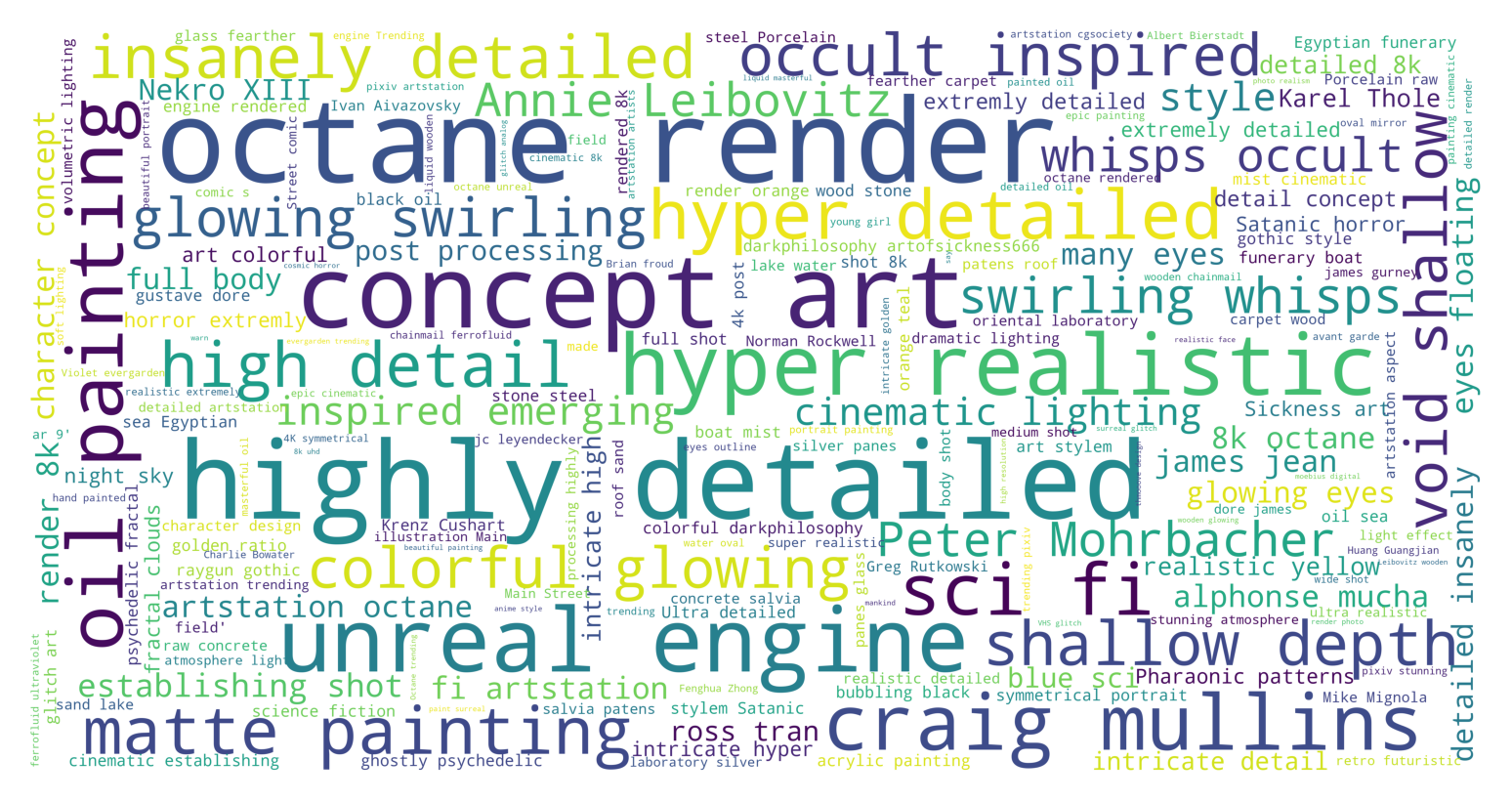}
  }
  \subfigure[Top 20 words in MidJourney's prompts]{
    \includegraphics[width=0.45\linewidth]{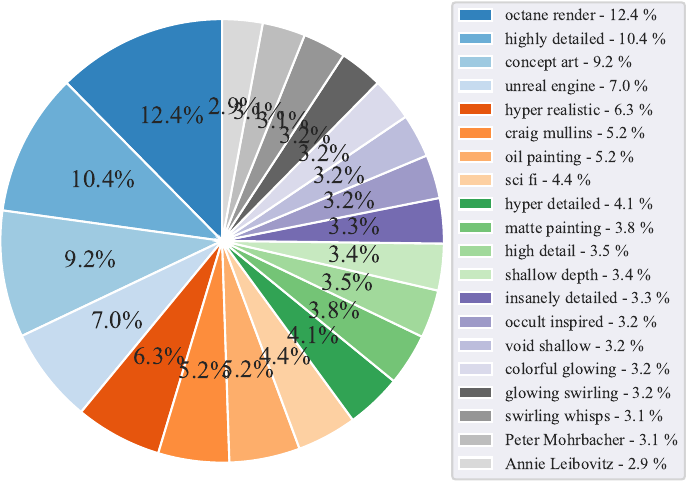}
  }
  \\
  \subfigure[Wordcloud of Personalized Diffusions]{
    \includegraphics[width=0.45\linewidth]{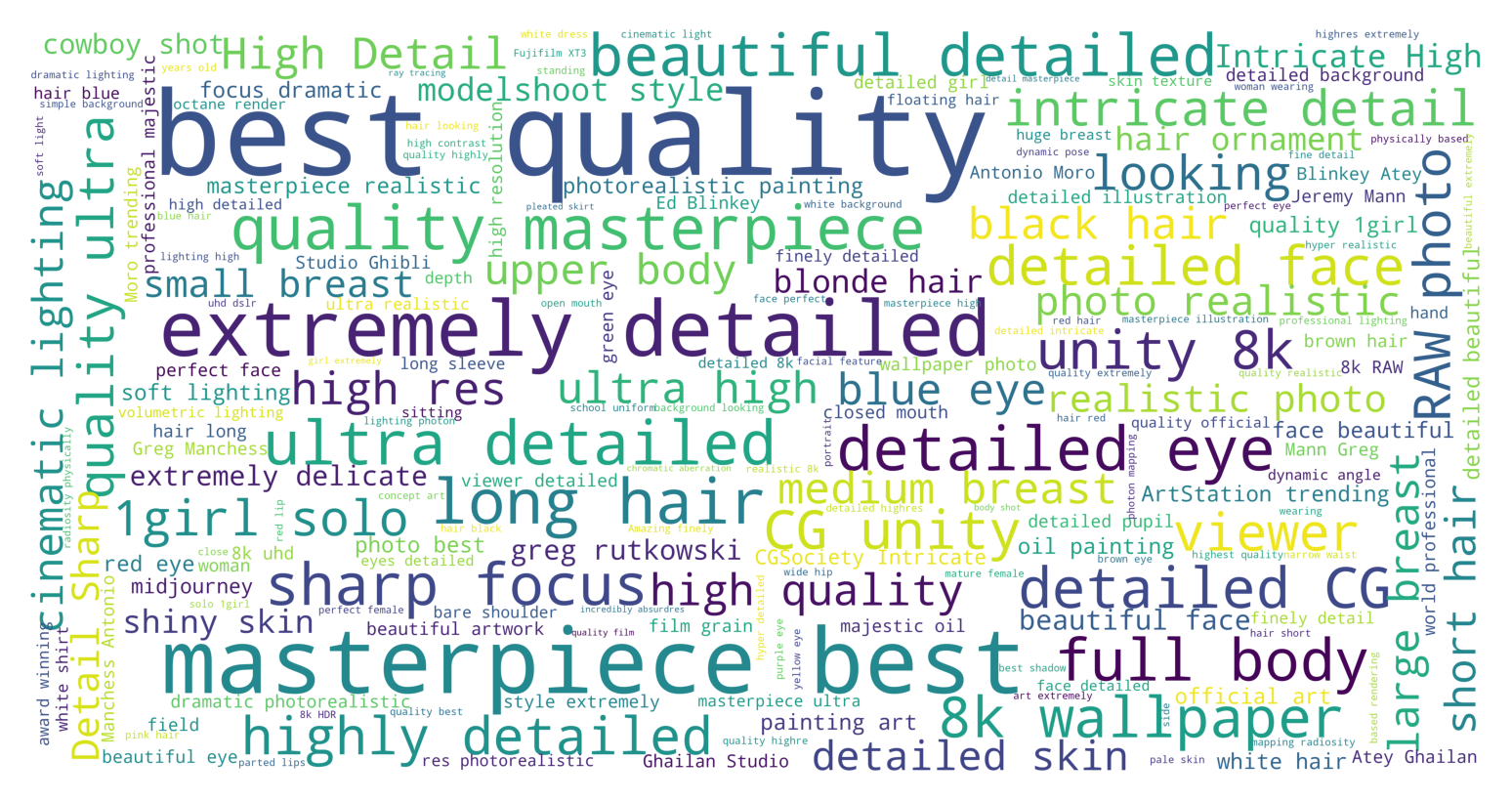}
  }
  \subfigure[Top 20 words in Personalized Diffusions' prompts]{
    \includegraphics[width=0.45\linewidth]{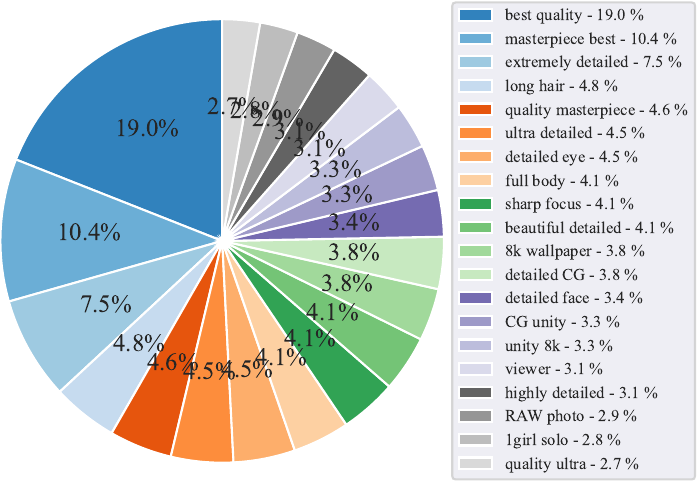}
  }
  \caption{Word clouds of our collected textual prompts that are used for different generative models.}
  \label{fig:wordcloud}
\end{figure}

Figure~\ref{fig:wordcloud} depicts wordclouds showcasing the top 200 most commonly used words for each generative model on the left side, while the right side illustrates the top 20 words used in prompts. Based on our analysis, it is evident that different models display distinct preferences for specific words. However, a consistent observation across all generative models is the prominence of adjectives that describe the quality of the generated images. These adjectives include terms such as `digital art', `highly detailed', `hyper-realistic', and `best quality'.
Despite the abstract and subjective nature of these concepts,
users' behavior indicates that providing generative models with adjectives that directly describe image quality can significantly improve the quality of the image generation results. However, we observe that this mechanism varies across different models. For example, our analysis shows that users of the DALL·E model tend to use fewer quality-related descriptors.

\begin{figure}
  \centering
  \subfigure[Wordcloud of Personalized Diffusions' negative prompts]{
    \includegraphics[width=0.45\linewidth]{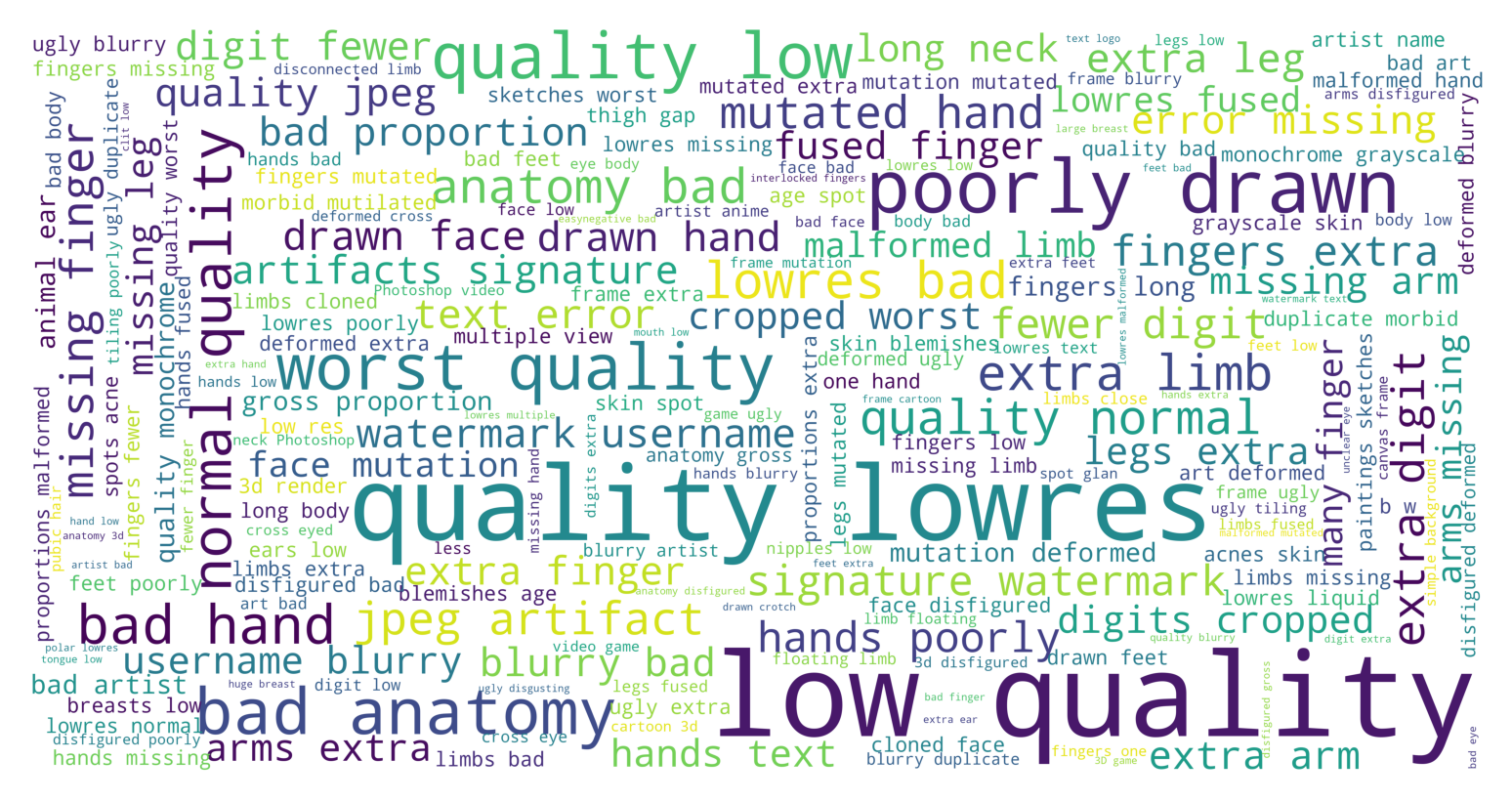}
  }
  \subfigure[Top 20 words in Personalized Diffusions' negative prompts]{
    \includegraphics[width=0.45\linewidth]{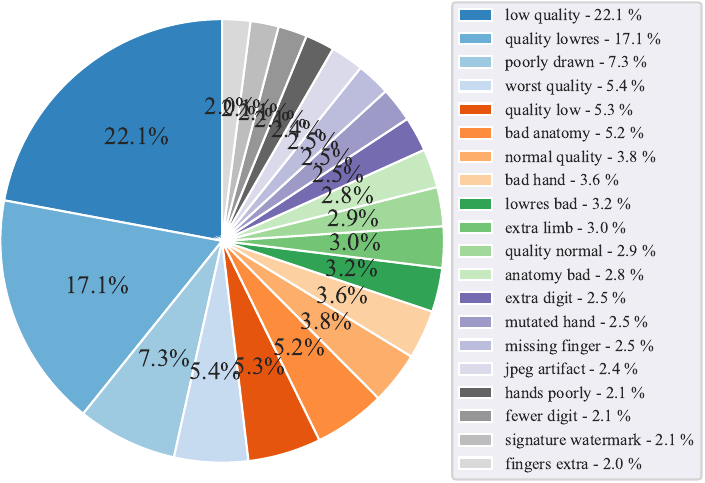}
  }
  \caption{Word clouds of Personalized Diffusions' negative prompts.}
  \label{fig:ngwordcloud}
\end{figure}

Furthermore, for personalized diffusions, users also use negative prompts to steer models away from certain concepts~\cite{Automatic11112022}. 
Figure~\ref{fig:ngwordcloud} displays the word frequency analysis of negative prompts. We find that these negative prompts typically include adjectives that describe low image quality, such as `low quality', `poorly drawn', and `worst quality'. The high frequency of these words suggests that providing negative descriptors can also remarkably enhance the quality of the generated images.

\subsection{Bias Analysis}

\yabin{ Like any large image dataset scraped from the internet, DFLIP-3K may inherently contain biases related to gender, race, and additional attributes. Conducting a meticulous quantification and mitigation of these biases is an imperative task.
We provide a thorough quantitative data that reflects the bias of the dataset. Given the high degree of freedom in text-to-image generation, the resulting image content is remarkably diverse. Therefore, bias investigation within this domain poses significant challenges. To tackle this, we leverage state-of-the-art (SOTA) techniques, including object detection, image classification, and age-gender detection, to perform statistical analysis of our dataset. These codes have been made publicly available on our project page.}

\yabin{Specifically, we use MiVOLO~\cite{kuprashevich2023mivolo} and FairFace~\cite{karkkainenfairface} for age, gender and race estimations on our dataset, which provide invaluable insights into the potential bias. The results are shown in Table.~\ref{tab:gender_distribution} \ref{tab:mivolo_age_results} \ref{tab:fairface_gender_results} \ref{tab:fairface_age_results} \ref{tab:fairface_race_results}.}

\yabin{We acknowledge the disproportion representation of gender and racial groups. A plausible explanation could be that these data are derived from the internet, where current users of generative models exhibit a preference for generating images of certain categories. This discrepancy does not stem from a bias in the data collection methodology, but rather encapsulates the context within which the data was sourced. It is vital for the users of DFLIP-3K to be cognizant of this inherent. We are dedicate to perpetually enhancing the balance and representatives of our datasets in forthcoming iterations. }

\begin{table}[]
\centering
\begin{tabular}{l|r|r|r}
\hline
 & Female & Male & None \\
\hline
    SD & 3,278 & 3,087 & 8,214 \\
    PD & 80,830 & 12,888 & 28,252 \\
    DALL·E 2 & 2,925 & 5,166 & 24,644 \\
    MidJourney & 3,140 & 2,345 & 16,373 \\
    Parti & 9 & 9 & 175 \\
    Imagen & 1 & 8 & 214 \\
\hline
\end{tabular}
\caption{Gender Distribution detected by MiVOLO}
\label{tab:gender_distribution}
\end{table}

\begin{table}[]
\centering
\begin{tabular}{l|r|r|r|r|r|r|r|r|r}
\hline
 & 0-2 & 3-9 & 10-19 & 20-29 & 30-39 & 40-49 & 50-59 & 60-69 & 70+ \\
\hline
SD & 0 & 12 & 237 & 1,554 & 1,547 & 669 & 215 & 97 & 34 \\
PD & 2 & 246 & 12,056 & 58,453 & 19,221 & 2,763 & 585 & 279 & 113 \\
DALL·E 2 & 0 & 29 & 260 & 1,902 & 3,793 & 1,571 & 379 & 116 & 41 \\
MidJourney & 0 & 23 & 293 & 1,414 & 2,178 & 956 & 368 & 193 & 60 \\
Parti & 0 & 0 & 0 & 2 & 11 & 4 & 0 & 1 & 0 \\
Imagen & 0 & 0 & 0 & 5 & 4 & 0 & 0 & 0 & 0 \\
\hline
\end{tabular}
\caption{Age Distribution detected by MiVOLO}
\label{tab:mivolo_age_results}
\end{table}

\begin{table}[]
\centering
\begin{tabular}{l|r|r}
\hline
 & Female & Male \\
\hline
SD & 12,707 & 16,011 \\
PD & 100,495 & 31,641 \\
DALL·E 2 & 7,934 & 11,322 \\
MidJourney & 7,971 & 7,591 \\
Parti & 22 & 27 \\
Imagen & 52 & 24 \\
\hline
\end{tabular}
\caption{Gender Distribution detected by FairFace}
\label{tab:fairface_gender_results}
\end{table}

\begin{table}[]
\centering
\begin{tabular}{l|r|r|r|r|r|r|r|r|r}
\hline
 & 0-2 & 3-9 & 10-19 & 20-29 & 30-39 & 40-49 & 50-59 & 60-69 & 70+ \\
\hline
SD & 459 & 1,342 & 889 & 15,487 & 5,147 & 2,356 & 1,151 & 951 & 936 \\
PD & 1,250 & 9,747 & 10,608 & 91,585 & 9,947 & 3,889 & 2,081 & 1,681 & 1,348 \\
DALL·E 2 & 387 & 890 & 430 & 10,622 & 3,564 & 1,423 & 622 & 572 & 746 \\
MidJourney & 336 & 822 & 305 & 7,612 & 2,726 & 1,025 & 777 & 968 & 991 \\
Parti & 1 & 2 & 1 & 31 & 9 & 2 & 2 & 1 & 1 \\
Imagen & 4 & 3 & 1 & 47 & 10 & 8 & 2 & 1 & 1 \\
\hline
\end{tabular}
\caption{Age Distribution detected by FairFace}
\label{tab:fairface_age_results}
\end{table}

\begin{table}[]
\centering
\begin{tabular}{l|r|r|r|r|r|r|r}
\hline
 & White & Black & East Asian & SEA & ME & Indian & Latino Hispanic \\
\hline
SD & 17,692 & 3,399 & 3,167 & 562 & 2,023 & 1,052 & 823 \\
PD & 79,283 & 6,846 & 33,247 & 1,382 & 6,645 & 2,085 & 2,648 \\
DALL·E 2 & 11,535 & 2,412 & 2,071 & 401 & 1,406 & 860 & 571 \\
MidJourney & 9,191 & 2,010 & 1,755 & 326 & 1,015 & 970 & 295 \\
Parti & 22 & 3 & 14 & 2 & 3 & 2 & 3 \\
Imagen & 42 & 8 & 9 & 1 & 13 & 3 & 0 \\
\hline
\end{tabular}
\caption{Race Distribution detected by FairFace, SEA stands for Southeast Asian and ME stands for Middle Eastern.}
\label{tab:fairface_race_results}
\end{table}

\subsection{Imbalance Problem}

\yabin{The DFLIP-3K exhibits an imbalance, primarily manifesting in the unequal distribution of images across different generative models, as well as in the object categories within the images. This phenomenon is largely attributable to the natural class distribution in the real world, supplemented by potential biases in the image generation process. For instance, images of females far exceed those of males and some models may enjoy wider usage than others. These distributions mirror real-world scenarios. 
The intrinsic imbalance within the DFLIP-3K dataset may cause deepfake detection models exhibit superior performance on the majority class at the expense of the minority class. Users should pay attention to this issue, as it is a common occurrence in the real world.Designing a universal effective deepfake detector is  challenging.}

\subsection{NSFW Filtering Specifics}

\yabin{
In conducting web scraping activities, we strictly adhere to the terms of service provided by each platform. Our scraping solely retrieved content made available by the websites. Given that not all websites guarantee that their data is Safe For Work (SFW), and that the criteria may vary across website providers, these factors are taken into account in our approach.
Consequently, we employed four SOTA safety detectors to filter and flag any potentially inappropriate images. These comprise: LAION-SAFETY~\cite{laionsafety}, GantMAN~\cite{man}, CLIP-based-NSFW-Detector~\cite{clipnsfw} and SD-SAFETY~\cite{SDSAFETY}. As such, we believe the risk of DFLIP-3K resulting in privacy violations or harm is in control.}

\yabin{Detail implementations of these NSFW Detector are presented in their project page, we there give a brief introduction. LAION-SAFETY and GantMAN give multi-label predictions of samples with the following five classes: drawings (safe for work drawings), hentai (hentai and pornographic drawings), neutral (safe for work neutral images), porn (pornographic images, sexual acts) and sexy (sexually explicit images, not pornography). We designate images from the "Drawing" and "Neutral" classes as safe for work (SFW), whereas those from the "Hentai", "Porn", and "Sexy" classes are deemed not safe for work (NSFW). Subsequently, we employ the softmax function to convert the detectors' output into a probability distribution, which is then summed to yield a composite score reflective of the NSFW degree of the input data. The CLIP-based-NSFW-Detector provides a binary score. We set a threshold of 0.5 to determine whether an image is SFW. And the SD-SAFETY model simply flags whether the input is SFW. Notably, we only exclude images identified as NSFW by all detectors, and we provide flags to assist researchers in making judicious decisions when utilizing our dataset. Upon conducting a manual review of a subset of the filtered images, we verified that these models effectively eliminated: overtly sexual imagery (nudity), extreme violence, hate symbols and rhetoric, and illegal content.}

\yabin{The results are shown in Table.~\ref{tab:nsfw1} \ref{tab:nsfw2} \ref{tab:nsfw3} \ref{tab:nsfw4}, and more details and sample-level annotations are provided in the project page.}

\begin{table}[ht]
\centering
\begin{tabular}{l|r|r}
\hline
    & SFW & NSFW \\
\hline
    Stable Diffusion & 12,881 & 2,119 \\
    Personalized Diffusion & 77,845 & 44,122 \\
    DALL·E 2 & 30,068 & 2,667 \\
    MidJourney & 20,787 & 1,071 \\
    Parti & 184 & 9 \\
    Imagen & 213 & 10 \\
\hline
\end{tabular}
\caption{Gantman NSFW Detector Results.}
\label{tab:nsfw1}
\end{table}

\begin{table}[ht]
\centering
\begin{tabular}{l|r|r}
\hline
    & SFW & NSFW \\
\hline
    Stable Diffusion & 14,727 & 273 \\
    Personalized Diffusion & 88,249 & 33,718 \\
    DALL·E 2 & 32,449 & 286 \\
    MidJourney & 21,767 & 91 \\
    Parti & 193 & 0 \\
    Imagen & 223 & 0 \\
\hline
\end{tabular}
\caption{CLIP-based-NSFW-Detector Results.}
\label{tab:nsfw2}
\end{table}

\begin{table}[ht]
\centering
\begin{tabular}{l|r|r}
\hline
    & SFW & NSFW \\
\hline
    Stable Diffusion & 13,927 & 1,073 \\
    Personalized Diffusion & 84,575 & 37,392 \\
    DALL·E 2 & 31,820 & 915 \\
    MidJourney & 21,407 & 451 \\
    Parti & 191 & 2 \\
    Imagen & 216 & 7 \\
\hline
\end{tabular}
\caption{LAION-SAFETY Results.}
\label{tab:nsfw3}
\end{table}

\begin{table}[ht]
\centering
\begin{tabular}{l|r|r}
\hline
    & SFW & NSFW \\
\hline
    Stable Diffusion & 12,050 & 2,950 \\
    Personalized Diffusion & 106,222 & 15,745 \\
    DALL·E 2 & 21,497 & 11,238 \\
    MidJourney & 17,765 & 4,093 \\
    Parti & 186 & 37 \\
    Imagen & 157 & 36 \\
\hline
\end{tabular}
\caption{SD-SAFETY Results.}
\label{tab:nsfw4}
\end{table}

\subsection{Misuse Problem}

\yabin{Misuse behaviour, characterized by disregard for inherent biases and imbalance, can result in a skewed affinity of models towards particular inputs. Moreover, the potential presence of Not-Safe-For-Work (NSFW) elements could precipitate unanticipated behaviors in the resultant Prompt Prediction models. In light of this potentiality, it becomes a requisite for researchers to diligently scrutinize the datasets for imbalances and biases, and to enforce protective measures that ensure the resulting models uphold standards of fairness, safety, and accountability. }

\yabin{To help mitigate the chances of misuse, we plan to release DFLIP-3K under a responsible AI license that prohibits unlawful, unethical, dangerous, or harmful use cases. We will require users to accept this license before granting access to the dataset. We also plan to track dataset downloads and may require submittal of a description of intended usage for auditing purposes. Access will be revoked in cases of terms violation. Furthermore, we offer some tools to facilitate researchers in their usage of DFLIP-3K.
However, we believe the potential positive impacts of releasing DFLIP-3K outweigh the risks, provided appropriate safeguards are in place. The dataset will enable researchers to better detect deepfakes, ultimately fighting their harmful use. We hope DFLIP-3K will encourage development of protective technologies in tandem with generative models.}

\subsection{Personalized Diffusion's Details and Diversity}

\yabin{ The origin of the Personalized Diffusion models lies with individual model creators. People fine-tune Stable-Diffusion on their own private datasets and share the models online.
This allows anyone to access open-source models and utilize them to create deepfakes. Consequently, Personalized Diffusions share the same architecture as Stable Diffusion. Most of these models can be procured from Civitai.com and HuggingFace.co, two of the largest communities for model sharing.
The training data of Personalized Diffusions are private and we get not access.}

\begin{figure}[http]
  \centering
    \includegraphics[width=0.8\linewidth]{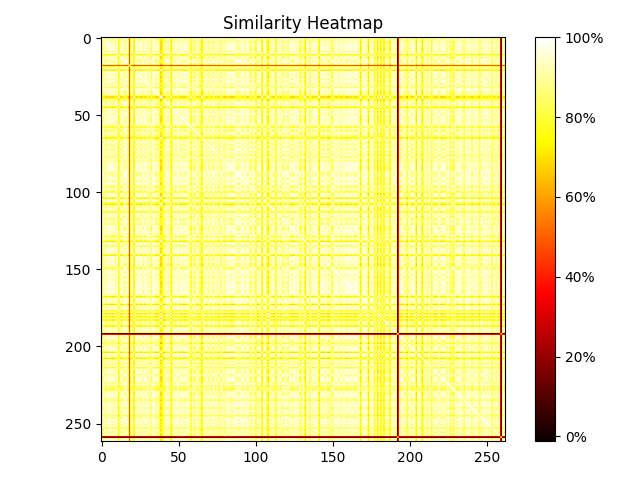}
  \caption{Similarity matrix of Personalized Diffusions.}
  \label{Fig.similarity}
\end{figure}

\yabin{Verifying the diversity of the DFLIP-3K is a challenging task. Directly identifying the model weights has several problems, due to the gradient optimization training process of deep neural networks. Model weights will definitely differ with different initializations and random seeds.
For Stable Diffusion, ASimilarityCalculatior~\cite{ASimilarityCalculatior} provides a more suitable measure. This method compares the self-attention similarities between two models across layers, assessing the overall similarity of the two models' outputs. Since self-attention can accurately reflect the model's internal representations of the input, their similarities can indicate how alike the two models are.
We matched these models pairwise for similarity and plotted a similarity matrix to understand the models' diversity as shown in Figure~\ref{Fig.similarity}. 
Table~\ref{tab:similarity} provides a statistical overview of similarity data across various models.
Due to the high computational demands of calculating pairwise similarity (approximately 1 million calculations for 2,000 models), we randomly selected a subset of 260 models from the full set of Personalized Diffusions.
The results indicate a certain level of similarity among these models, but they are not the same model. 
This observation substantiates the conclusion that these models exhibit significant differentiation.}

\begin{table}[h]
\centering
\begin{tabular}{l|llllllllllll}
\hline
Similarity (\%) & <70 & 70-75 & 75-80 & 80-85 & 85-90 & 90-95 & 95-100 \\ \hline
Num.Pair & 873 & 1,550 & 1,731 & 2,207 & 2,869 & 8,892 & 12,683 \\ \hline
\end{tabular}
\caption{Model Similarity Statistics}
\label{tab:similarity}
\end{table}

\subsection{Ground-truth of Prompts and Hyper-parameters}

\yabin{The ground truth of the prompts and generated models are important, and we have ensured their clarity.
Personalized diffusion models use prompts that follow a specific grammatical structure called the Stable Diffusion Webui grammar. This grammar uses symbols to control the model's attention and output. For example, parentheses "()" around words make the model focus more on those words, while brackets "[]" decrease the model's focus on enclosed words. Additional networks like LoRA can be incorporated into the base Stable Diffusion model by specifying "lora:LoRA\_Name" in the prompt. 
Additional networks such as LoRA, Embedding, Hyper-Network, and ControlNet are primarily designed to incorporate new features, semantic layouts, and poses into the base model. However, the content of the generated images are largely determined by the base models. These auxiliary networks are merely included in the prompts as hyper-parameters, offering guidance on usage and selection of plugins.
We have made available all raw prompts with these supplemental details in the metadata we provide, which may have led to some confusion. For the training of Flamingo, we opted to omit these details as they functioned merely as noise in our implementation.
To make personalized models easier to train, we provide a cleaned up version of the prompts that removes all special symbols and formatting. This simplified prompt format still contains the key phrases and instructions but without the formatting complexity.}

\yabin{The ground truth label that determines whether an image is a deepfake is reliable. Its generative model is determined by the model hash, a technique widely employed in numerous applications including blockchain technology, data storage, and network security.
We identify generative models using the model hash provided by uploaders. Hashing reliably determines if two models are identical, since collisions are extremely unlikely for good algorithms like SHA-256. Different files producing the same hash is theoretically possible but exceedingly rare. With model files as random data, SHA-256 would need $3.4*10^38$ models to get a collision. Erroneous labels are thus highly improbable.
As for watermarks, we only mask out obvious ones, such as those added by DALL·E and Imagen in the lower right corner of images. Indeed, it is possible that some users might add their own watermarks when publishing images, but such instances are rare, at least from our visual inspection perspective. Moreover, Stable Diffusion does not automatically add watermarks. The intention behind our approach is to attain more credible evaluations, not to provide erroneous ground truth.}

\subsection{Hyper-parameter Analysis}

As part of our research, we conduct an extensive analysis of the hyper-parameters used for image generation in personalized diffusions. We focus on personalized diffusions because  the other generative models, such as DALL·E and MidJourney, provide limited information beyond prompts, making it difficult to conduct a detailed analysis of their hyper-parameters.
By quantifying the frequency of each hyper-parameter, our analysis provides valuable insights into which parameters are commonly used by users. This analysis can uncover patterns that potentially contribute to generating improved results.

\begin{figure}[t]
\begin{floatrow}
\ffigbox{
  \centering
  \includegraphics[width=0.37\textwidth]{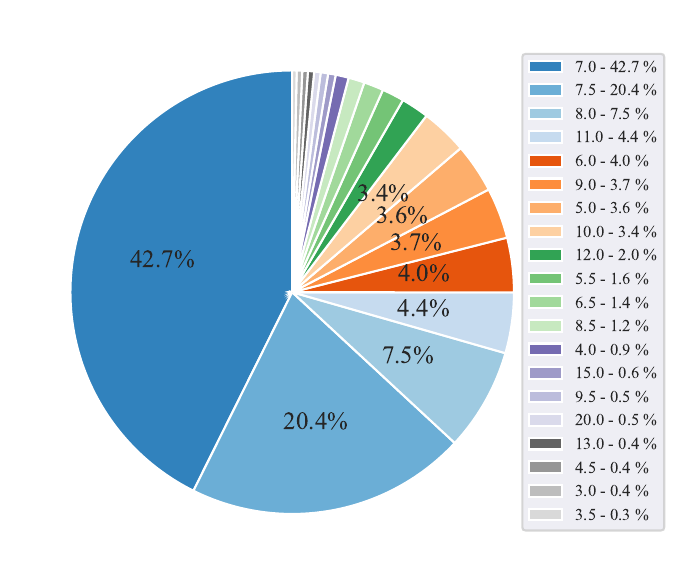}
}{
  \caption{Statistical analysis of the Classifier Free Guidance (CFG) scale used in Personalized Diffusions.}
  \label{fig:CFGscale}
}
\ffigbox{
  \centering
  \includegraphics[width=0.45\textwidth]{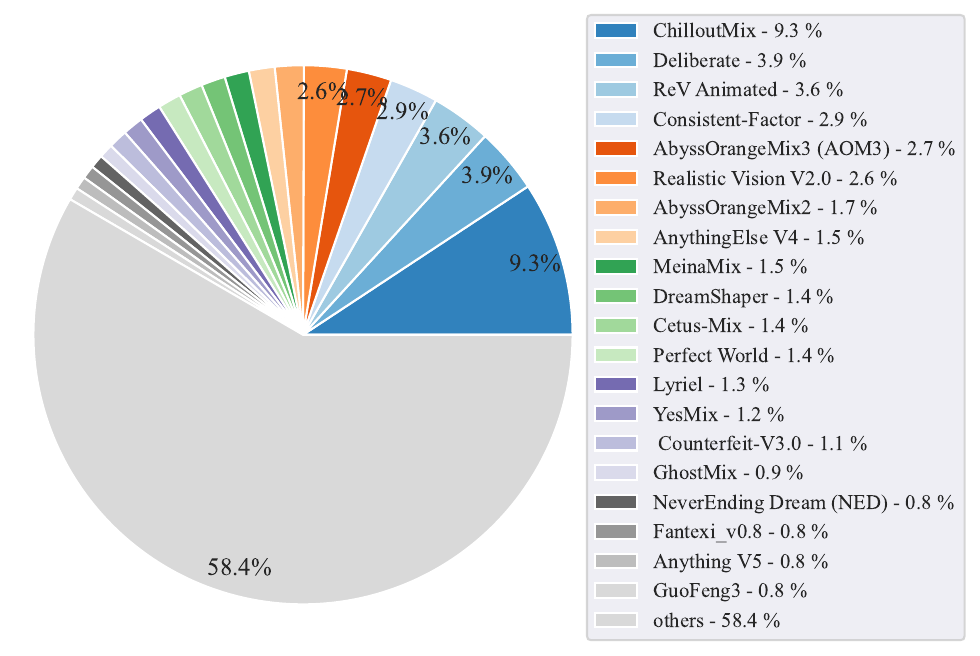}%
}{
  \caption{Statistical analysis of models in Top 20 words in Personalized Diffusions, only displays the names of the top 20 models based on the number of images.}%
  \label{fig:topmodels}
}
\end{floatrow}
\end{figure}

Our initial analysis focuses on the Classifier Free Guidance (CFG) scale, which adjusts the resemblance of the generated image to the input prompt. A higher CFG scale value pormotes closer alignment between the output and the prompt, although it may also introduce distortions in generated image. Figure~\ref{fig:CFGscale} presents the findings of our analysis, revealing that CFG scale values of 7-8 are used in approximately 75\% of cases, with a preference for 7. This finding suggests that 7 is a commonly favored value for this parameter.

\begin{figure}[t]
\begin{floatrow}
\ffigbox{
  \centering
  \includegraphics[width=0.37\textwidth]{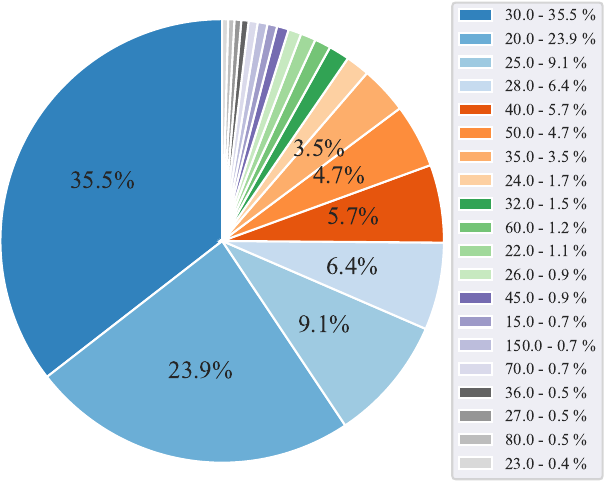}%
}{
  \caption{Statistical analysis of the number of sampling steps used in Personalized Diffusions.}%
  \label{fig:steps}
}
\ffigbox{
  \centering
  \includegraphics[width=0.45\textwidth]{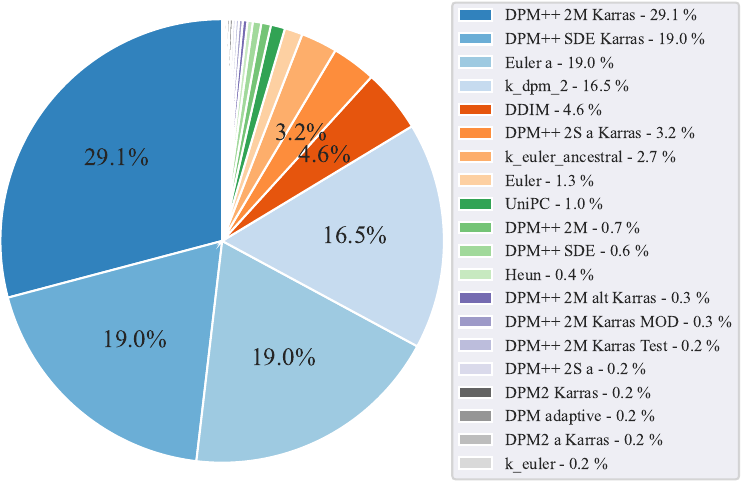}%
}{
  \caption{Statistical analysis of the sampling methods used in Personalized Diffusions.}%
  \label{fig:sampler}
}
\end{floatrow}
\vspace{-0.5cm}
\end{figure}

Our analysis also focused on the hyper-parameter of Sampling Steps, which refers to the number of iterations that Diffusion runs to transform random noise into a recognizable image.  Figure~\ref{fig:steps} displays the results of our analysis, which showed that users most commonly employed 20-30 steps. However, we note that the value of steps is closely related to the Sampling Method employed, which we further analyzed in Figure~\ref{fig:sampler} to identify the most commonly used samplers by users.

In our final analysis, we examine the distribution of Personalized Diffusions models used by users. Figure~\ref{fig:topmodels} displays the proportion of the top 20 models. Our analysis provides valuable insights into the most commonly used models by users in the context of Personalized Diffusions, providing a comprehensive understanding of user preferences in this domain.

\begin{figure*}[t] 
\centering 
\includegraphics[width=0.95\textwidth]{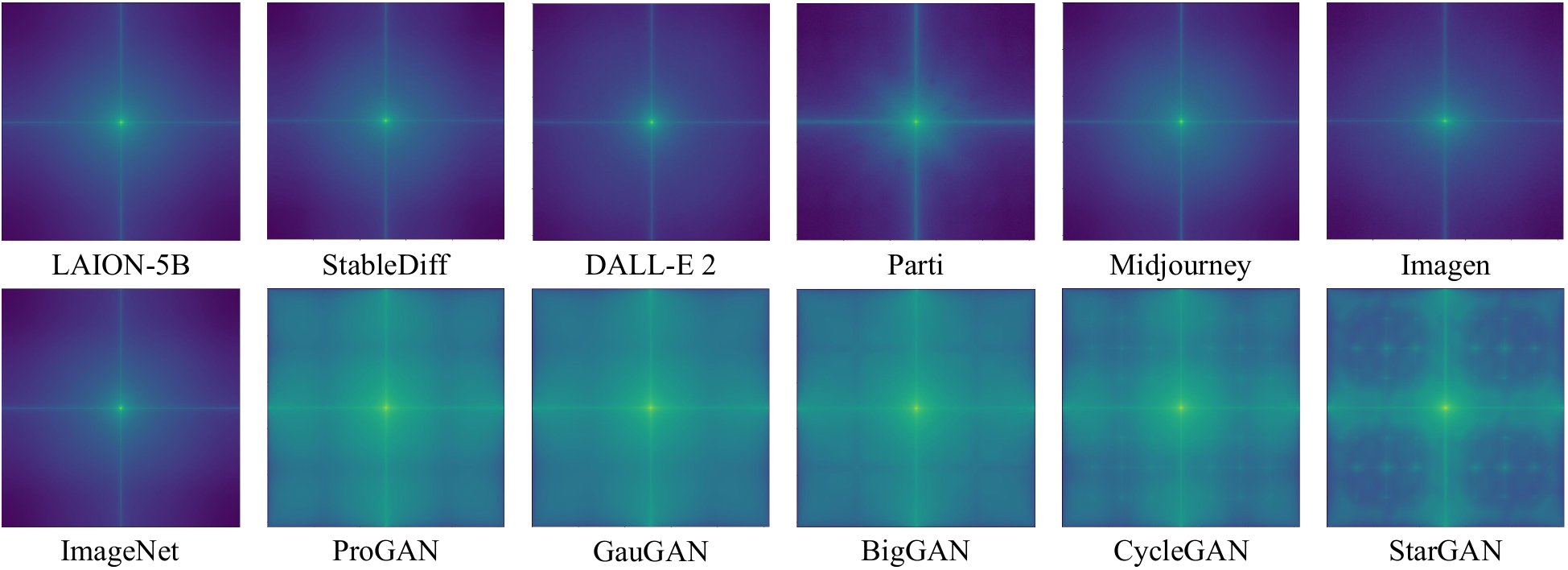} 
\caption{Frequency analysis on each deepfake model generated images. The periodic patterns (dots or lines) of our used deepfake models (Stable Diffusion, DALL·E2, Parti, MidJourney, Imagen) generated deepfake images are highly close to real images from LAION-5B and ImageNet. In contrast, images generated by early deepfake models (ProGAN~\cite{karras2018progressive},GauGAN~\cite{park2019SPADE}, BigGAN~\cite{brock2018large}, CycleGAN~\cite{zhu2017unpaired}, StarGAN~\cite{choi2018stargan}) have different periodic patterns.}
\label{fig:vis1}
\end{figure*}

\subsection{Database Examples}

\begin{figure}
\centering
{\scriptsize
    \def\teaserwid{0.15\linewidth}
    \begin{tabular}{c@{\hspace{.5mm}}*{11}{c@{\hspace{.5mm}}}}
    \raisebox{-0.45\height}{\includegraphics[width=\teaserwid]{Figs/samples/dalle/0000060.jpg}}&  
    \raisebox{-0.45\height}{\includegraphics[width=\teaserwid]{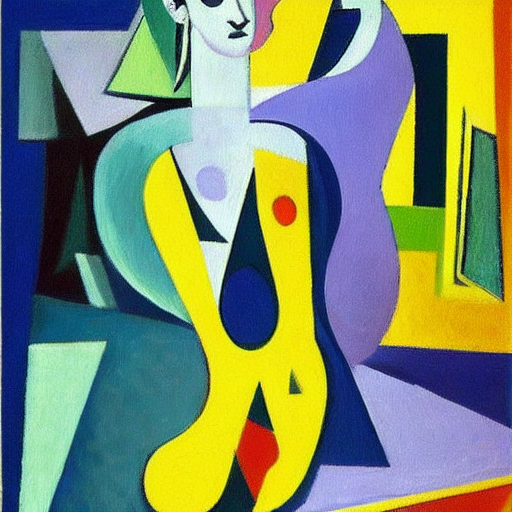}}&
    \raisebox{-0.45\height}{\includegraphics[width=\teaserwid]{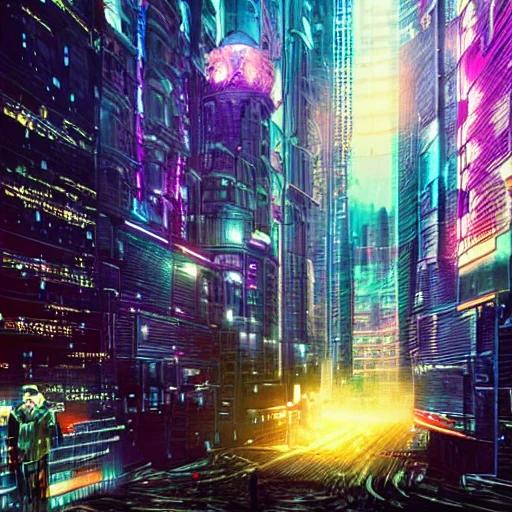}}&
    \raisebox{-0.45\height}{\includegraphics[width=\teaserwid]{Figs/samples/mj/000041.png}}&
    \raisebox{-0.45\height}{\includegraphics[width=\teaserwid]{Figs/samples/imagen/FXfWDS-XkAAr8jc.jpg}}&
    \raisebox{-0.45\height}{\includegraphics[width=\teaserwid]{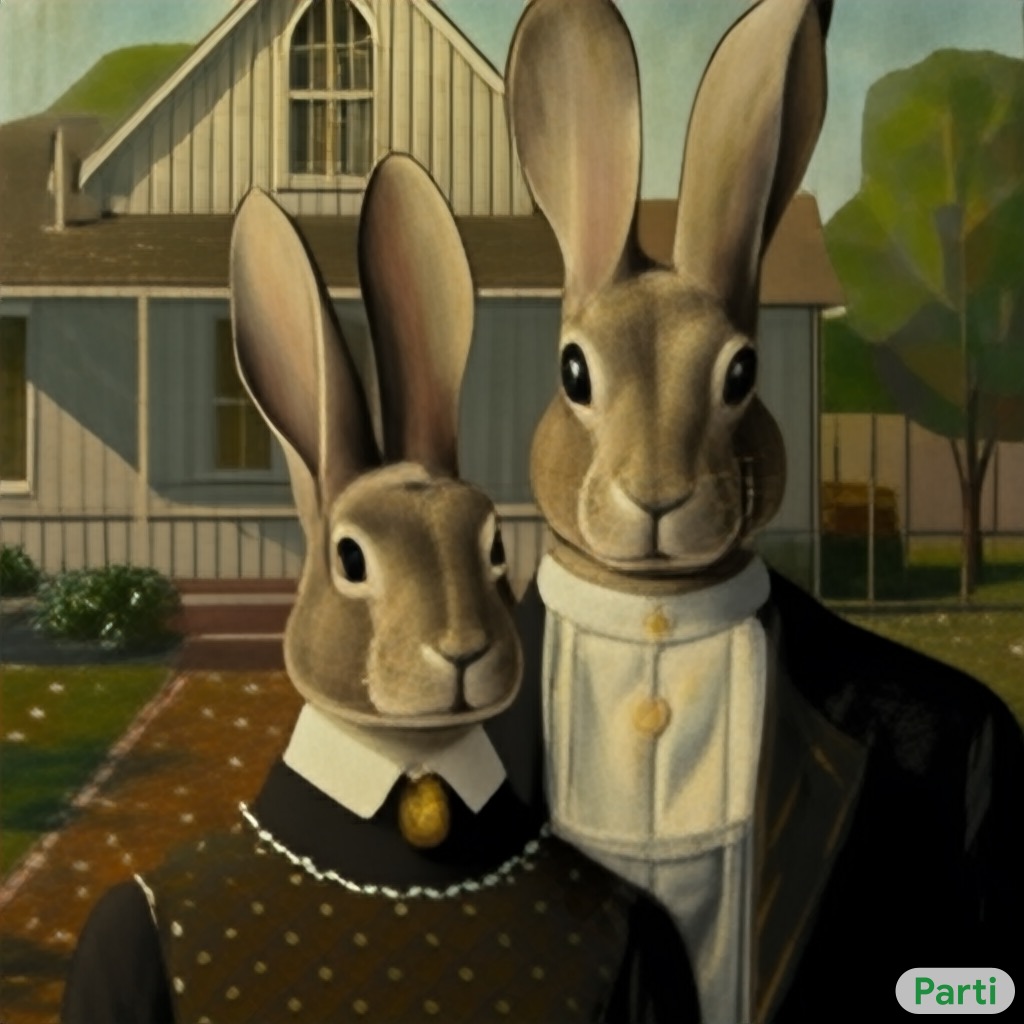}}&
    \vspace{.05in}
    \\
    \raisebox{-0.45\height}{\includegraphics[width=\teaserwid]{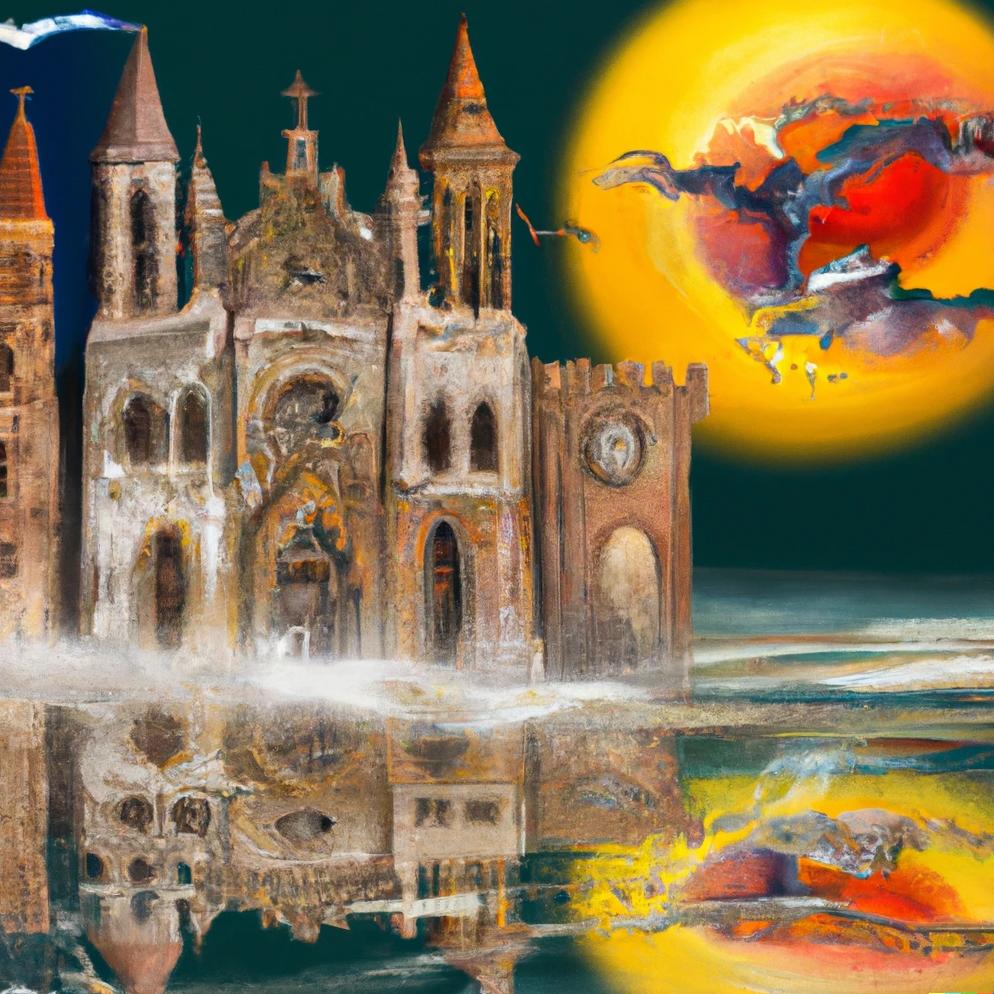}}&  
    \raisebox{-0.45\height}{\includegraphics[width=\teaserwid]{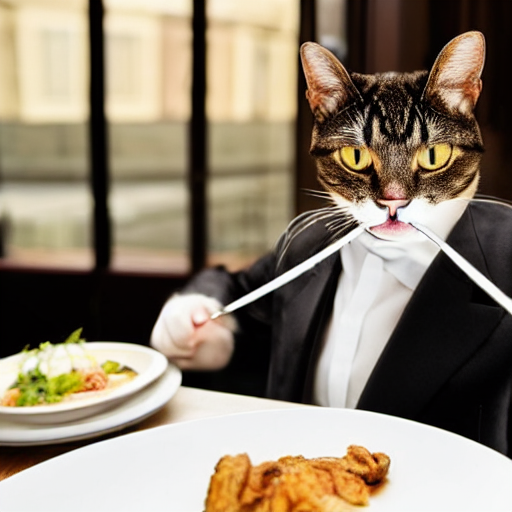}}&
    \raisebox{-0.45\height}{\includegraphics[width=\teaserwid]{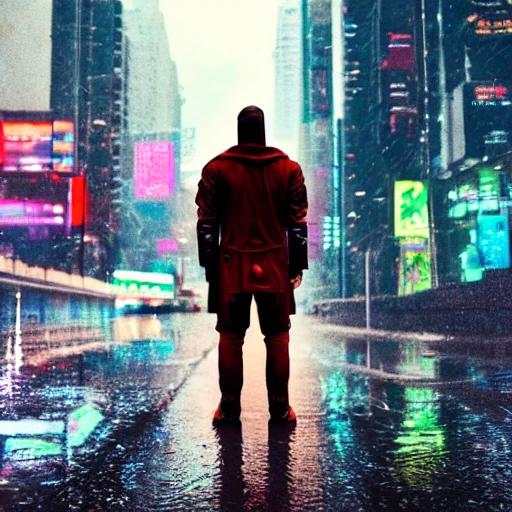}}&
    \raisebox{-0.45\height}{\includegraphics[width=\teaserwid]{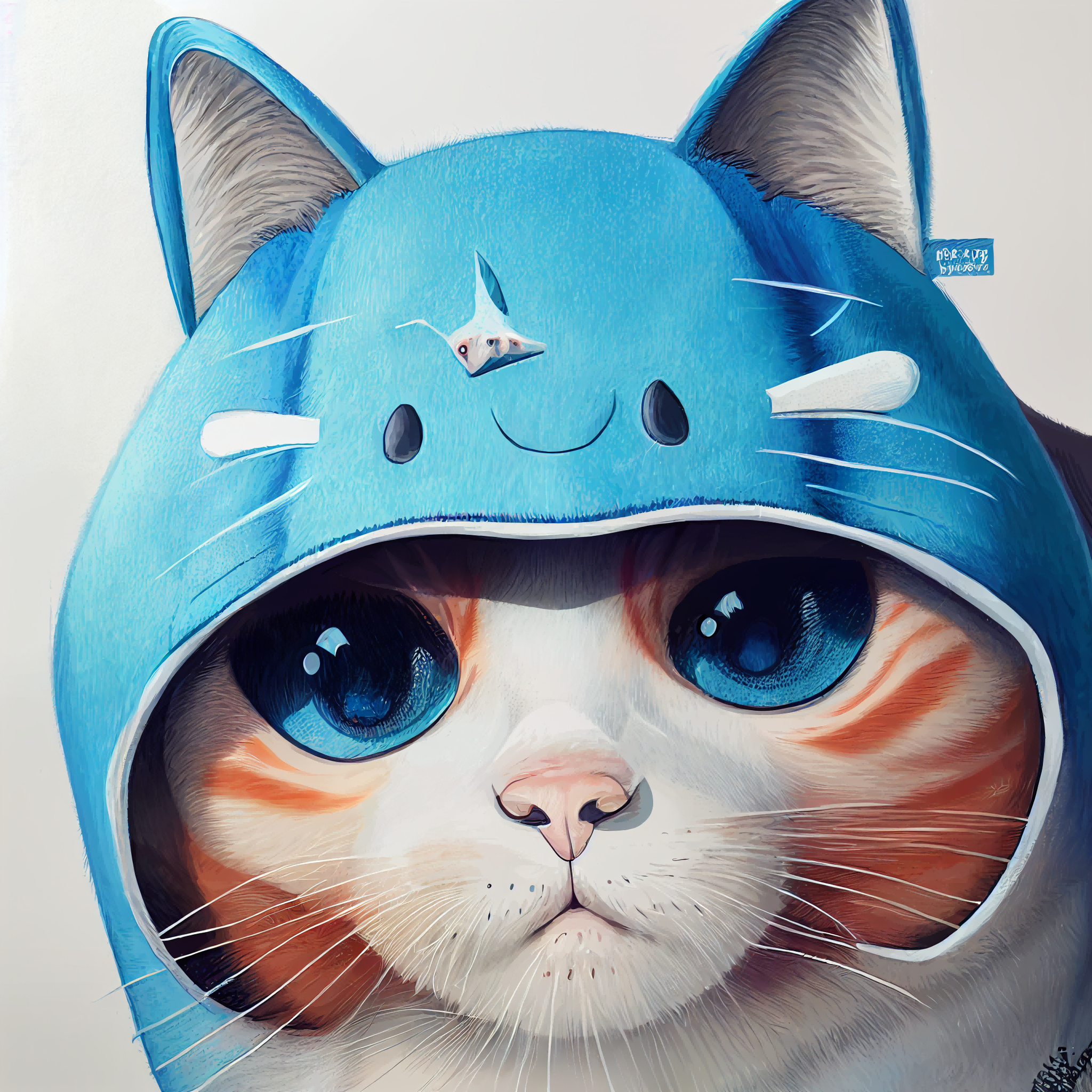}}&
    \raisebox{-0.45\height}{\includegraphics[width=\teaserwid]{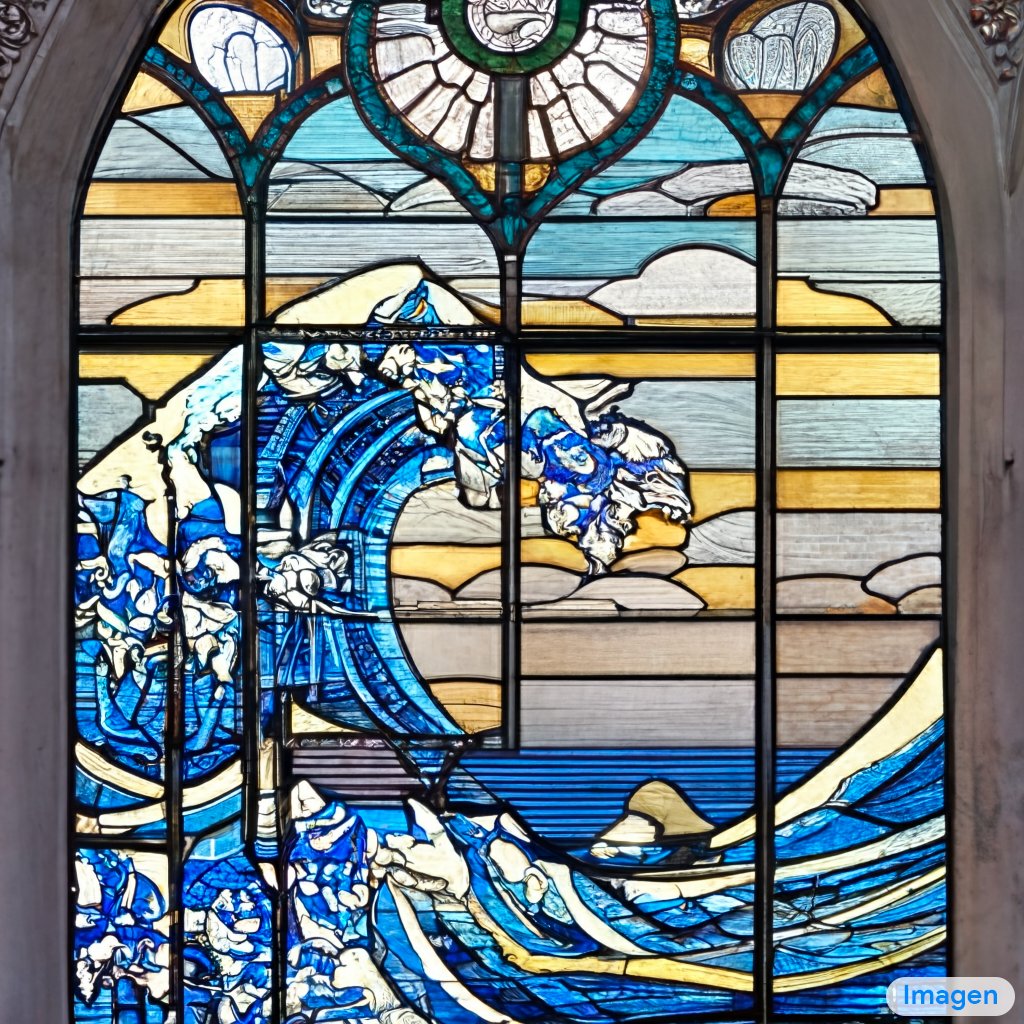}}&
    \raisebox{-0.45\height}{\includegraphics[width=\teaserwid]{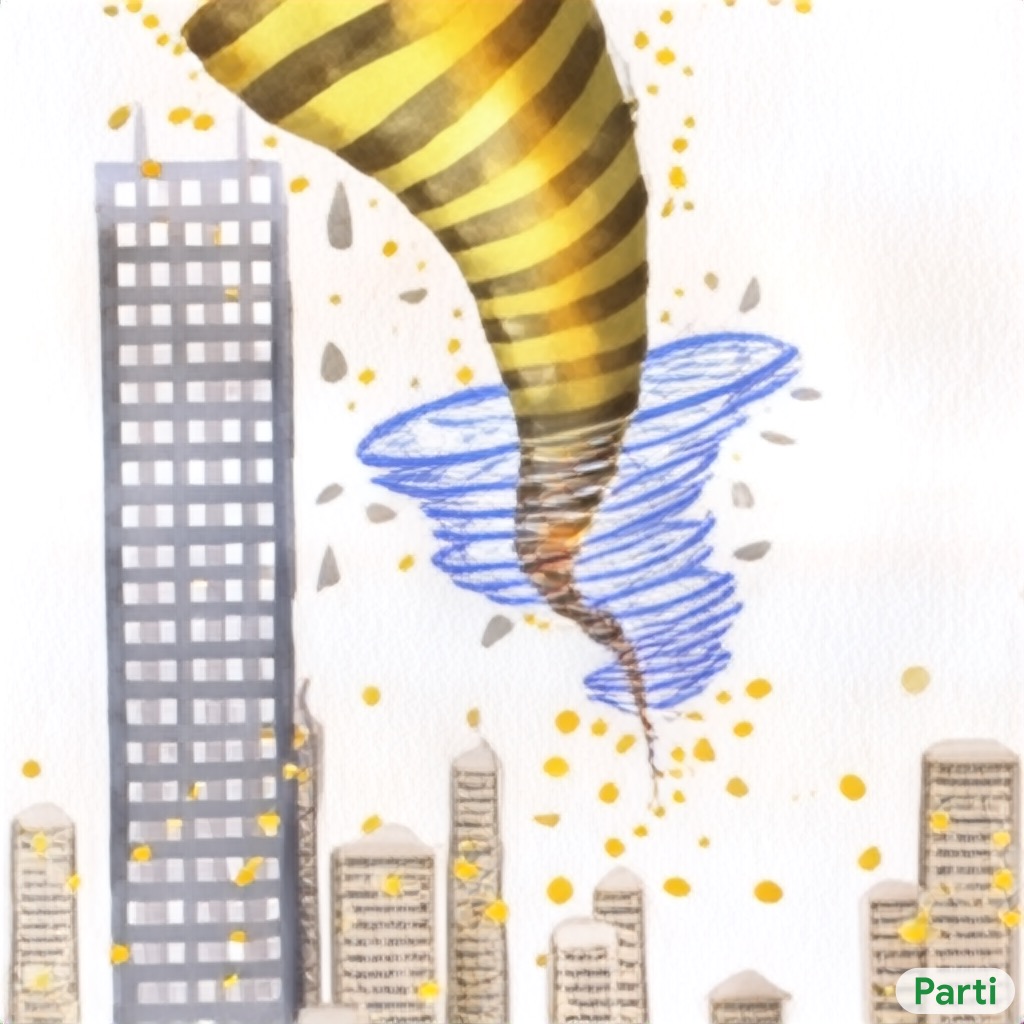}}&
    \vspace{.05in}
    \\
    \raisebox{-0.45\height}{\includegraphics[width=\teaserwid]{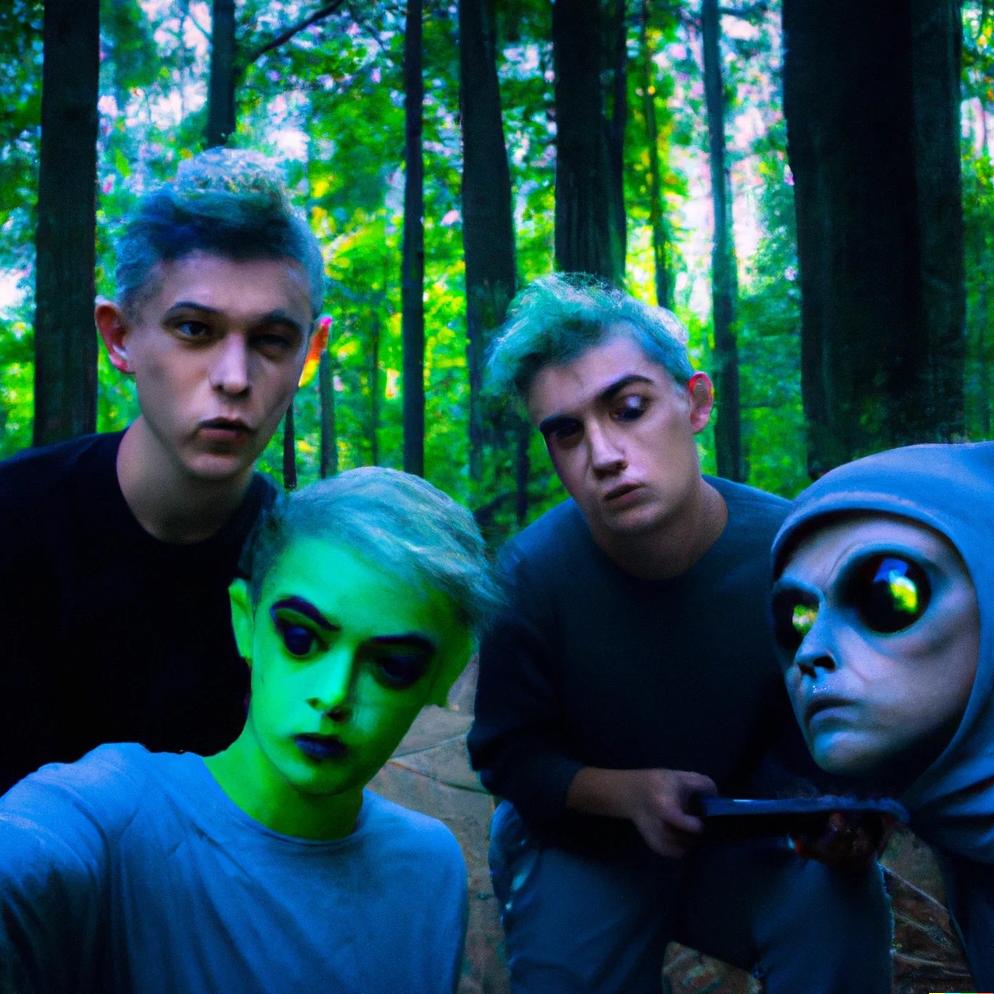}}&  
    \raisebox{-0.45\height}{\includegraphics[width=\teaserwid]{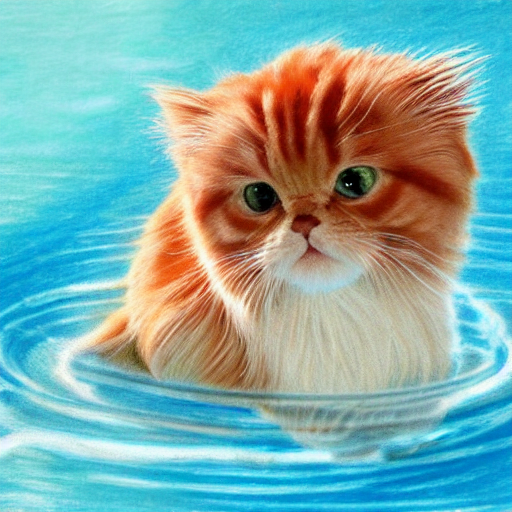}}&
    \raisebox{-0.45\height}{\includegraphics[width=\teaserwid]{Figs/samples/sdft/0017845.jpg}}&
    \raisebox{-0.45\height}{\includegraphics[width=\teaserwid]{Figs/samples/mj/000061.png}}&
    \raisebox{-0.45\height}{\includegraphics[width=\teaserwid]{Figs/samples/imagen/FXkgcDIWYAE6nNo.jpg}}&
    \raisebox{-0.45\height}{\includegraphics[width=\teaserwid]{Figs/samples/parti/beetles_1.jpg}}&
    \vspace{.05in}
    \\
    \raisebox{-0.45\height}{\includegraphics[width=\teaserwid]{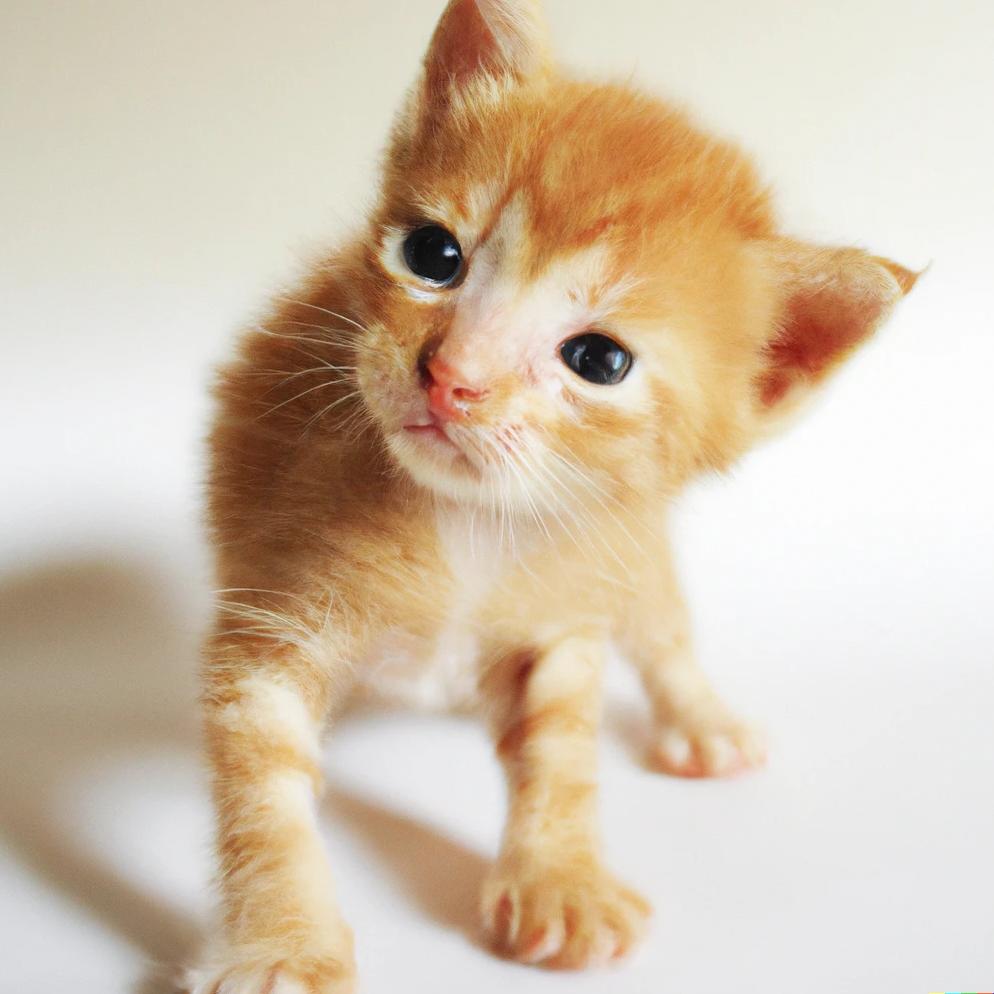}}&  
    \raisebox{-0.45\height}{\includegraphics[width=\teaserwid]{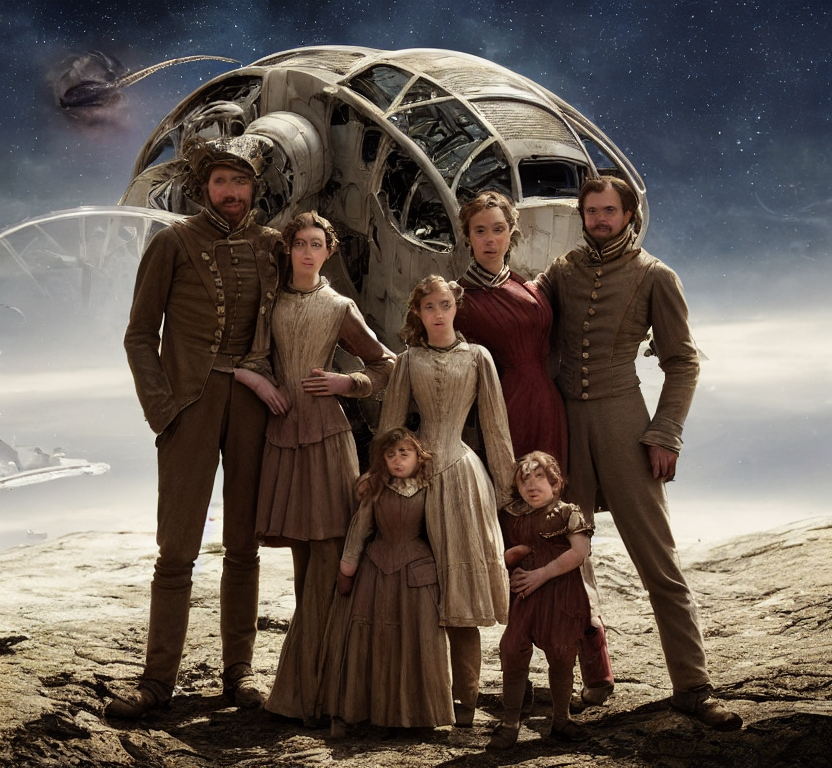}}&
    \raisebox{-0.45\height}{\includegraphics[width=\teaserwid]{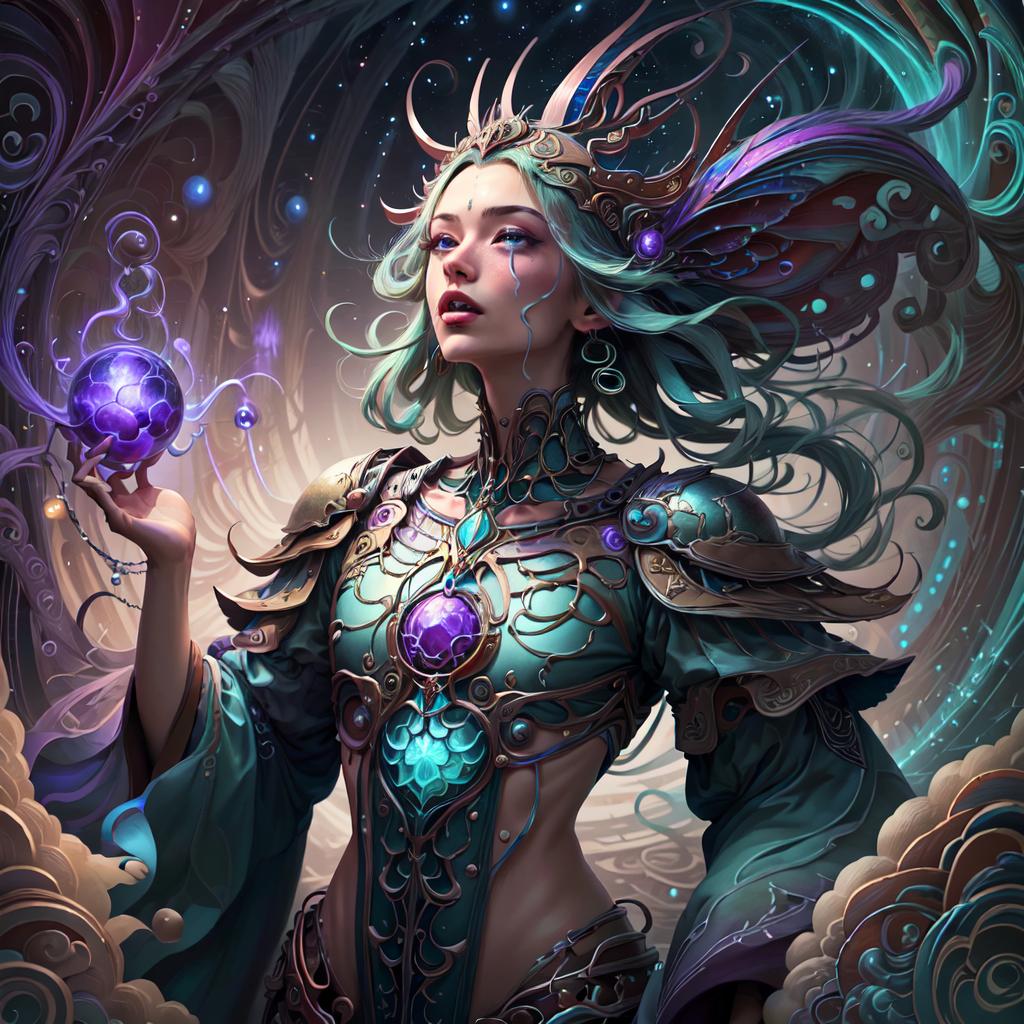}}&
    \raisebox{-0.45\height}{\includegraphics[width=\teaserwid]{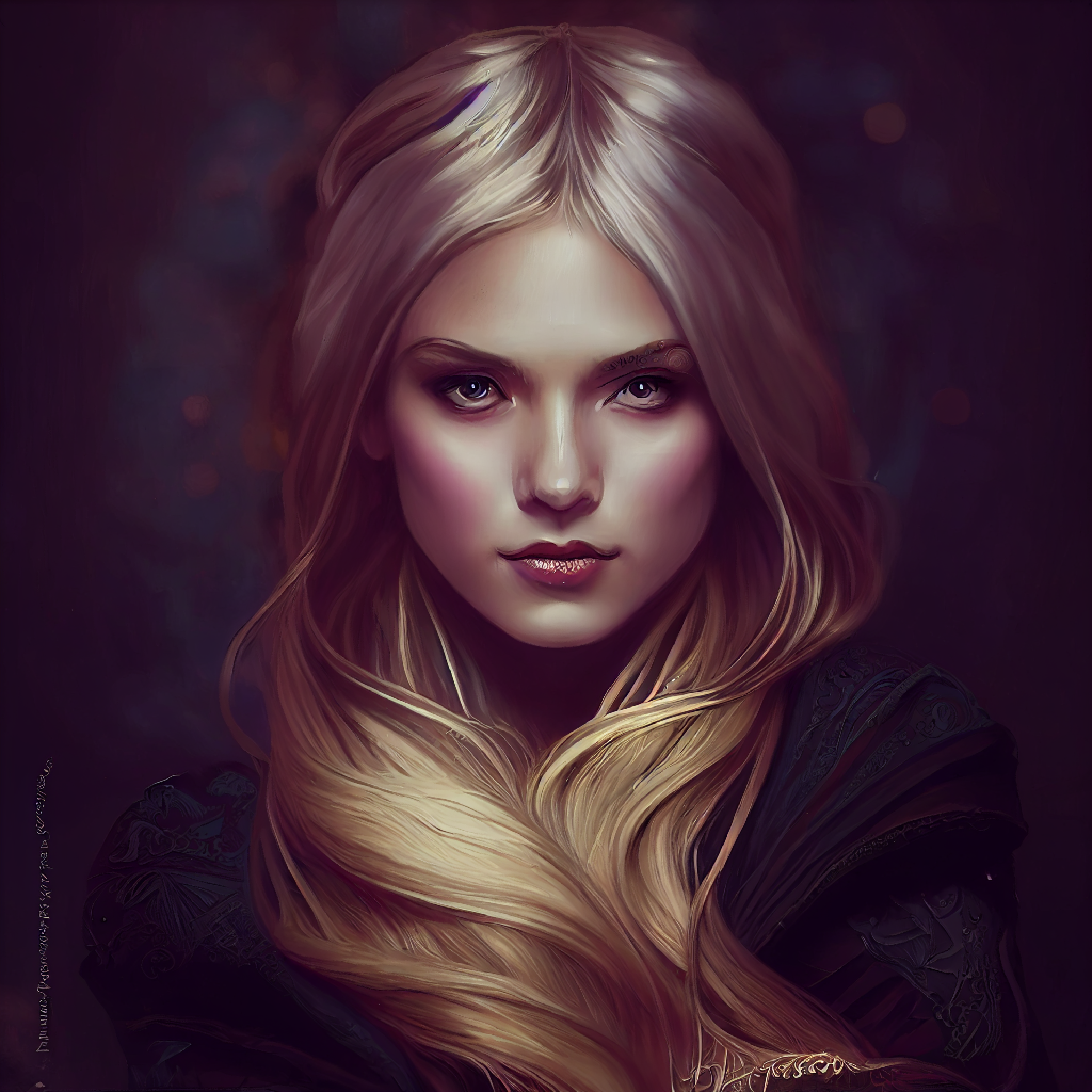}}&
    \raisebox{-0.45\height}{\includegraphics[width=\teaserwid]{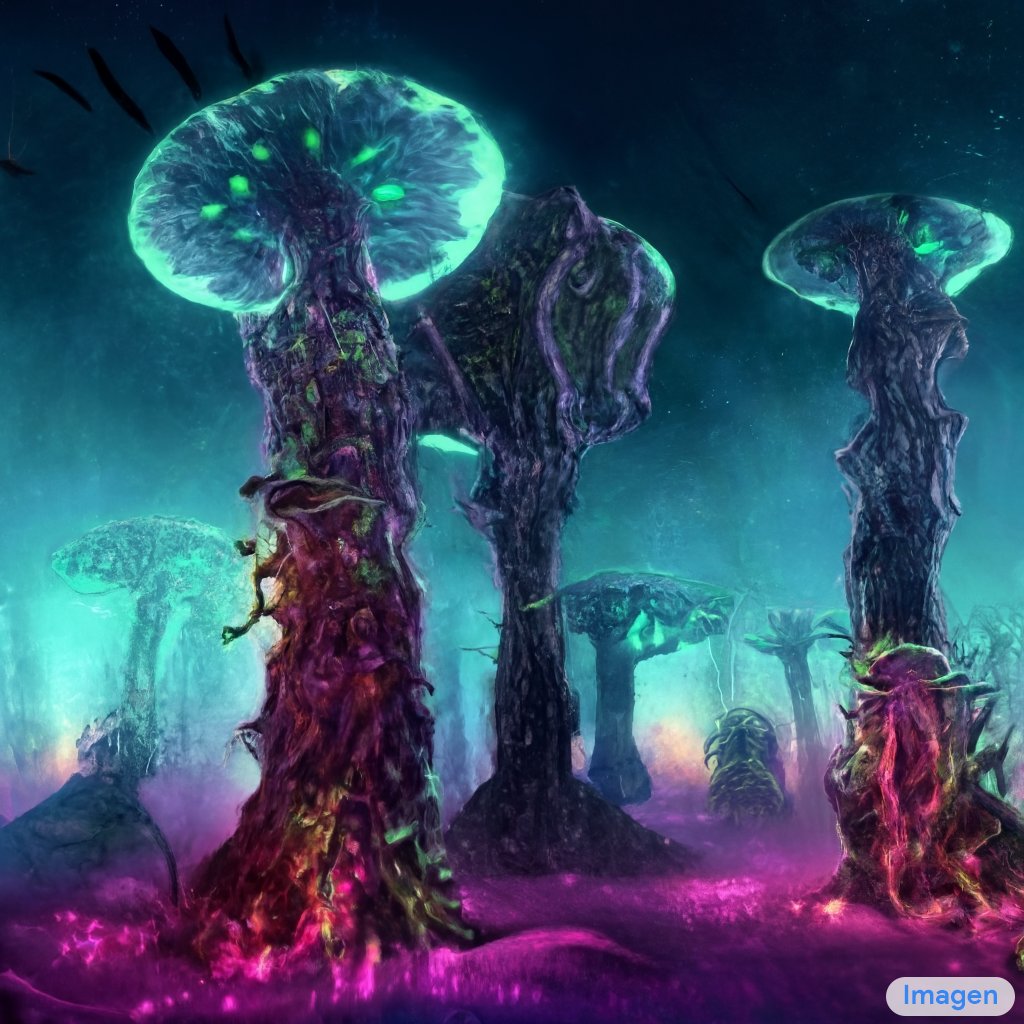}}&
    \raisebox{-0.45\height}{\includegraphics[width=\teaserwid]{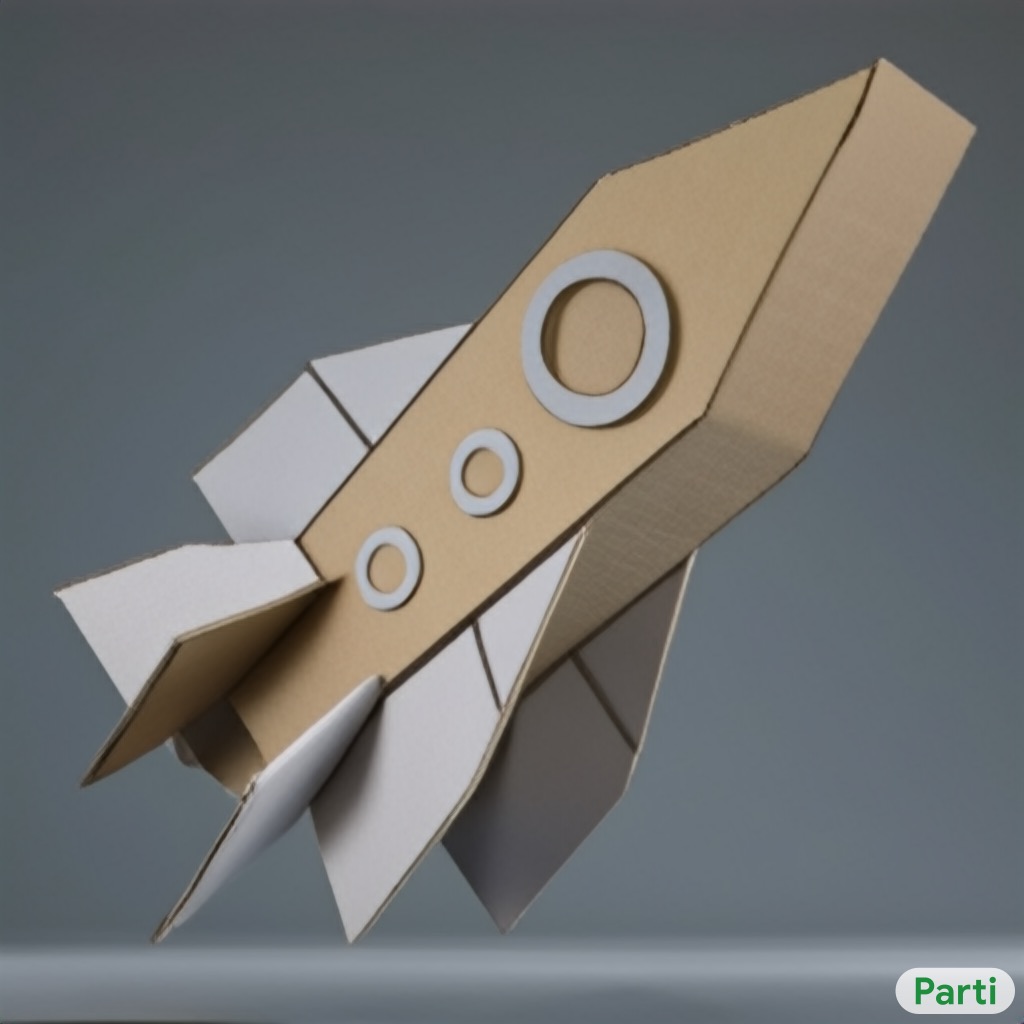}}&
    \vspace{.05in}
    \\
    \raisebox{-0.45\height}{\includegraphics[width=\teaserwid]{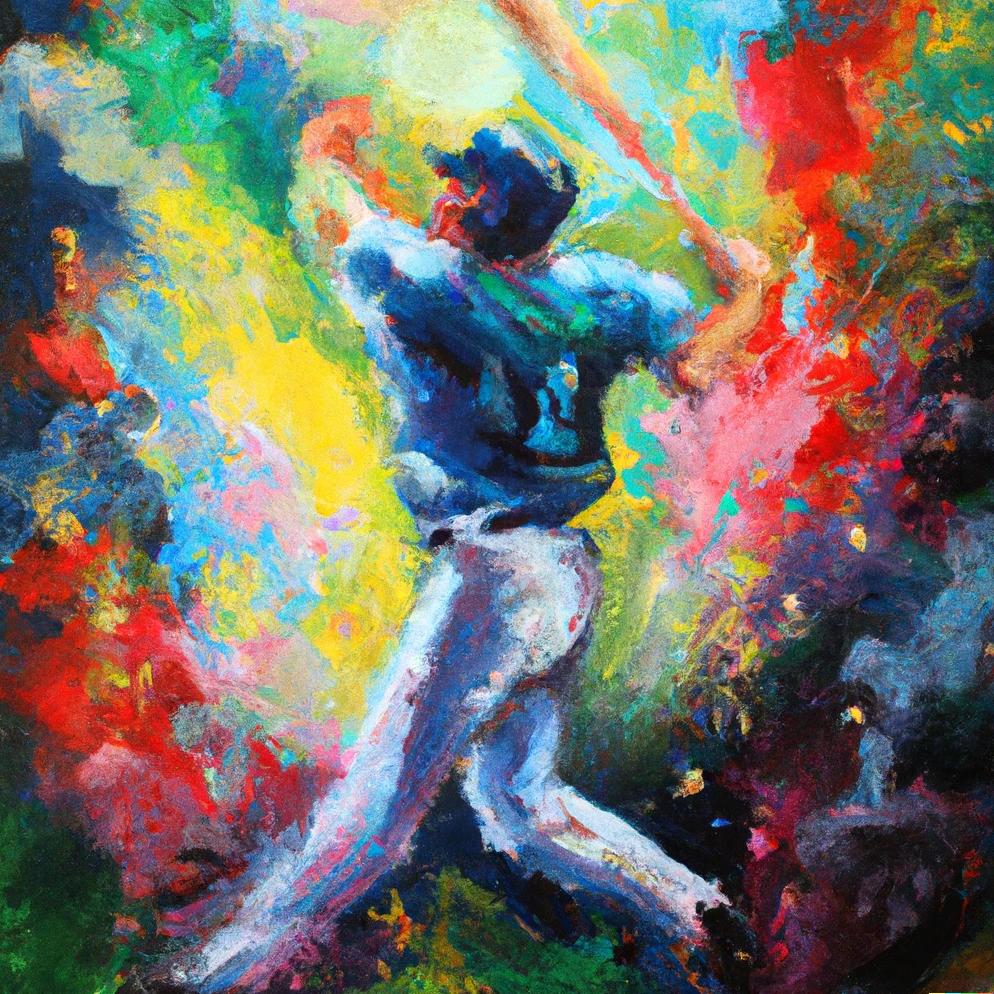}}&  
    \raisebox{-0.45\height}{\includegraphics[width=\teaserwid]{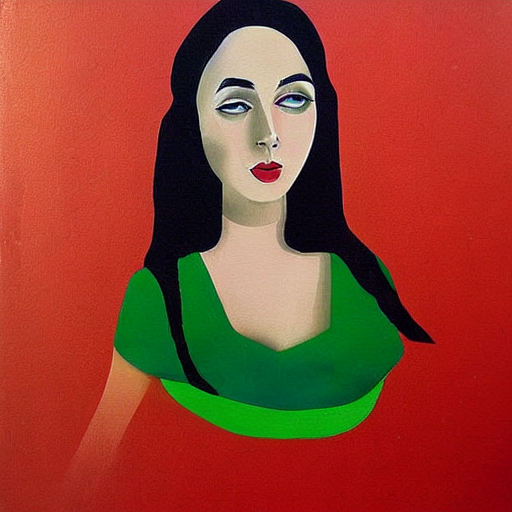}}&
    \raisebox{-0.45\height}{\includegraphics[width=\teaserwid]{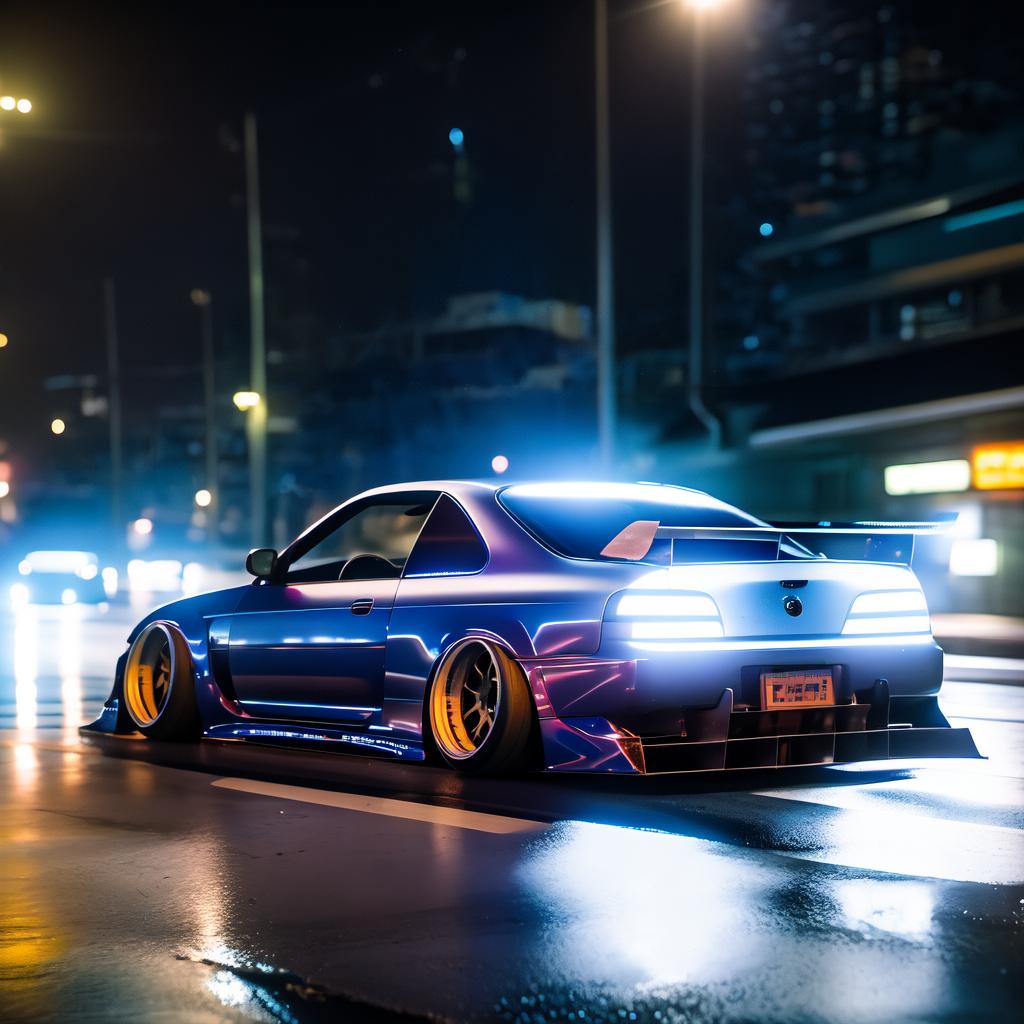}}&
    \raisebox{-0.45\height}{\includegraphics[width=\teaserwid]{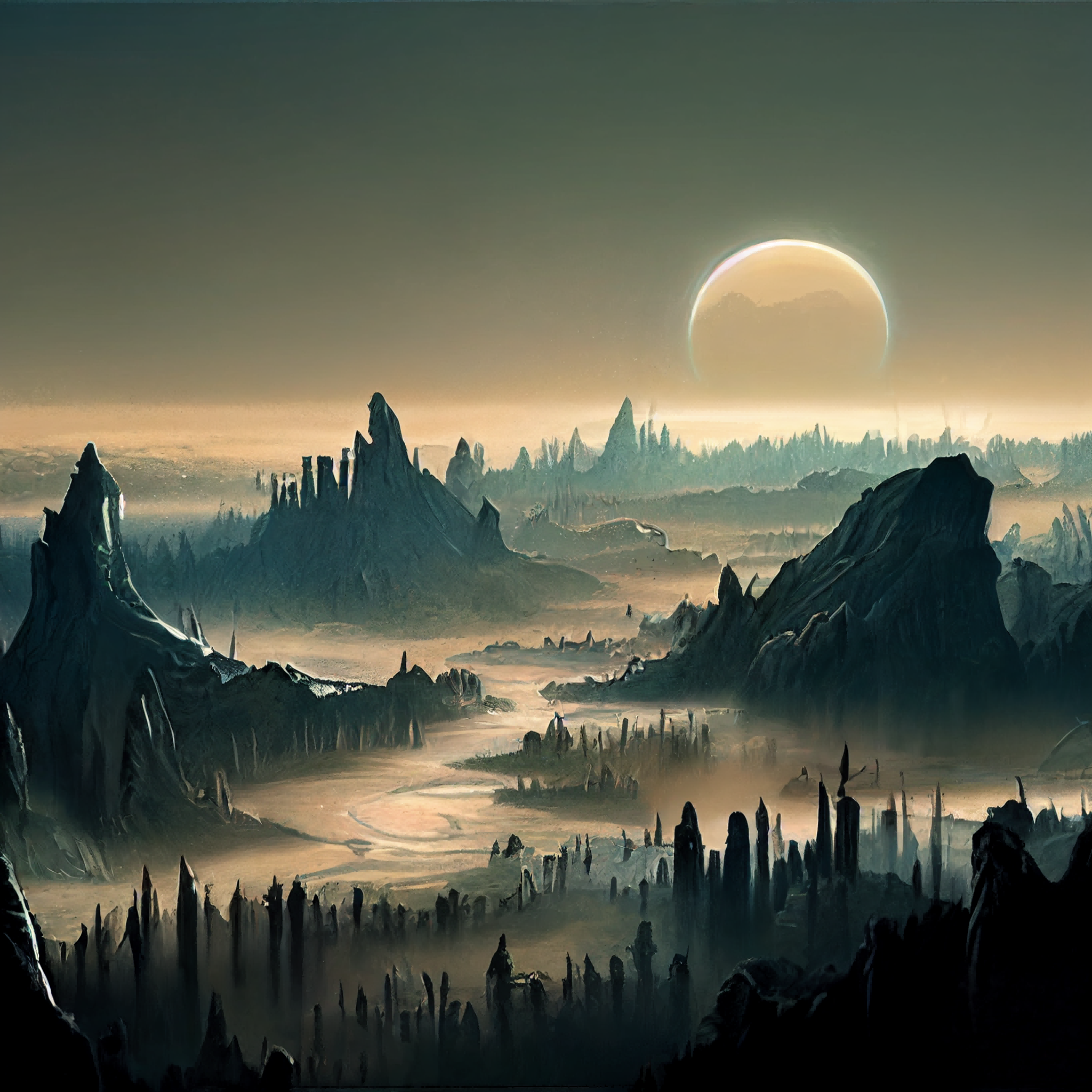}}&
    \raisebox{-0.45\height}{\includegraphics[width=\teaserwid]{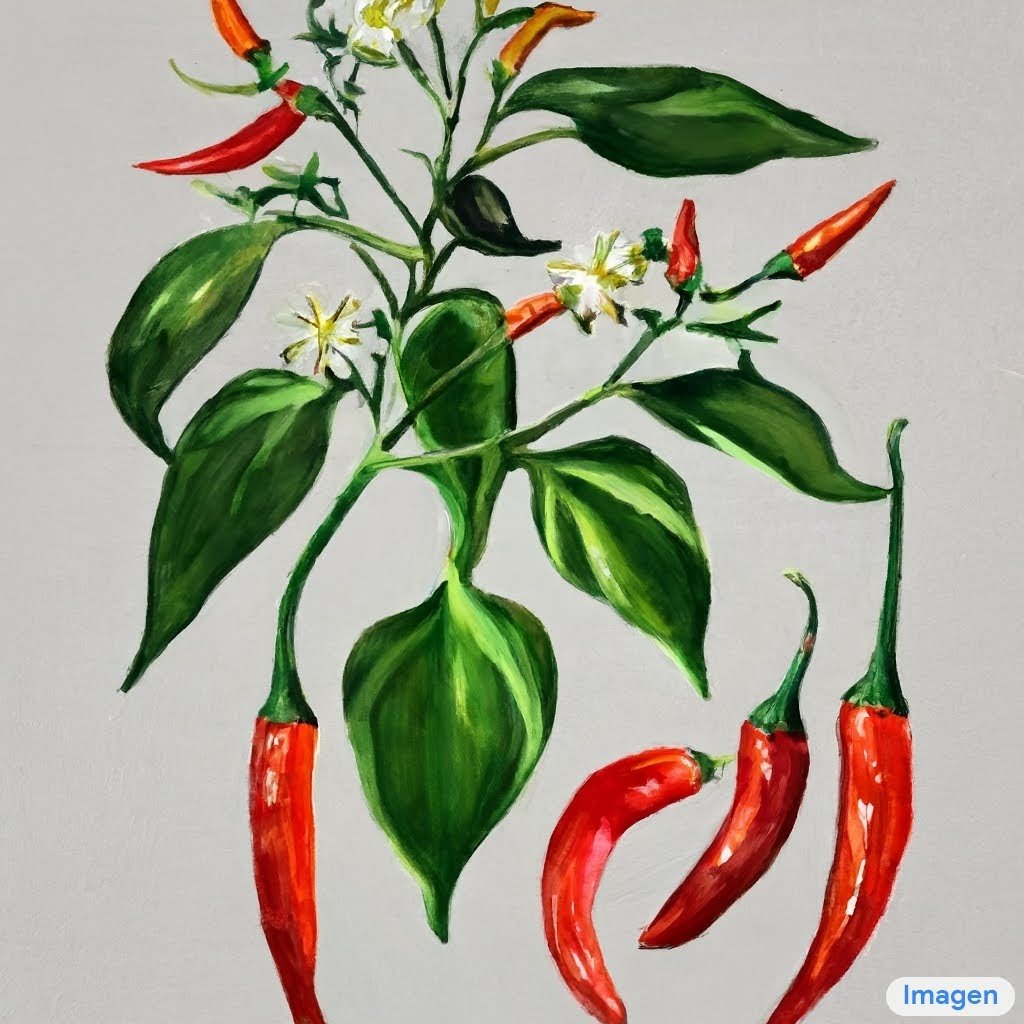}}&
    \raisebox{-0.45\height}{\includegraphics[width=\teaserwid]{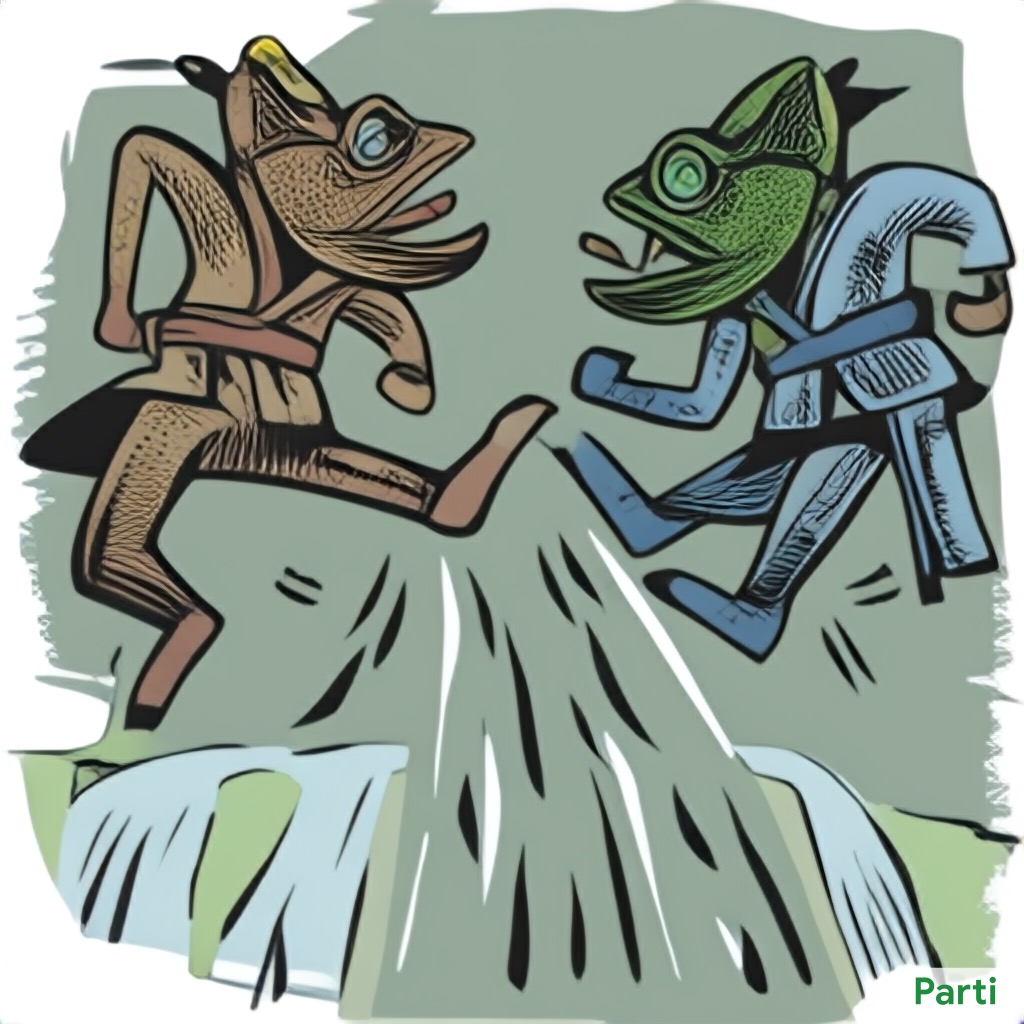}}&
    \vspace{.05in}
    \\
    \raisebox{-0.45\height}{\includegraphics[width=\teaserwid]{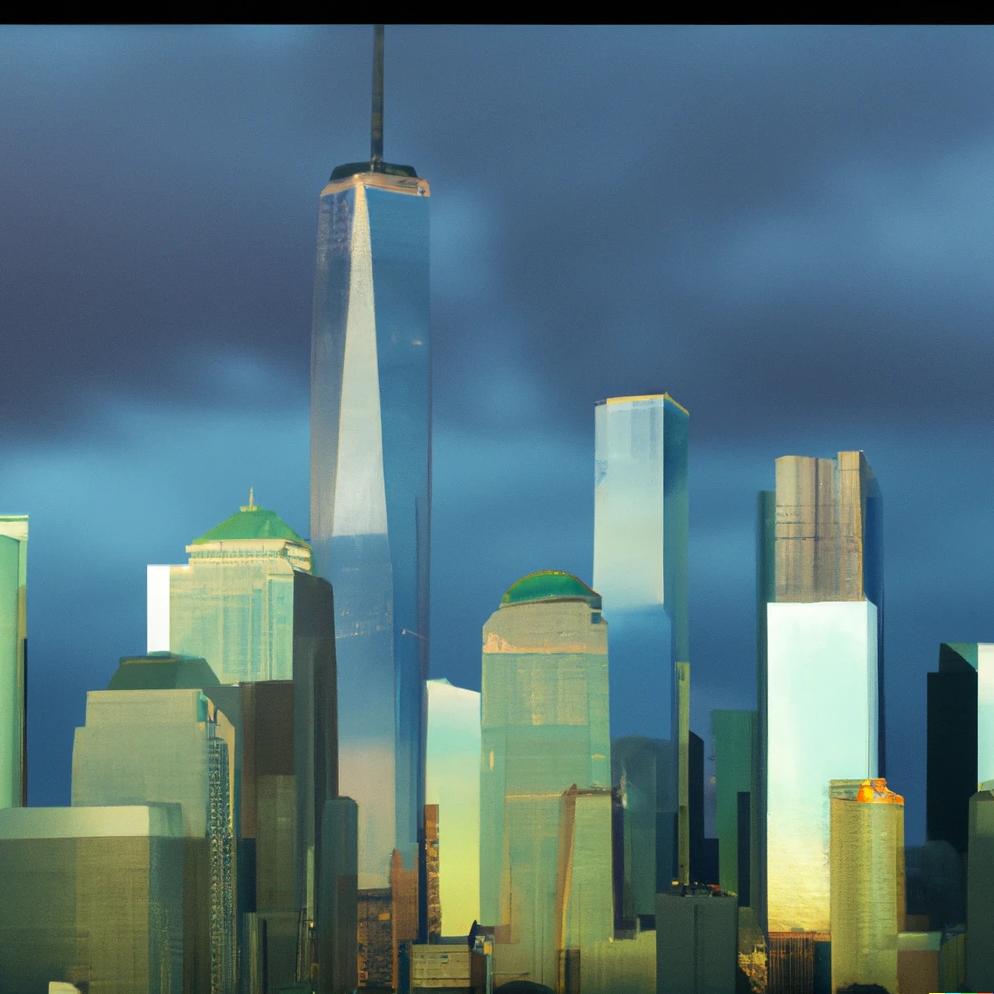}}&  
    \raisebox{-0.45\height}{\includegraphics[width=\teaserwid]{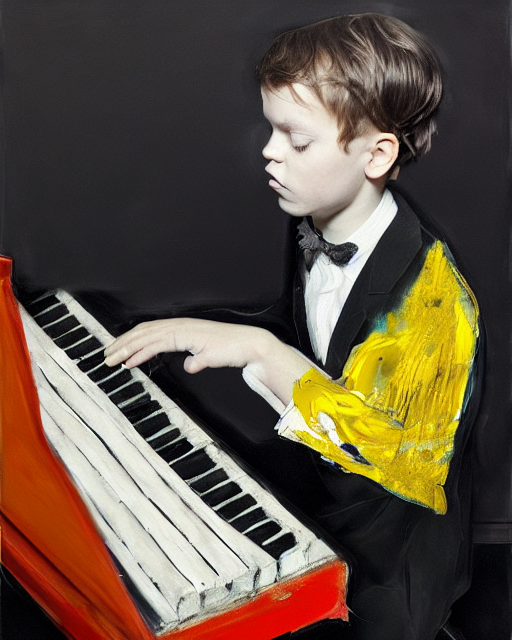}}&
    \raisebox{-0.45\height}{\includegraphics[width=\teaserwid]{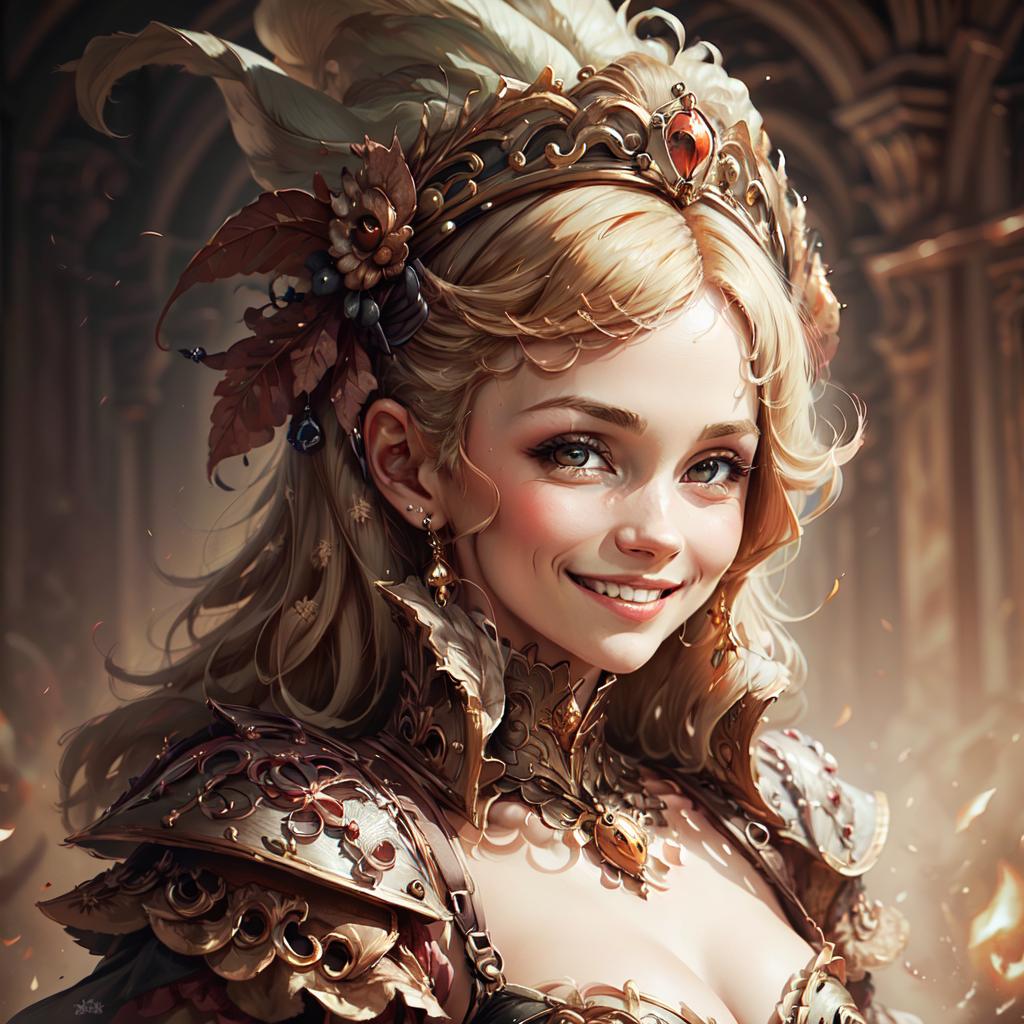}}&
    \raisebox{-0.45\height}{\includegraphics[width=\teaserwid]{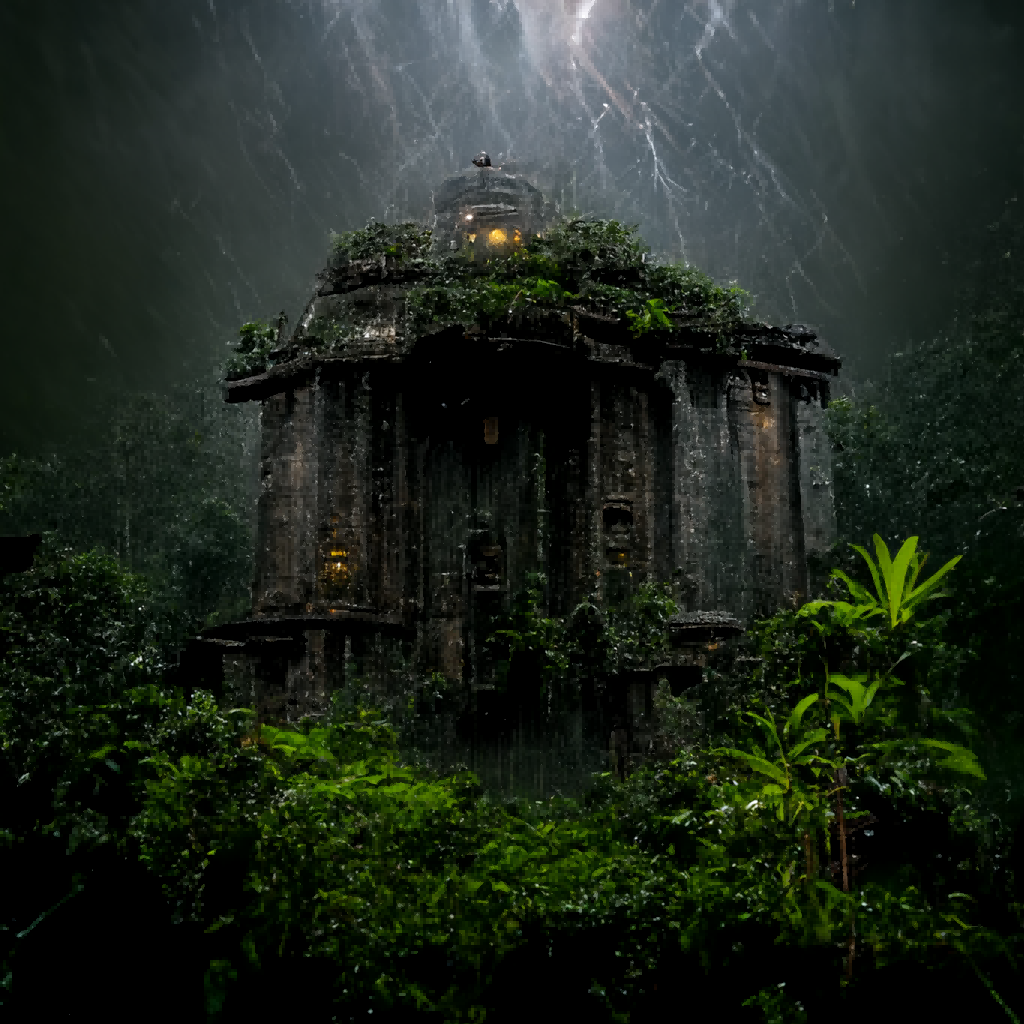}}&
    \raisebox{-0.45\height}{\includegraphics[width=\teaserwid]{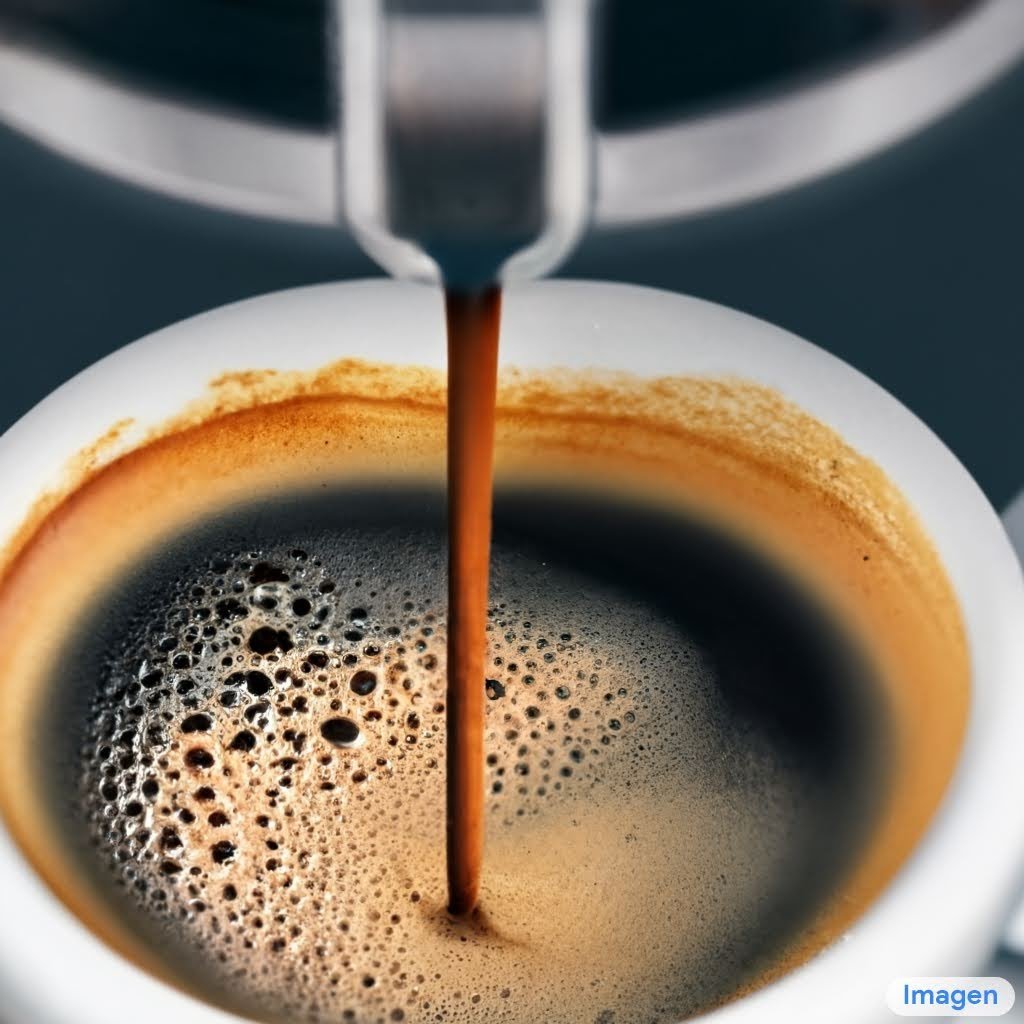}}&
    \raisebox{-0.45\height}{\includegraphics[width=\teaserwid]{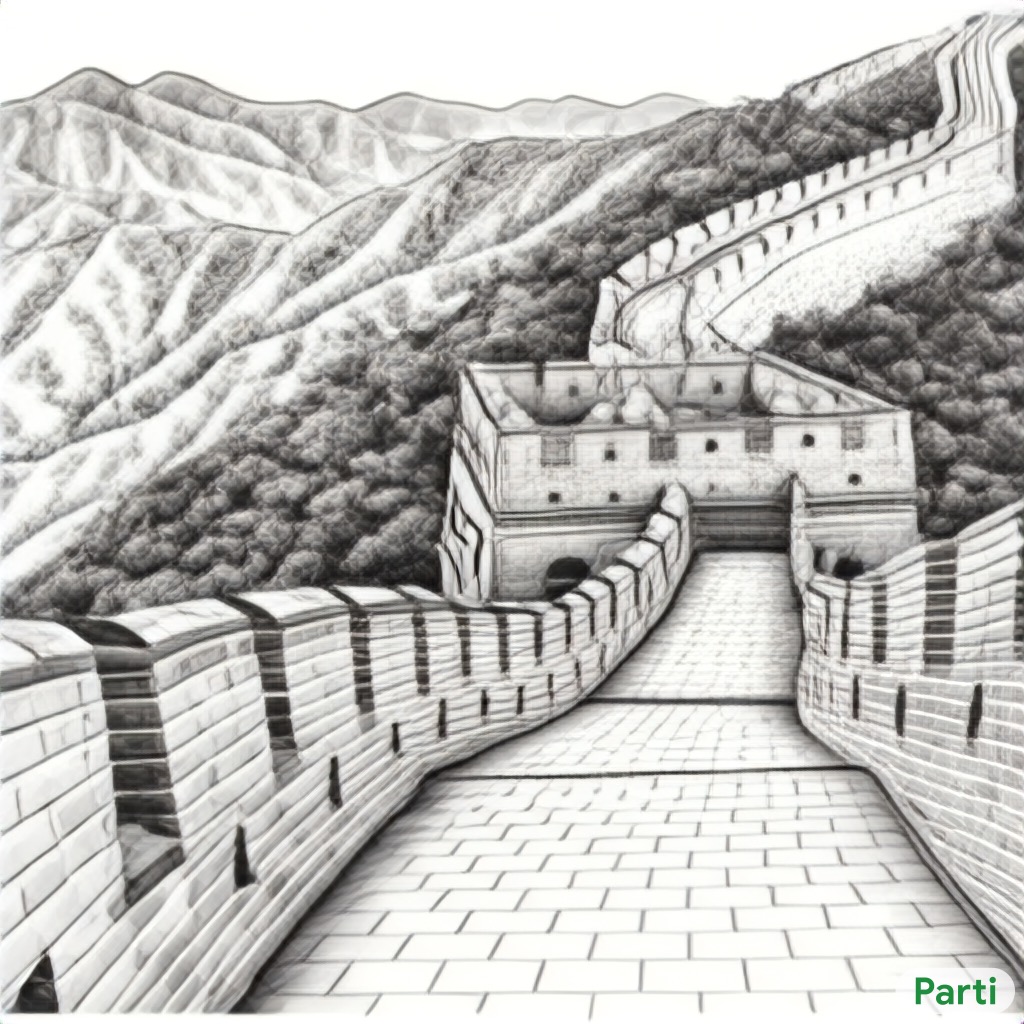}}&
    \vspace{.05in}
    \\
    \raisebox{-0.45\height}{\includegraphics[width=\teaserwid]{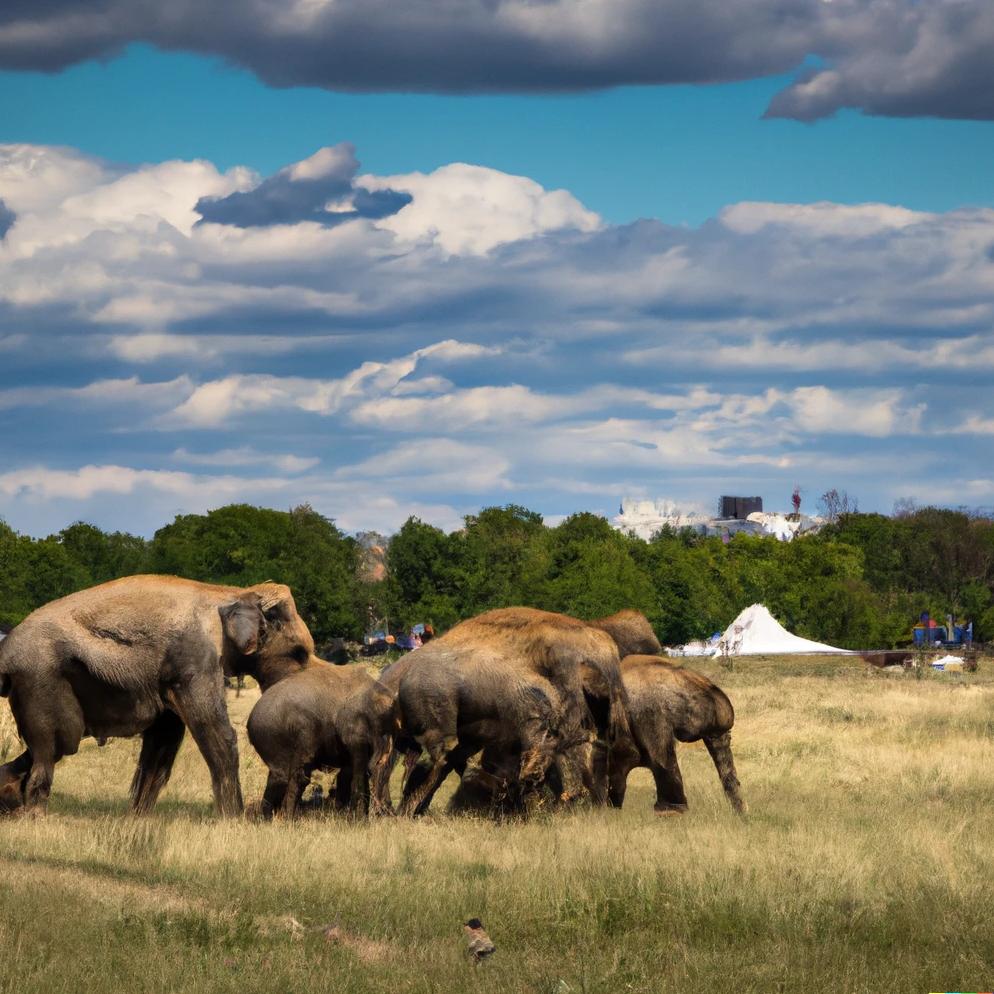}}&  
    \raisebox{-0.45\height}{\includegraphics[width=\teaserwid]{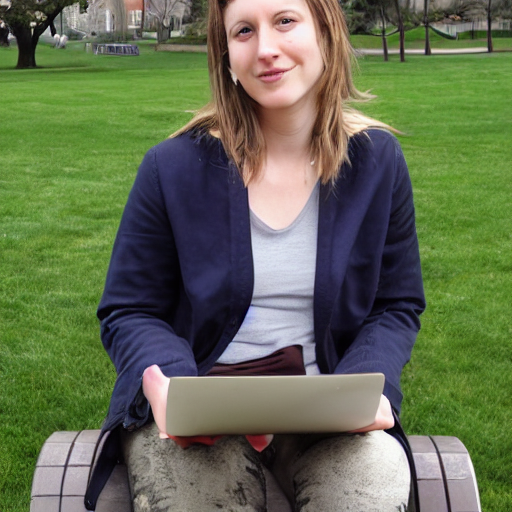}}&
    \raisebox{-0.45\height}{\includegraphics[width=\teaserwid]{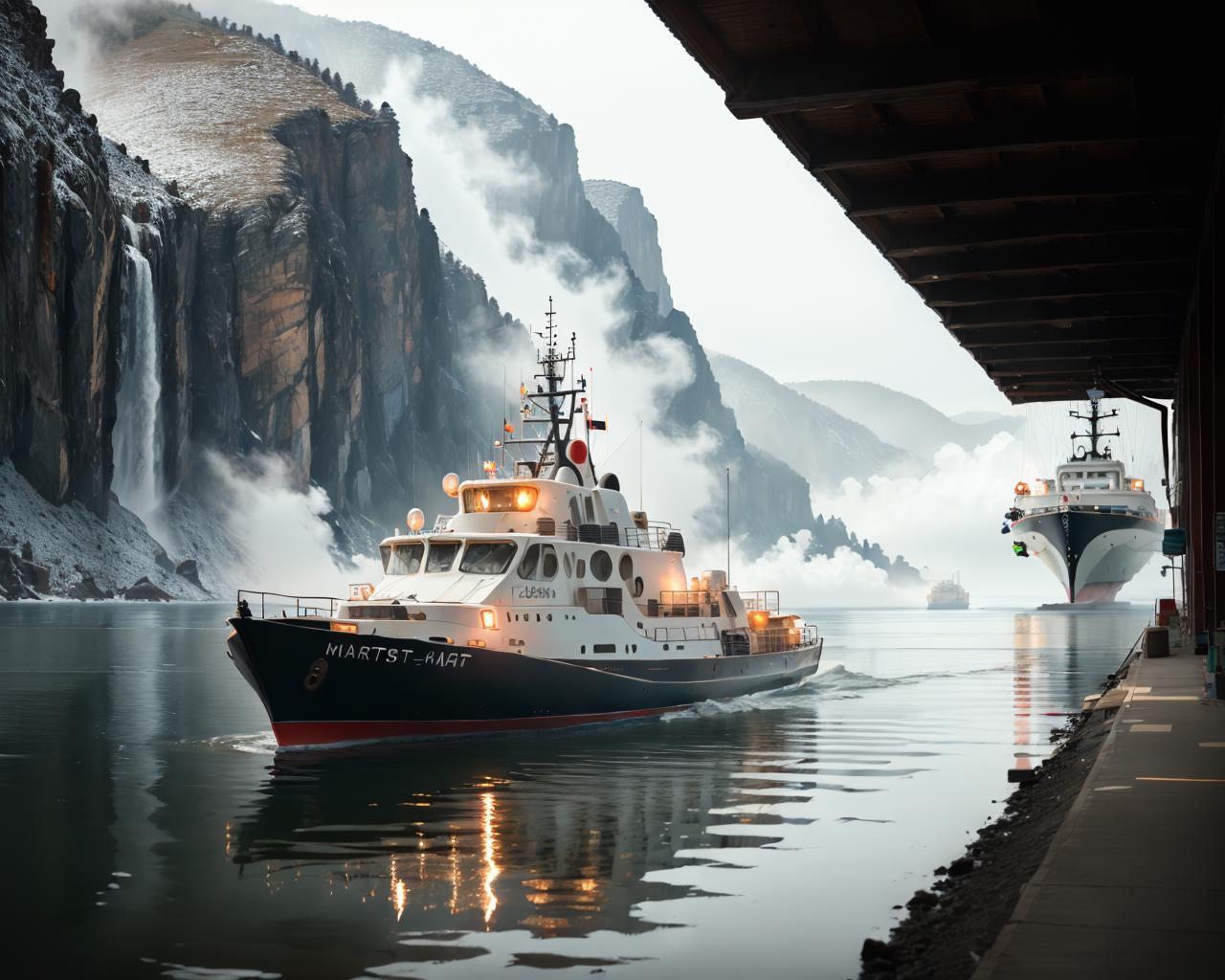}}&
    \raisebox{-0.45\height}{\includegraphics[width=\teaserwid]{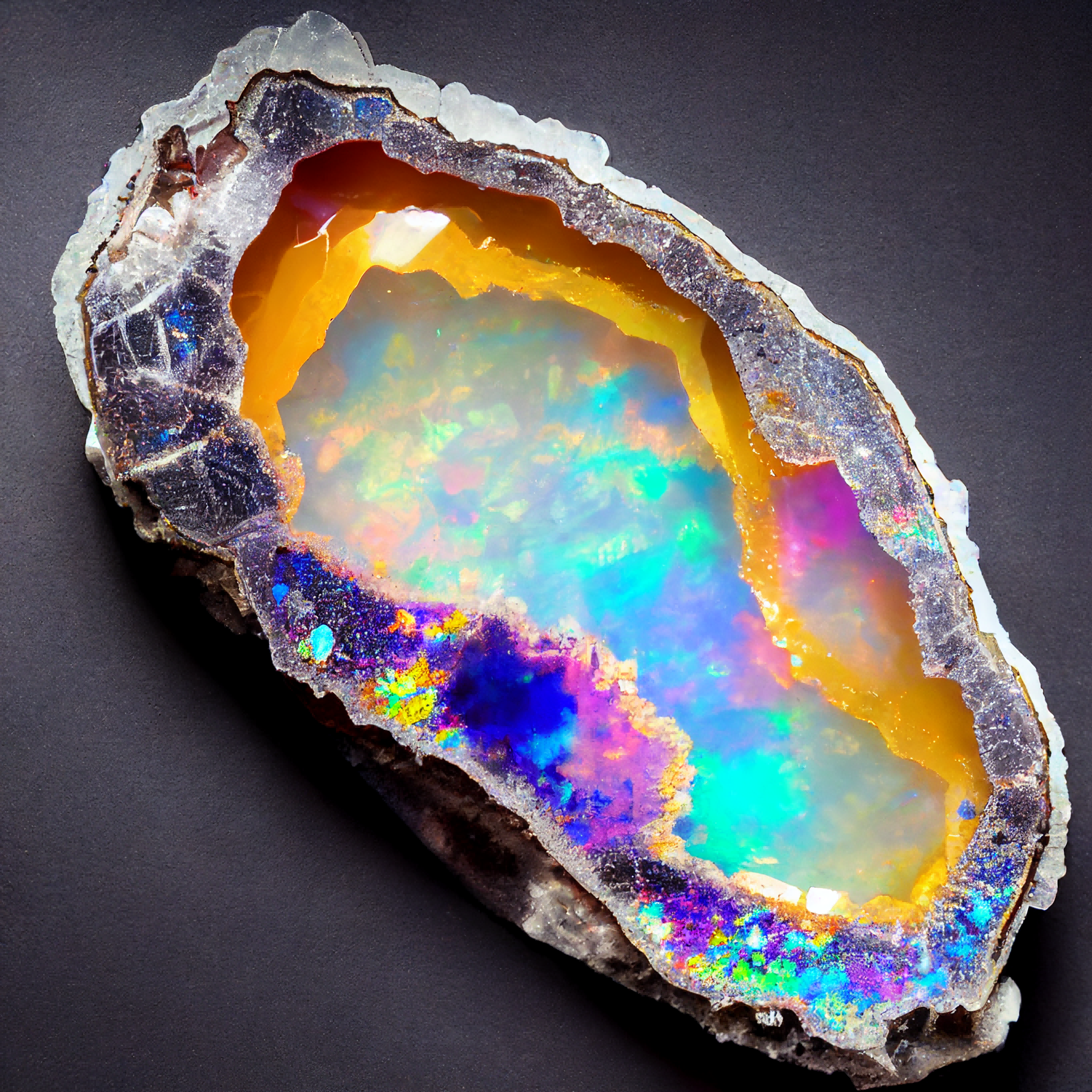}}&
    \raisebox{-0.45\height}{\includegraphics[width=\teaserwid]{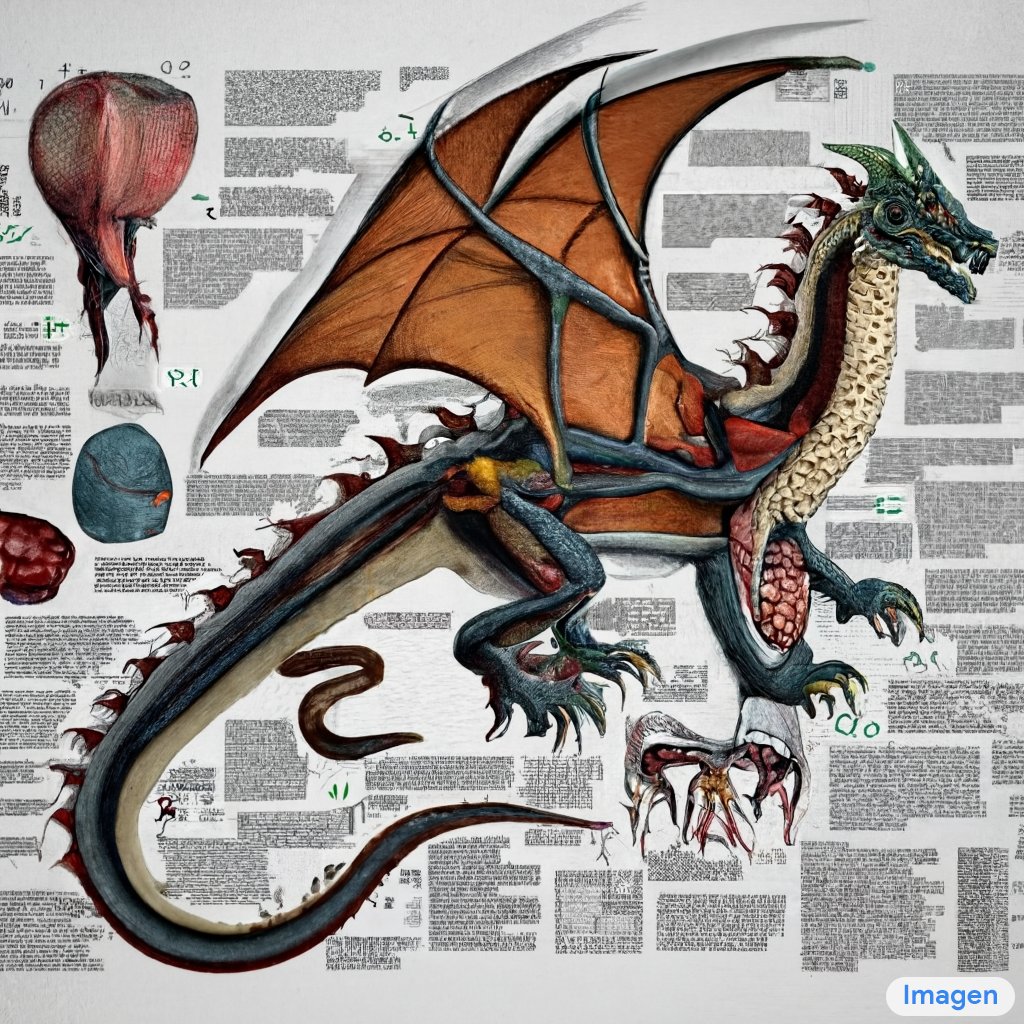}}&
    \raisebox{-0.45\height}{\includegraphics[width=\teaserwid]{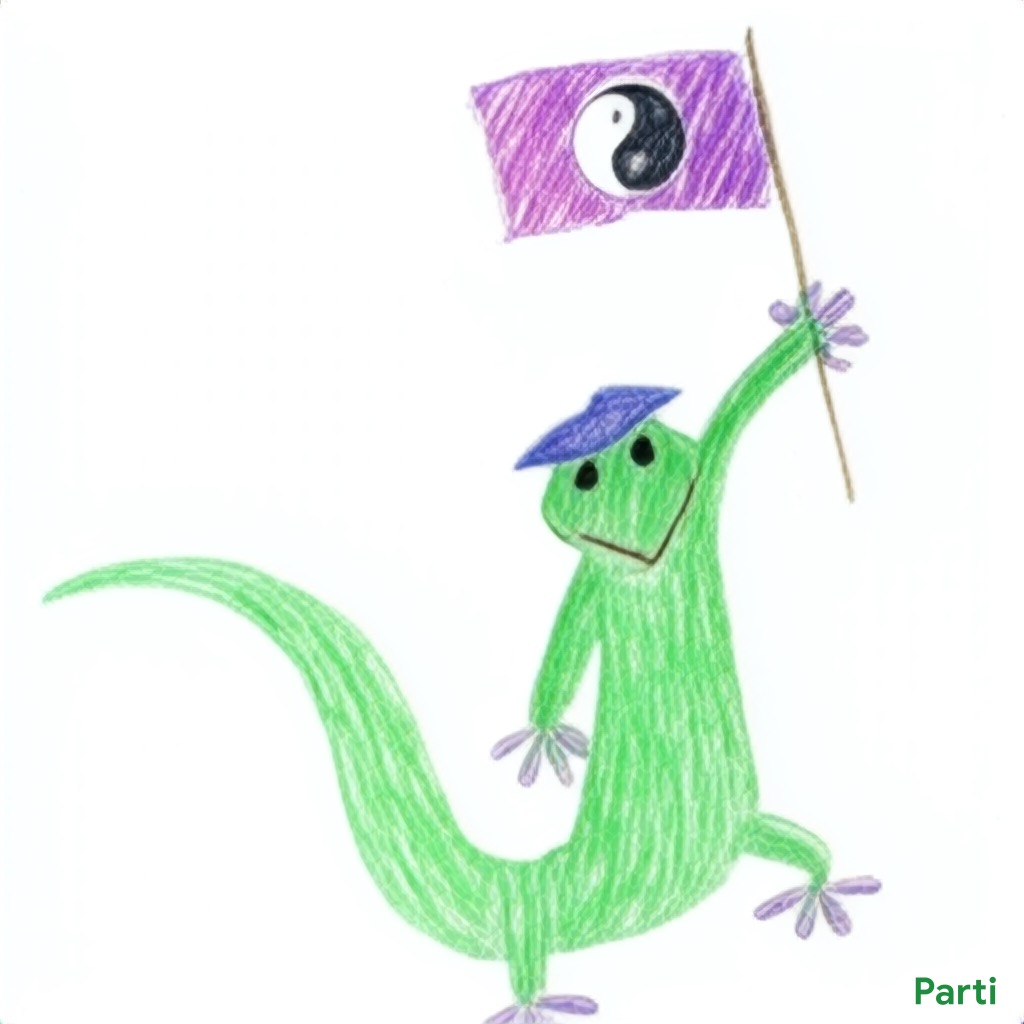}}&
    \vspace{.05in}
    \\
    \raisebox{-0.45\height}{\includegraphics[width=\teaserwid]{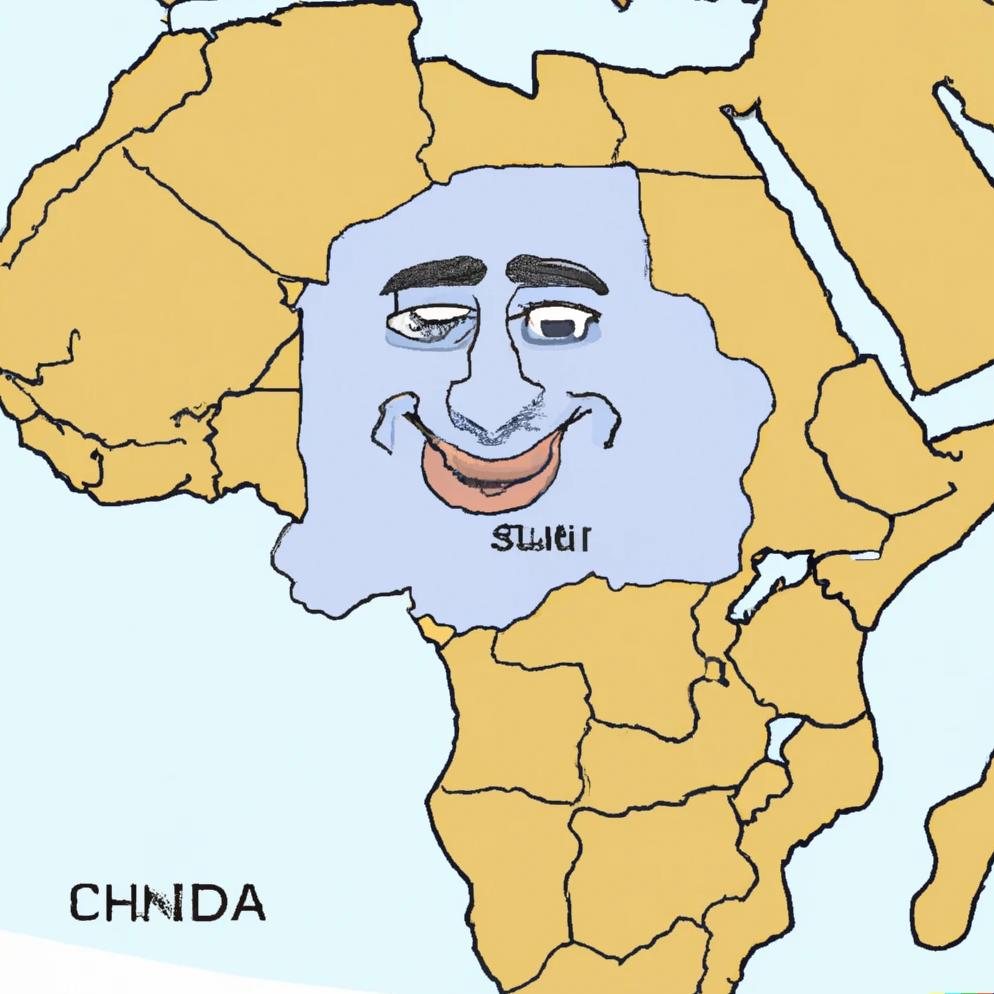}}&  
    \raisebox{-0.45\height}{\includegraphics[width=\teaserwid]{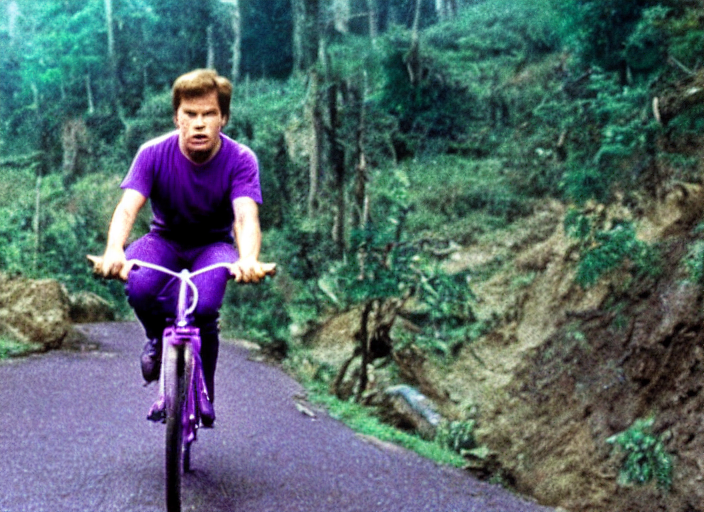}}&
    \raisebox{-0.45\height}{\includegraphics[width=\teaserwid]{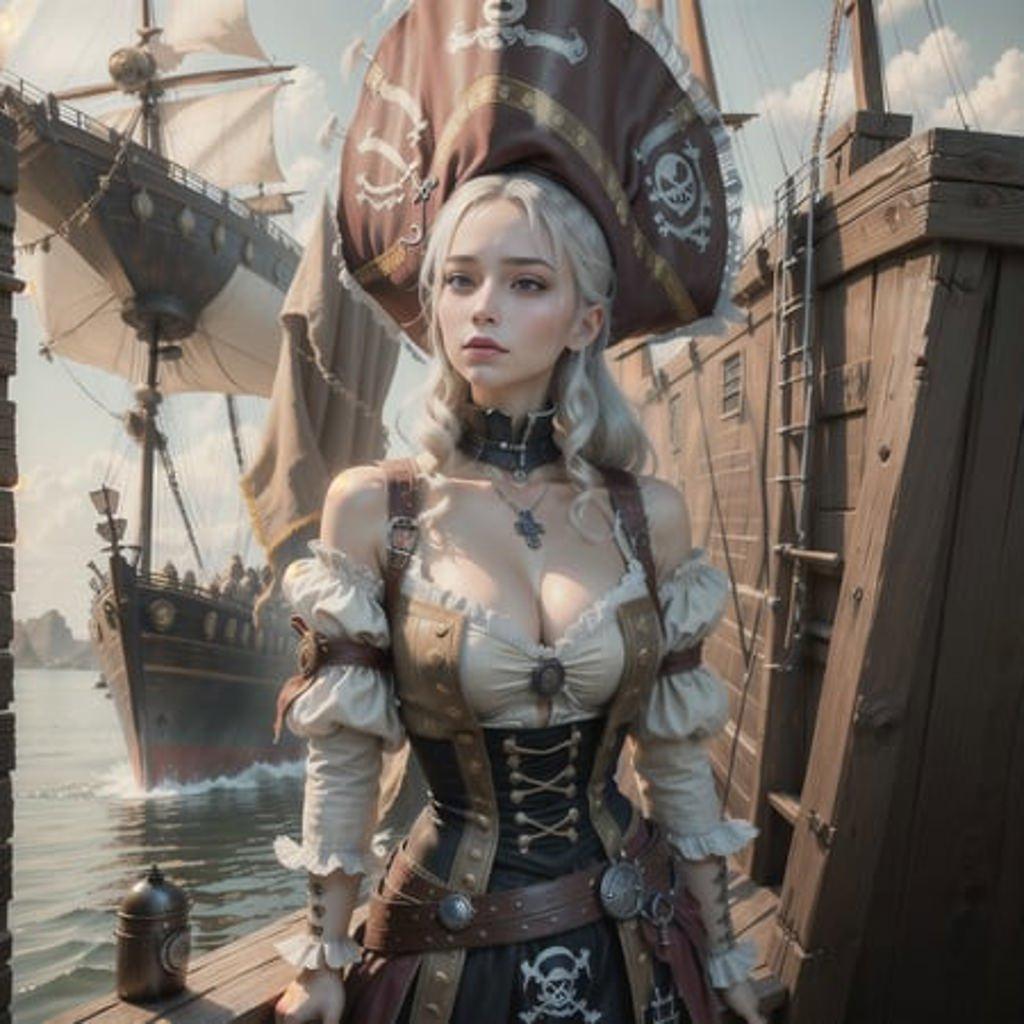}}&
    \raisebox{-0.45\height}{\includegraphics[width=\teaserwid]{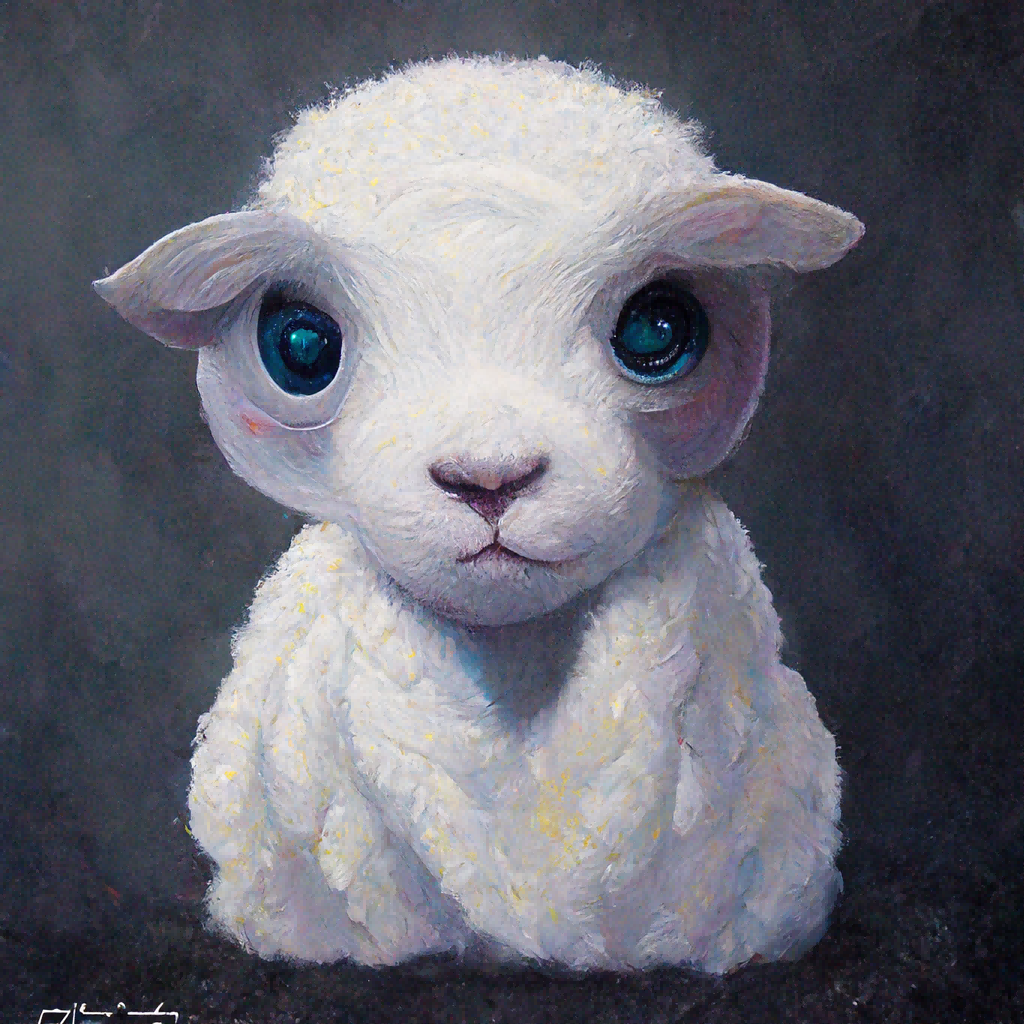}}&
    \raisebox{-0.45\height}{\includegraphics[width=\teaserwid]{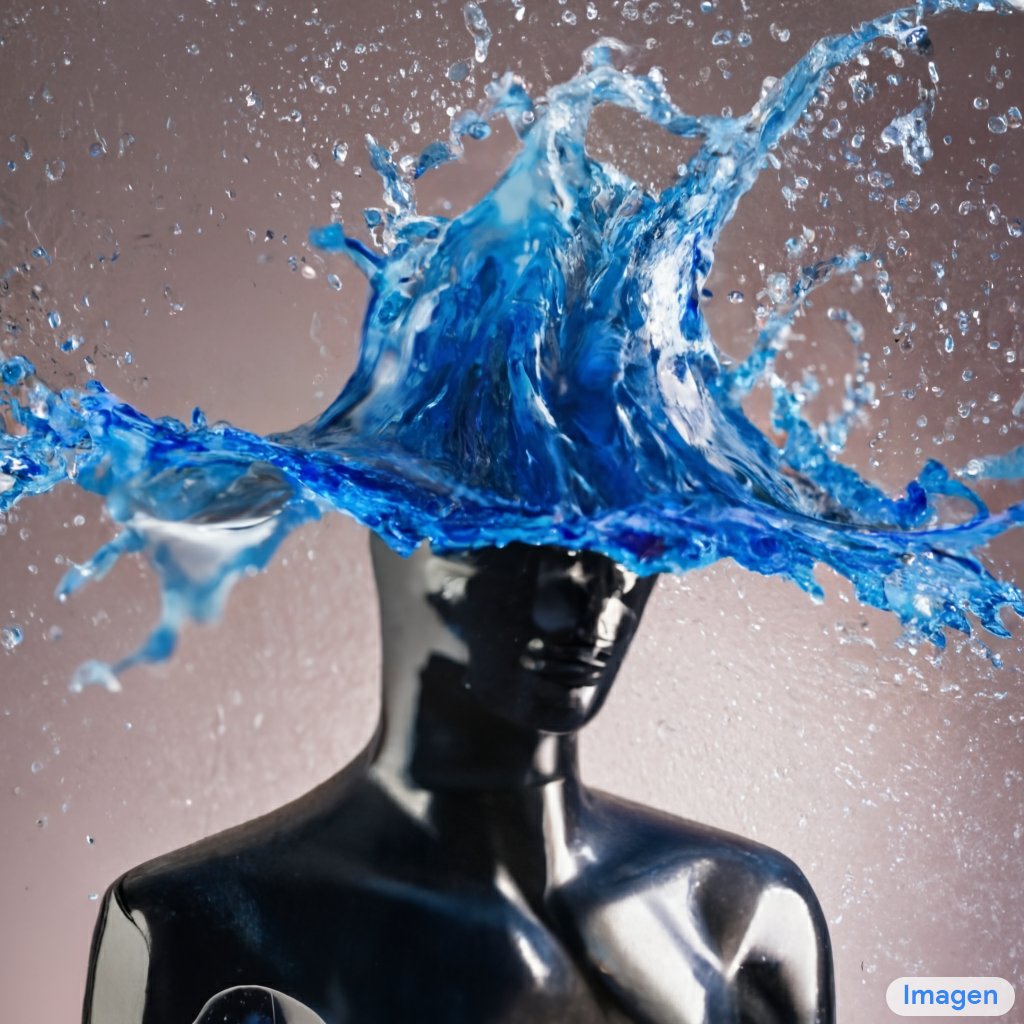}}&
    \raisebox{-0.45\height}{\includegraphics[width=\teaserwid]{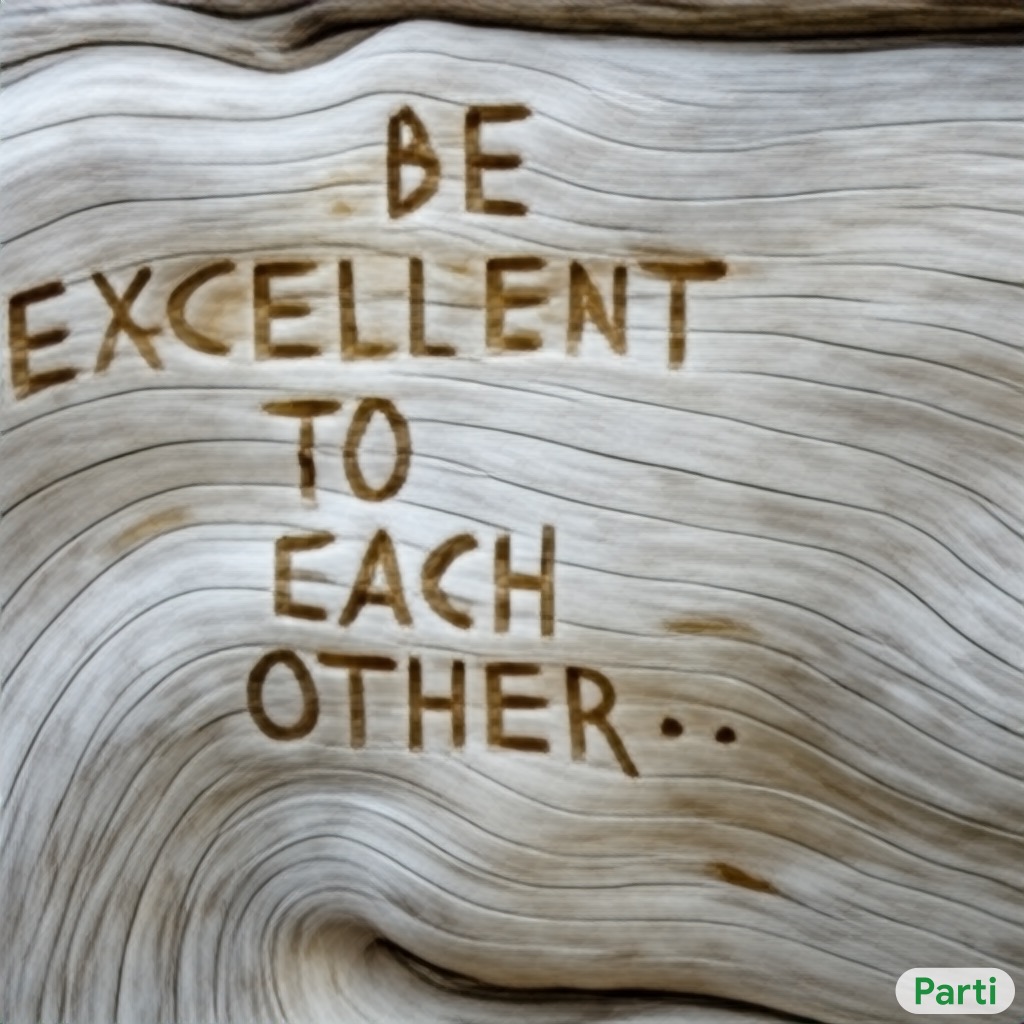}}&
    \vspace{.05in}
    \\
    \raisebox{-0.45\height}{\includegraphics[width=\teaserwid]{Figs/samples/dalle/277975157_1163397197807381_4055136767262611038_n.jpg}}&  
    \raisebox{-0.45\height}{\includegraphics[width=\teaserwid]{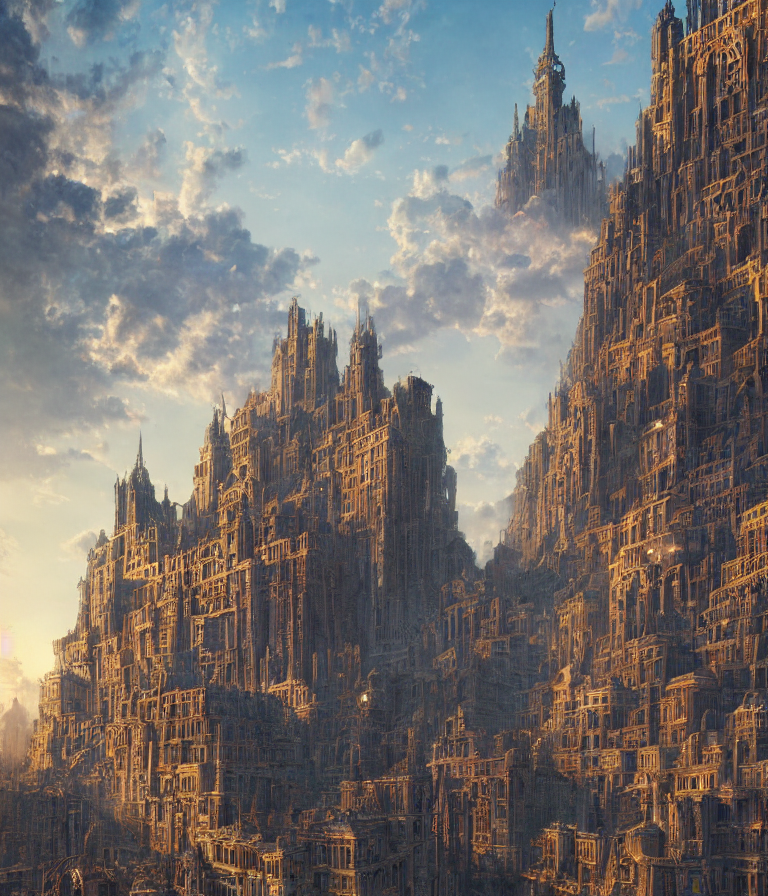}}&
    \raisebox{-0.45\height}{\includegraphics[width=\teaserwid]{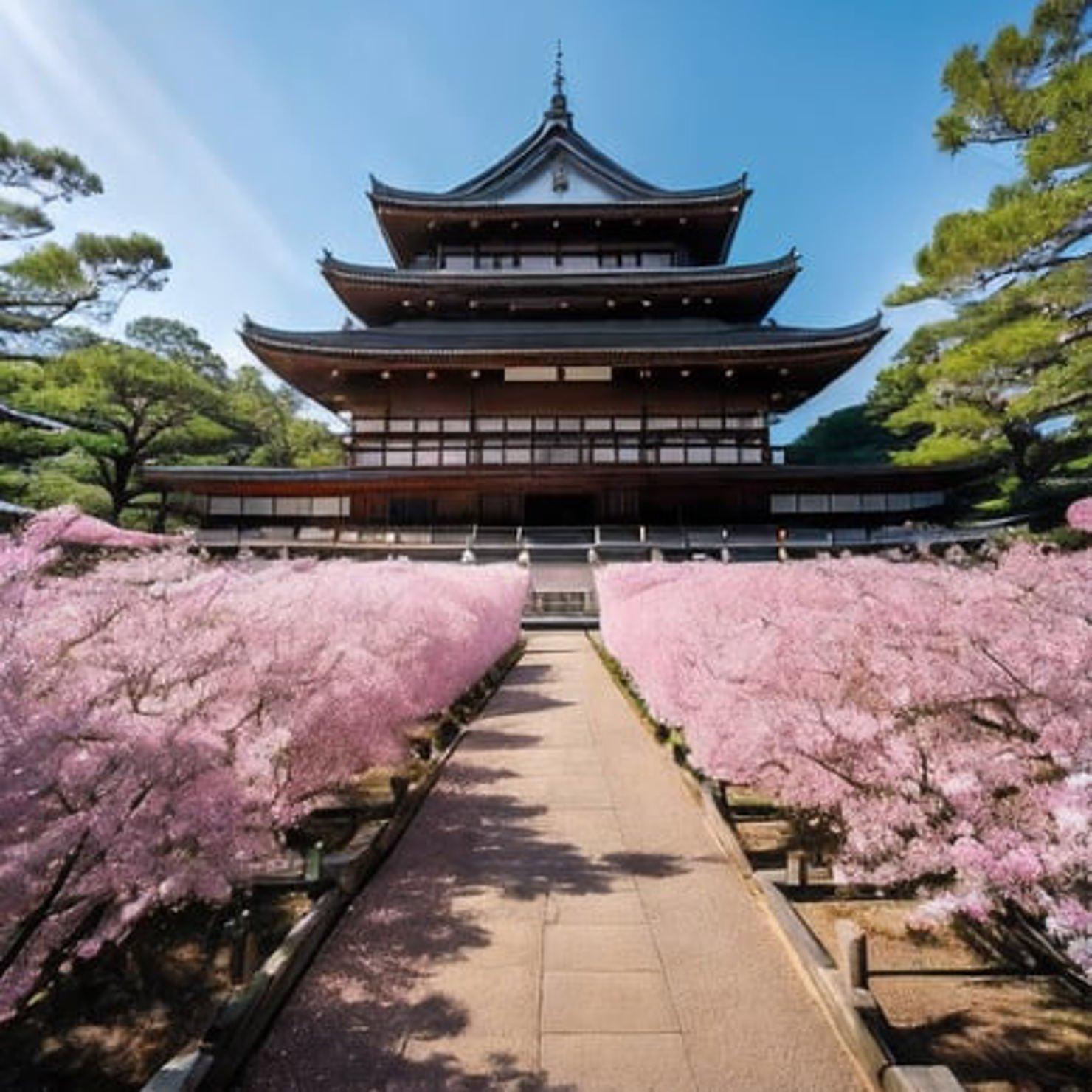}}&
    \raisebox{-0.45\height}{\includegraphics[width=\teaserwid]{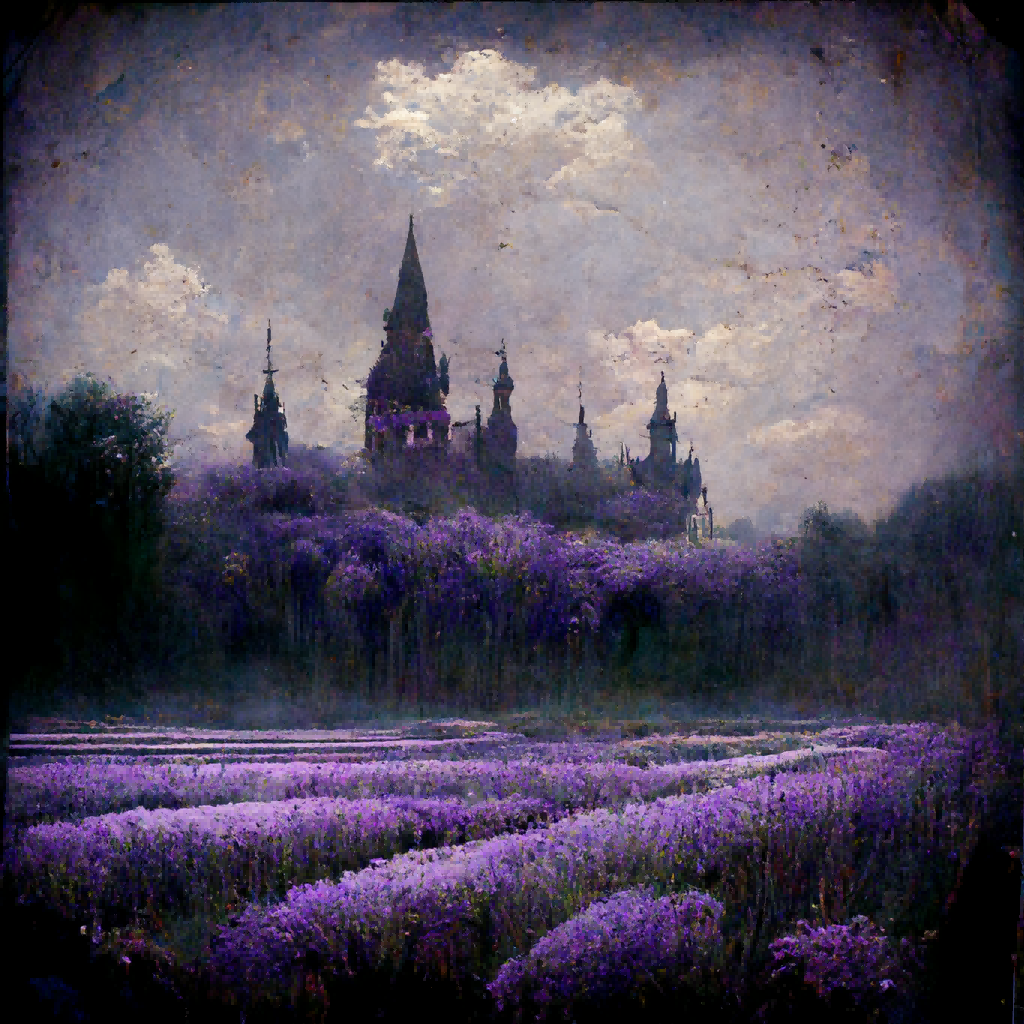}}&
    \raisebox{-0.45\height}{\includegraphics[width=\teaserwid]{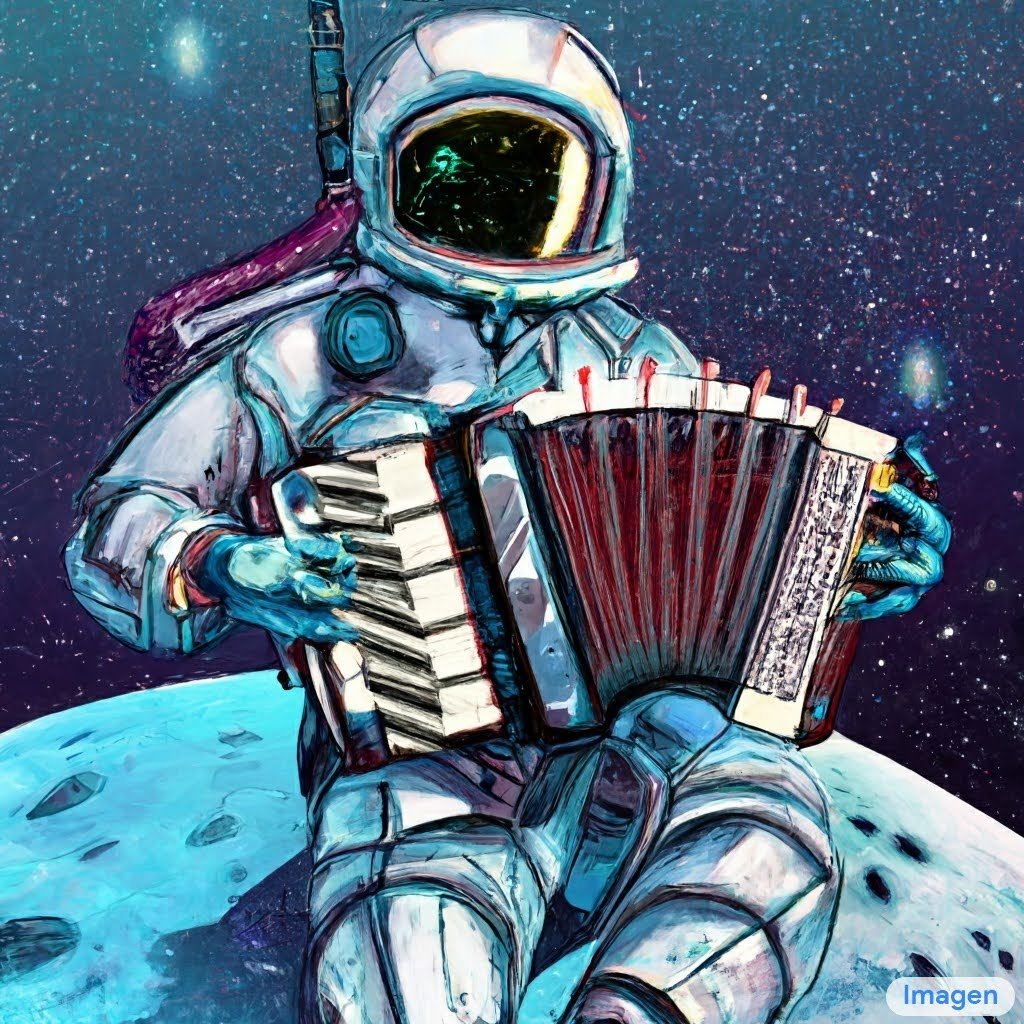}}&
    \raisebox{-0.45\height}{\includegraphics[width=\teaserwid]{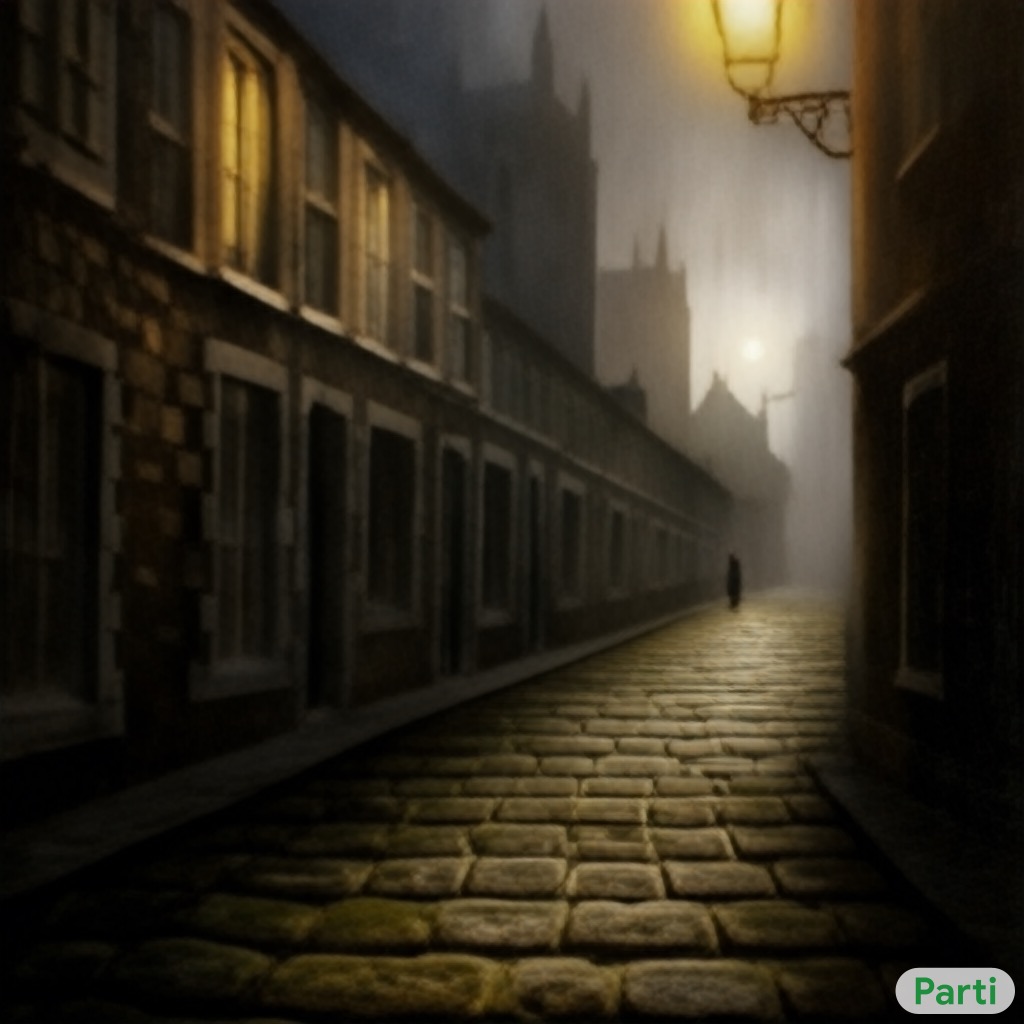}}
    \vspace{.05in}
    \\
    DALL·E\cite{ramesh2022hierarchical} &  
    SD \cite{rombach2021highresolution} &
    PD &
    MJ \cite{david2022mj}&
    Imagen \cite{saharia2022photorealistic} &
    Parti \cite{yu2022scaling}
    \\
    \end{tabular}
}
\caption{Examples of the established DFLIP-3K. The examples are deepfakes produced by generative models. Our data and code will be released.}
\label{fig:samples}
\end{figure}

Figure~\ref{fig:samples} exhibits a collection of randomly selected samples of DFLIP-3K generated by six distinct groups of generative models.

\section{Benchmark Details}

\subsection{Metrics}
Following~\cite{wang2020cnn}, we utilize the task-wise average detection accuracy, which calculates the average test accuracy of all generative models. 
The detection accuracy is computed as follows:

\begin{equation}
A_{det}=\frac{1}{n}\sum_{i=1}^{n}B_{n},
\label{eq:aa}
\end{equation}
where $B_{i}$ represents the test accuracy of the $i$-th deepfake model, and $n$ is the total number of deepfake methods. This metric provides an evaluation of the overall performance of the detector.

We suggest using multiclass accuracy to report the identification accuracy, which calculates the accuracy of classifying images correctly into either the  deepfake models or the group of real images. The calculation of the deepfake identification accuracy is:

\begin{equation}
A_{id} = \frac{N^{TP}_{fake}+N^{TN}_{real}}{N_{total}},
\label{eq:ca}
\end{equation}
where $N^{TP}_{fake}$ represents the true positives of deepfakes classified as their corresponding generative models, and $N^{TN}_{real}$ indicates the true negatives of real images classified correctly. $N_{total}$ denotes the total number of test samples. All real images are classified as a single class in the calculation.

To evaluate the quality of predicted prompts and models, which are two critical components of the deepfake generation process, we suggest measuring the similarity between each input image and the corresponding reconstructed deepfake. We use three different metrics to assess the semantic, perceptual, and aesthetic similarities, namely CLIP-Score~\cite{gal2022image}, Learned Perceptual Image Patch Similarity (LPIPS)\cite{zhang2018perceptual}, and LAION-Aesthetic Score\cite{laionaesthetics}, respectively.

\subsection{Training Details}
To optimize our deepfake detection baseline methods, we follow the implementation of CNNDet \cite{wang2020cnn} and S-Prompts \cite{wang2022sprompt} and employ ResNet-50 and ViT-B/16 as backbones, both of which are pre-trained on ImageNet. We then fine-tune these pre-trained models on DFLIP-3K and utilize the recommended data augmentation methods of Blur+JPEG with varying scales of 0.1 and 0.5. 
Additionally, we adopt the CLIP model used by S-Prompts as another baseline and utilize the suggested prompt-tuning strategies on DFLIP-3K. 
All methods are implemented with their original hyper-parameters as claimed in their respective papers.
For deepfake model identification, we implement ViT-B/16 as our robust baseline and utilize the AdamW optimizer with a learning rate of 0.001, a weight decay of 0.01, 300 epochs, and a batch size of 1024. For Flamingo \cite{Alayrac2022FlamingoAV}, we adopt a learning rate of 0.00001 and use a batch size of 1 with 10 epochs for training. It is noteworthy that we only train a single Flamingo model to accomplish all the sub-tasks of the proposed linguistic deepfake profiling.
Our training is conducted on a server equipped with four NVIDIA 4090 GPUs. To facilitate the reproducibility of our work, we will release our source code and implementation as an open-source project. 

\subsection{Testing Details}
For deepfake detection and identification, traditional vision models typically rely on classifier output logits to make predictions. 
In contrast, as a vision-language model, Flamingo takes a different approach by posing a question to the model using the test image as input, specifically `Is this image generated by AI?'. 
The model then provides an answer, which could be `This is an AI-generated image by MidJourney' or `This is a real image.'.
Then we use the string-matching technique to obtain the prediction from the vision-language model.

For prompt prediction during testing, we randomly select $1,000$  images in the test set as reference images. 
As BLIP \cite{li2022blip} merely outputs image captions, we use Stable Diffusion v1.5 to accept its predicted captions for the generation of the reconstructed deepfakes. 
By comparison, Flamingo can jointly identify deepfake models and prompts, and thus we use its predicted models and prompts to generate the reconstructed deepfakes.
Based on our hyper-parameter analysis, we use the most commonly used hyper-parameters to generate the reconstruted images, specifically DPM++ 2M Karras sampler with 30 sampling steps and 7.0 CFG scale. Additionally, we use the mostly used top 10 words as negative prompts: "low quality, low quality, poorly drawn, worst quality, quality low, bad anatomy, normal quality, normal quality, bad hand, lowres bad".
We generate $10$ images for each predicted prompt, and calculate the average score between each generated deepfake and its reference images.

\section{Additional Experiments and Visualization}

In our analysis, we additionally compare our text-to-image deepfake detection database, DFLIP-3K, with existing deepfake detection benchmark datasets. The comparison is presented in Table~\ref{tab:deepfake}. The results demonstrate that DFLIP-3K significantly differs from the previous datasets. Deepfake detectors trained on the ProGAN dataset show excellent performance in detecting images generated by traditional deepfake methods. However, these detectors prove to be nearly ineffective, performing close to random guessing, when applied to the DFLIP-3K database. This disparity suggests that the detection of deepfakes generated through text-to-image methods requires novel approaches for the detection task.

\begin{figure}[http]
  \centering
    \includegraphics[width=0.98\linewidth]{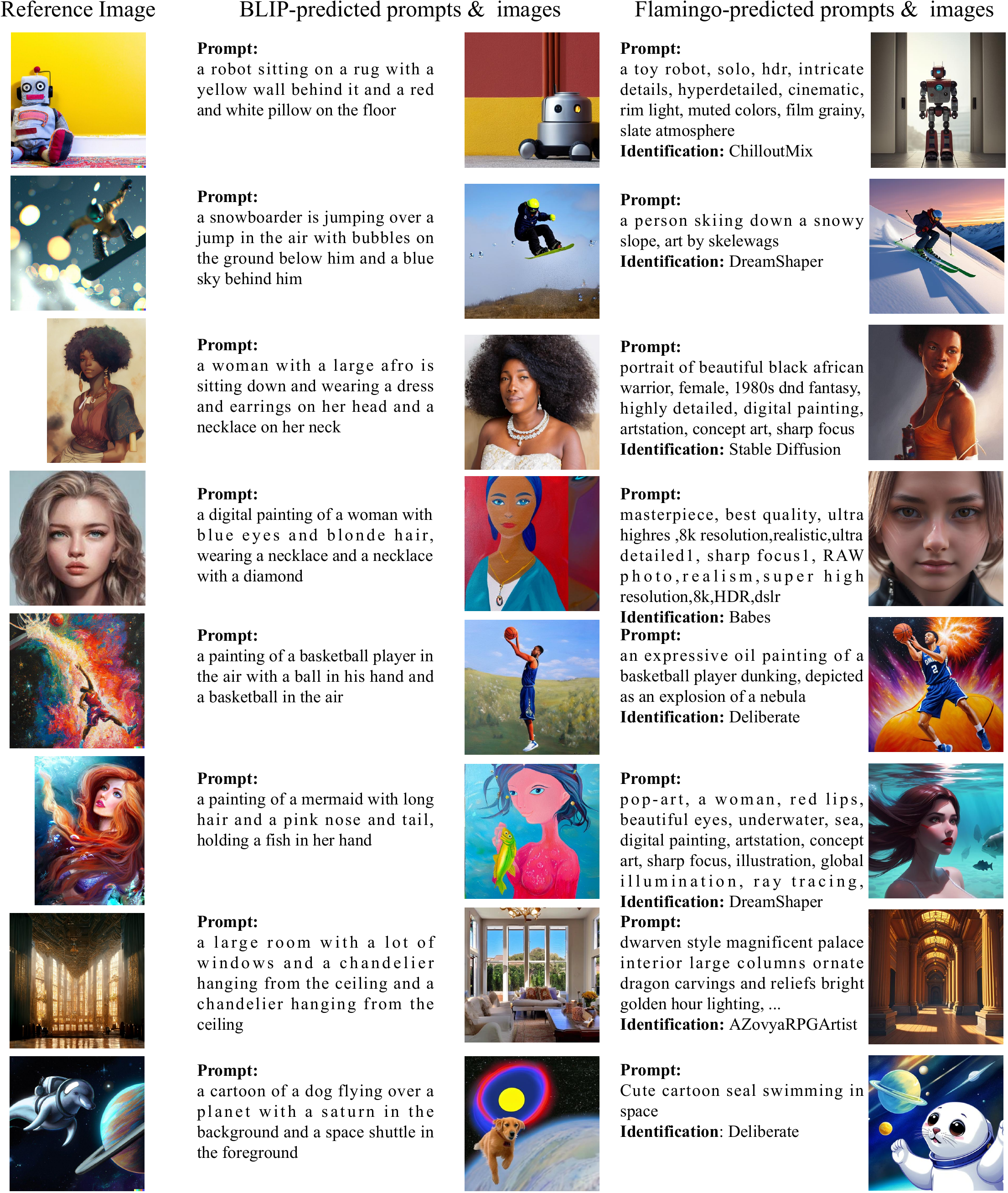}
  \caption{Visualization of Prompt Prediction Results. The first column displays the input reference images used for prompt prediction, while the second and third columns showcase the prompts predicted by BLIP and the corresponding images generated by Stable Diffusion v1.5, respectively. Similarly, the fourth and fifth columns exhibit the fine-tuned Flamingo-predicted prompts and the images generated by models predicted by Flamingo, respectively. \emph{Compared with BLIP, Flamingo can jointly perform model identification and prompt prediction, and its predicted prompts interpret the image contents more faithfully (like `palace' vs. `room'), providing more global styles (like `pop-art') and  more local attributes (like `a basketball player dunking'). Thanks to the better dissection on the two essentials (source model and prompt), Flamigo's resulting images look visually closer to the reference images.} }
  \label{Fig.visp}
\end{figure}

In addition, we present more results of prompt and model prediction in Figure~\ref{Fig.visp}. In comparison to BLIP, Flamingo offers the advantage of performing both model identification and prompt prediction jointly. The predicted prompts by Flamingo demonstrate a more faithful interpretation of the image contents, distinguishing between concepts such as `palace' and `room'. Furthermore, Flamingo provides a broader range of global styles, such as `pop-art', as well as more localized attributes, such as `a basketball player dunking'. This improved dissection and understanding of the two key evidences of deepfake detection, the source model and the source prompt, contribute to Flamingo's ability to generate visually closer images to the reference images.

We follow \cite{wang2020cnn} to visualize the average frequency spectra using Discrete Fourier Transform on each deepfake model' generated images in Figure \ref{fig:vis1}. The results show that the real images (LAION-5B and ImageNet), and our collected deepfake images look very alike (all have very few periodic patterns), while previous GAN-based deepfake images (from GAN models used in \cite{li2022continual}) have many visible patterns (dots or lines). This reflects that the recent T2I deepfakes are very close to the real ones and much more photorealistic than previous deepfakes, showing the great challenge of our suggested deepfake detection.

\begin{table}[]
\resizebox{1\linewidth}{!}{
\begin{tabular}{lrrrrrrrrrr}
\hline
\multicolumn{1}{c}{Methods} & \multicolumn{1}{c}{ProGAN} & \multicolumn{1}{c}{FaceForensic++~\cite{rossler2019faceforensics++}} & \multicolumn{1}{c}{SD} & \multicolumn{1}{c}{PD} & \multicolumn{1}{c}{DALL·E} & \multicolumn{1}{c}{MJ}  & \multicolumn{1}{c}{Imagen} & \multicolumn{1}{c}{Parti}\\ 
\hline
CNNDet(0.1) \cite{wang2020cnn} & 100.00 & 91.39 & 49.50 & 49.34 & 50.25 & 51.45 & 50.45 & 53.36   \\
CNNDet(0.5) \cite{wang2020cnn} & 100.00 & 92.11 & 49.50 & 49.61 & 49.90 & 49.65 & 50.00 & 49.48  \\
\hline
\end{tabular}
}
\caption{Detection accuracy of deepfake detection that \emph{fine-tunes pre-trained model on the ProGAN dataset collected by \cite{wang2020cnn}}. (0.1) and (0.5) mean the augmentation with 10\% or 50\% probability~\cite{wang2020cnn}, respectively. SD: Stable Diffusion, MJ: MidJourney, PD: Personalized Diffusions.}
\label{tab:deepfake}
\end{table}

\yabin{
We want to explore the effectiveness of the huge generator numbers. 
We use a collection of images derived from the Personalized Diffusions model as the training set. The distinction lies in the establishment of three sets, namely Set1, Set2, and Set3, with incrementally increasing model quantities. Set1 contains images from 100 models, Set2 encompasses all models from Set1 plus an additional 400 models, and Set3 includes all models from the previous sets, supplemented by yet another 500 models. We also release this data split in the project page. 
In this setting, both the models and the images were selected at random, ensuring a fixed scale for the training set. Subsequently, the CNNDet-ViT was trained on these varying sets, followed by tests on Out-of-distribution datasets.
The results are shown in Table.~\ref{tab:continualincrease}, which verify that more generators for training lead to better generalization ability.}

\begin{table}[ht]
\centering
\begin{tabular}{l|r|r|r|r|r|r}
\hline
Set & SD & DALLE & MJ & Parti & Imagen & Avg \\
\hline
Set1 & 60.15 & 63.15 & 70.90 & 60.62 & 57.17 & 62.40 \\
Set2 & 67.40 & 60.00 & 79.80 & 52.59 & 61.88 & 64.33 \\
Set3 & 67.55 & 64.25 & 80.35 & 55.96 & 64.79 & 66.58 \\
\hline
\end{tabular}
\caption{Accuracy results with different number of Generators.}
\label{tab:continualincrease}
\end{table}

\section{Related Datasets}

Several datasets focusing on T2I have been proposed, such as DiffusionDB \cite{wang2022diffusiondb}, Simulacra Aesthetic Captions (SAC) \cite{pressmancrowson2022}, ImageReward \cite{xu2023imagereward}, and Pick-a-Pic \cite{kirstain2023pick}. These datasets primarily focus on stable diffusion and investigate the relationship between prompts and generated images to assist in creating aesthetically valuable images that align with human aesthetics.
Additionally, some datasets have been proposed for studying deepfake detection based on the diffusion model. However, these datasets are also limited to a few diffusion models. 

DiffusionDB \cite{wang2022diffusiondb} is a notable example, encompassing a large-scale dataset for the T2I generative model Stable Diffusion. It consists of 2 million images generated by Stable Diffusion 1.4. The dataset was constructed by scraping the official Stable Diffusion Discord server, collecting images, hyper-parameters, and prompts.

Simulacra Aesthetic Captions (SAC) \cite{pressmancrowson2022} is a dataset containing 238,000 human-rated images generated by GLIDE \cite{nichol2021glide} and Stable Diffusion \cite{rombach2021highresolution}. This dataset was created by forty thousand user-submitted prompts, with their aesthetic value ranging from 1 to 10. SAC serves the purpose of training Prompt Generators or evaluating the aesthetic quality of generated images.

ImageReward \cite{xu2023imagereward} employed crowd workers to gather image preference ratings for prompts and images selected from the DiffusionDB dataset. This effort resulted in 136,892 examples originating from 8,878 prompts. 

Pick-a-Pic \cite{kirstain2023pick} is a dataset created by logging user interactions with the Pick-a-Pic web application for T2I generation. It contains over 500,000 examples and 35,000 distinct prompts. Each example includes a prompt, two generated images, and a label indicating which image is preferred or if there is a tie when no image is significantly preferred over the other. The images in this dataset were generated using multiple backbone models such as Stable Diffusion 2.1, Dreamlike Photoreal 2.0, and Stable Diffusion XL variants while sampling different classifier-free guidance scale values.

HP \cite{wu2023better} is another concurrent work that collected a dataset of human judgments by scraping about 25,000 human ratings, including about 100,000 images from the Discord channel of StabilityAI.

Midjourney User Prompts \& Generated Images (250k) dataset,also known as MidJourney (250K)~\cite{turc2023midjourney}, gathers messages from the MidJourney's public Discord server during a four week period from June 20, 2002 to July 17, 2022. The dataset includes user-issued prompts, links to the generated image, and additional metadata. 

\yabin{Deephy~\cite{narayan2022deephy} is a dataset using multiple deepfake generation techniques sequentially to create more sophisticated fake videos that can evade detection. DeePhy containing over 5000 deepfake videos created by applying three different techniques (FSGAN, FaceSwap, FaceShifter) in chains up to three times.}

\yabin{In summary, DFLIP-3K is the first large-scale dataset covering thousands of state-of-the-art T2I deepfake generation techniques, thus facilitating a systematic evaluation of deepfake detection robustness.
In contrast, most existing T2I datasets concentrate on a single diffusion model like Stable Diffusion. Furthermore, datasets designed for studying deepfake detection also cover few generation methods.
The statistic of the comparison are summarized in the Table.~\ref{tab:comparasiondatasets}
DFLIP-3K contains over 350K images generated by applying 3000+ diverse deepfake generators. This makes it uniquely comprehensive in training and evaluating deepfake in real-world scenario. Other datasets are limited to only 1 to 13 generation methods. Additionally, DFLIP-3K provides a multi-dimensional analytical framework, incorporating image, prompt, and technique-level labels.}

\begin{table}[ht]
\centering
\resizebox{\textwidth}{!}{%
\begin{tabular}{llll}
\hline
Dataset & Real Source & Num. Tech & Num. Images(I)/Videos(V) \\ \hline
Deepfake-TIMIT~\cite{korshunov2018deepfakes} & VidTIMIT~\cite{sanderson2009multi} & 2 & 620 (V) \\
FaceForensics++~\cite{rossler2019faceforensics++} & YouTube & 4 & 1.8 million (I) \\
Celab-DF v2~\cite{li2020celeb} & YouTube & 1 & 5,639 (V) \\
DFDC~\cite{dolhansky2020deepfake} & Actors & 8 & 128,154 (V) \\
WildDeepfake~\cite{zi2020wilddeepfake} & Internet & Unknown & 7,314 (V) \\
CNNfake~\cite{wang2020cnn} & Multi-Datasets & 11 & 72.3k (I) \\
CDDB~\cite{li2022continual} & Multi-Datasets\&Internet & 13 & 825,440 (I) \\
Deephy~\cite{narayan2022deephy} & Youtube & 3 & 5,040 (V) \\
DiffusionDB~\cite{wang2022diffusiondb} & NA & 1 & 2 million (I) \\
SAC~\cite{pressmancrowson2022} & NA & 4 & 238,000 (I) \\
ImageReward~\cite{xu2023imagereward} & NA & 1 & 137k (I) \\
Pick-a-Pic~\cite{kirstain2023pick} & NA & 2 & 584,747 (I) \\
HP~\cite{wu2023better} & NA & 1 & 98,803 (I) \\
Midjourney~\cite{turc2023midjourney} & NA & Unknown & 250k (I) \\
DFLIP-3K & LAION-5B & 3K+ & 357,118 (I) \\
\hline
\end{tabular}
}
\caption{Comparison of different datasets.}
\label{tab:comparasiondatasets}
\end{table}